\documentclass[dvipsnames]{article}
\PassOptionsToPackage{table}{xcolor}
\usepackage[T1]{fontenc}
\usepackage{unipat_report}
\usepackage{graphicx}
\usepackage{xurl}
\usepackage{tcolorbox}
\tcbuselibrary{skins}

\usepackage{amsmath,amsfonts,bm}

\def\eqref#1{equation~\ref{#1}}

\def\1{\bm{1}}

\DeclareMathAlphabet{\mathsfit}{\encodingdefault}{\sfdefault}{m}{sl}
\SetMathAlphabet{\mathsfit}{bold}{\encodingdefault}{\sfdefault}{bx}{n}

\usepackage[table]{xcolor}
\usepackage{booktabs}
\usepackage{multirow}
\usepackage{graphicx}
\usepackage{amsmath,amssymb}
\usepackage{enumitem}
\setlist{nosep,leftmargin=1.2em}
\usepackage{xspace}
\usepackage{caption}
\usepackage{wrapfig}
\usepackage[section]{placeins}   
\usepackage{tikz}
\usetikzlibrary{positioning,arrows.meta,calc,fit,backgrounds,decorations.pathreplacing}
\usepackage[capitalize,noabbrev]{cleveref}

\newcommand{\bench}{\textsc{CivWorldBench}\xspace}
\newcommand{\Lone}{L1\xspace}      
\newcommand{\Ltwo}{L2\xspace}      
\newcommand{\Lthree}{L3\xspace}    
\newcommand{\QNT}{\textsc{Quantity}\xspace}
\newcommand{\EVT}{\textsc{Event}\xspace}
\newcommand{\TRT}{\textsc{Trait}\xspace}
\newcommand{\RUL}{\textsc{Rules}\xspace}
\newcommand{\INT}{\textsc{Intervene}\xspace}
\newcommand{\omniID}{\textsc{Omni-ID}\xspace}
\newcommand{\omniOOD}{\textsc{Omni-OOD}\xspace}
\newcommand{\ego}{\textsc{Ego}\xspace}
\newcommand{\expert}{\textsc{Expert}\xspace}
\newcommand{\ceilingline}{\textsc{Ceiling}\xspace}
\newcommand{\floorline}{\textsc{Floor}\xspace}
\definecolor{tableTeal}{HTML}{176D72}
\definecolor{tableRose}{HTML}{A04455}
\definecolor{tableGray}{HTML}{646A70}
\newcommand{\headroompct}[2]{\,{\fontsize{5.5}{6}\selectfont\textcolor{#1}{(#2\%)}}}
\newcommand{\tabledash}{\noalign{\vskip 2pt\hbox to \linewidth{\color{tableGray!70}\leaders\hbox{\rule{2pt}{0.3pt}\hspace{2pt}}\hfill}\vskip 2pt}}
\newsavebox{\promptboxcontent}
\newenvironment{promptbox}[1][]{%
  \setlength{\fboxsep}{8pt}%
  \begin{lrbox}{\promptboxcontent}%
  \begin{minipage}{\dimexpr\linewidth-2\fboxsep-2\fboxrule\relax}%
  \small\setlength{\parindent}{0pt}\setlength{\parskip}{4pt}%
  \textbf{#1}\par\smallskip
}{%
  \end{minipage}\end{lrbox}%
  \noindent\fcolorbox{tableGray!60}{black!2}{\usebox{\promptboxcontent}}%
}
\newcommand{\dox}{\ensuremath{\mathrm{do}}}

\newcommand{\Sset}{\ensuremath{\mathcal{S}}}

\newcommand{\Wset}{\ensuremath{\mathcal{W}}}
\newcommand{\st}[1]{\ensuremath{s_{#1}}}
\newcommand{\vis}[1]{\ensuremath{x_{#1}}}
\newcommand{\hid}{\ensuremath{\theta}}
\newcommand{\obs}[1]{\ensuremath{o_{#1}}}
\newcommand{\hist}[1]{\ensuremath{o_{\le #1}}}
\newcommand{\obsfn}[1]{\ensuremath{O_{#1}}}
\newcommand{\trans}{\ensuremath{T}}
\newcommand{\noise}[1]{\ensuremath{u_{#1}}}

\newif\ifdraft \draftfalse

\newcommand{\nWorlds}{475}

\newcommand{\nWorldsTeam}{60}
\newcommand{\nWorldsClassic}{128}
\newcommand{\nAnchors}{1{,}900}

\newcommand{\nRecords}{561{,}301}   

\newcommand{\nPatches}{5}
\newcommand{\nArmsPT}{7{,}104}
\newcommand{\nRows}{18{,}499}   
\newcommand{\nRowsOneA}{3{,}600}
\newcommand{\nRowsOneB}{3{,}600}
\newcommand{\nRowsTwo}{3{,}360}

\newcommand{\nTemplates}{11}   

\newcommand{\tokS}{17.9K}
\newcommand{\tokM}{47.6K}
\newcommand{\tokL}{78.1K}
\newcommand{\ceilOneABasic}{+0.68}

\newcommand{\ceilOneB}{+0.11}
\newcommand{\ceilPIT}{0.49}
\newcommand{\ceilTraits}{0.66}   
\newcommand{\ceilTraitsRange}{0.56--0.62}
\newcommand{\ceilTraitsGapForty}{0.63}   
\newcommand{\ceilRules}{0.93 / 0.86 / 0.90}
\newcommand{\ceilDisasters}{0.79 / 0.59 / 0.72}
\newcommand{\ceilDirAll}{0.71 / 0.71 / 0.68}
\newcommand{\ceilTraitsAll}{0.59}
\newcommand{\ceilCapture}{+0.39 / +0.39 / +0.30}

\newcommand{\ceilCapturePooled}{+0.29 / +0.29 / +0.20}
\newcommand{\ceilDirIdentifiable}{0.92}
\newcommand{\fracDirIdentifiable}{48\% (basic) / 42\% (generalization)}   
\newcommand{\nCeilAnchors}{92}
\newcommand{\nCeilReruns}{8}
\newcommand{\ceilBankruptTen}{-0.43}
\newcommand{\ceilBankruptPT}{+0.29}

\newcommand{\nApiAnchors}{190}
\newcommand{\nApiSampleFraction}{10\%}

\newcommand{\nHumanAnchors}{\nApiAnchors}
\newcommand{\nManualWorlds}{447}
\newcommand{\humOneA}{+0.37 / +0.29 / +0.37}

\newcommand{\nLinesShort}{eleven}

\newcommand{\nRowsThree}{7{,}939}   
\newcommand{\ceilRulesID}{0.93}
\newcommand{\ceilDirID}{0.92}

\newcommand{\directEffectSign}{0.77 / 0.75 / 0.78}

\newcommand{\directEffectNoMove}{2--4\%}
\newcommand{\directEffectRatio}{1.1--1.5}

\newcommand{\studyPairedGain}{+0.086}
\newcommand{\studyPairedInterval}{[+0.061, +0.111]}

\newcommand{\magnitudeOvershootRange}{17--35\%}

\newcommand{\pooledR}{0.73}
\newcommand{\pooledP}{0.009}
\newcommand{\pooledPairs}{20}
\newcommand{\pooledModels}{10}
\newcommand{\pooledRnoT}{0.73}
\newcommand{\pooledPnoT}{0.006}
\newcommand{\aaPermP}{0.007}

\renewcommand{\today}{October 12, 2026}
\hypersetup{
  pdftitle={CivWorldBench: Evaluating Language World Models in Structured Multi-Agent Sandbox Worlds},
  pdfauthor={Ang Li; Haofei Yu; Zhengwei Tao; Liang Chen; Kuan Li; Miaomiao Li; Chenran Cai; Bin Liang; Jiaxuan You; Kam-Fai Wong},
  pdfsubject={Language world models and structured multi-agent sandbox evaluation},
  pdfkeywords={CivWorldBench, world models, benchmark, Freeciv}
}
\title{\bench: Evaluating Language World Models in Structured Multi-Agent Sandbox Worlds}
\author{%
  {\fontsize{10}{13}\selectfont
  Ang Li\textsuperscript{1}, Haofei Yu\textsuperscript{2},
  Zhengwei Tao\textsuperscript{3,4}, Liang Chen\textsuperscript{3,4}\\
  Kuan Li\textsuperscript{3,6}, Miaomiao Li\textsuperscript{1},
  Chenran Cai\textsuperscript{5}, Bin Liang\textsuperscript{1}\\
  Jiaxuan You\textsuperscript{2}, Kam-Fai Wong\textsuperscript{1}\thanks{Corresponding author: Kam-Fai Wong. Contact: \nolinkurl{angli@link.cuhk.edu.hk}.}}\\[0.8em]
  {\fontsize{9}{12}\selectfont
  \textsuperscript{1}The Chinese University of Hong Kong\\
  \textsuperscript{2}University of Illinois at Urbana-Champaign\\
  \textsuperscript{3}UniPat AI\qquad\textsuperscript{4}Peking University\qquad\textsuperscript{5}TME\\
  \textsuperscript{6}The Hong Kong University of Science and Technology}%
}

\begin{document}
\begin{NoHyper}
\maketitle
\end{NoHyper}
\begin{abstract}
The growing use of language models for forecasting motivates a broader evaluation of their
world-model capabilities. Such evaluation must go beyond predictive accuracy to assess
hidden-state inference and intervention prediction.
Real-world forecasting benchmarks offer limited access
to hidden-state ground truth and controlled experiments for evaluating intervention predictions.
We introduce \bench, a
benchmark built on a multi-agent strategy game engine with recorded internal states,
controlled observations, and paired intervention rollouts. Its \nAnchors\ decision points from
\nWorlds\ worlds support three task levels: association forecasting, belief inference, and
intervention prediction.
Controlled observations and reference predictors support attributable evaluation of capability
gaps, with an expert harness and empirical ceiling estimates contextualizing scores.
Stored trajectories make intermediate predictions verifiable, while agent-play experiments
test whether world-model capabilities are usable for decision-making.
The main panel comprises \nLinesShort\ frontier language models spanning different
capability levels. Composite benchmark scores
correlate with real-world forecasting leaderboards on matched model panels, more closely than
with a general-capability index, supporting its use as a proxy for real-world forecasting
capability.
Strong association forecasts do not
consistently extend to belief inference or intervention prediction, and effect magnitudes remain
poorly calibrated despite informative direction predictions. Process verification reveals errors
missed by final-answer evaluation, and structured reasoning improves intervention prediction for
the strongest model in the process study. Agent-play experiments identify useful assisted plans
and further motivate evaluating decision-relevant capabilities beyond association forecasting.
These findings identify hidden-state inference, intermediate-state prediction, and effect
calibration as priorities for world-model training.
\end{abstract}

\begin{center}
  \small
  \github\quad\textbf{Code:} \url{https://github.com/Leon-Francis/CivWorldBench}
\end{center}
\clearpage

\section{Introduction}
\label{sec:intro}

\begin{figure}[!t]\centering
\includegraphics[width=\linewidth]{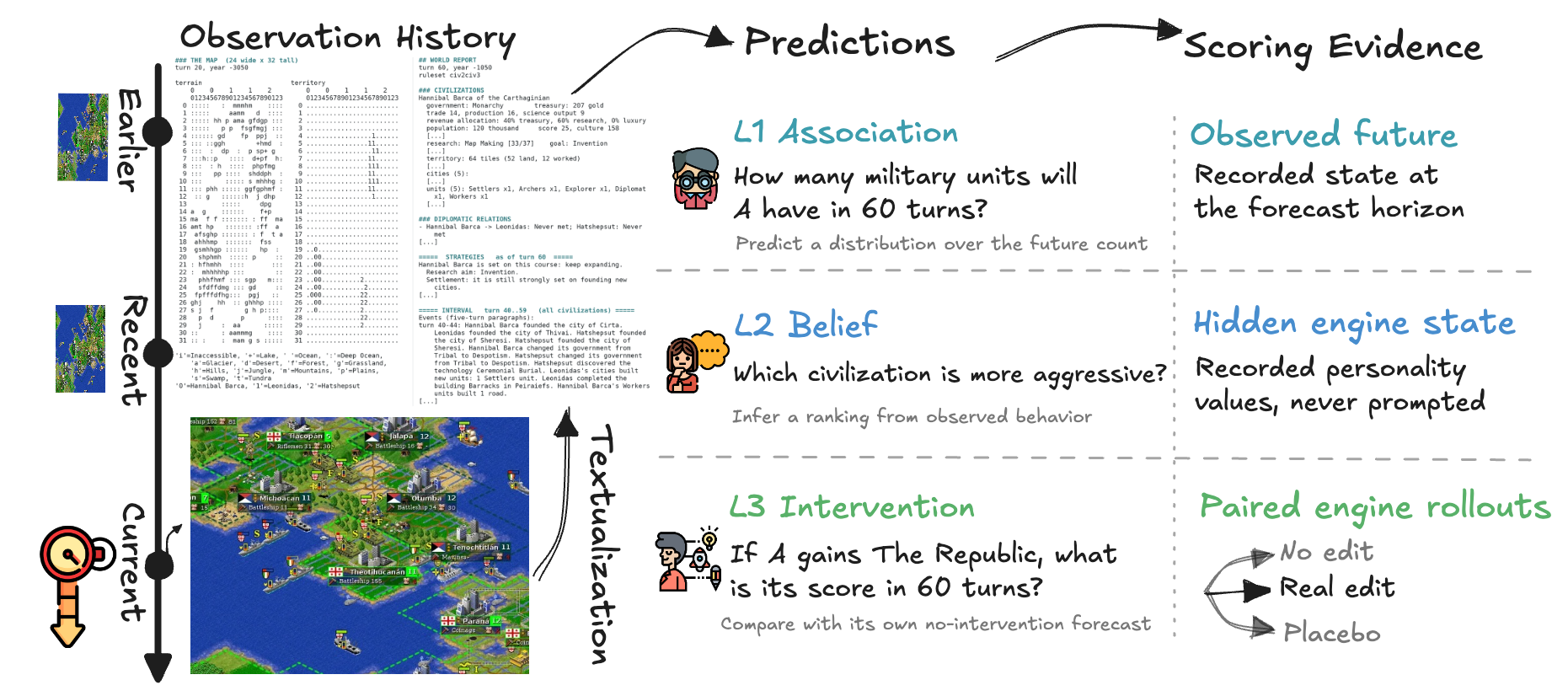}
\caption{Overview of \bench. Textual observation histories support association forecasting,
belief inference, and intervention prediction. Predictions are evaluated against recorded
futures, hidden engine states, and paired engine rollouts, respectively.}
\label{fig:overview}
\end{figure}

Language models increasingly support forecasting and decision-making in multi-agent systems.
They simulate agent populations \citep{park2023generative,piatti2024govsim,piao2025agentsociety},
forecast geopolitical and economic events
\citep{halawi2024approaching,karger2024forecastbench,xu2026forecasting}, negotiate
\citep{fair2022cicero,zhou2024sotopia}, and act in digital, social, and physical
environments \citep{zhu2026agenticlimits}. These applications require a \emph{world model} that predicts
observable dynamics, infers relevant hidden state, and anticipates the consequences of
interventions \citep{chen2026wmdef,chu2026agentic}. Forecasting is the observable output of
such a world model: a forecast commits implicitly to beliefs about hidden state and dynamics,
but its accuracy alone does not establish that these beliefs are correct
\citep{vafa2024evaluating}, and intervention prediction further requires mechanism assumptions
beyond observed associations \citep{pearl2009causality,bareinboim2022hierarchy}.

Existing benchmarks each isolate part of this problem. Real-world forecasting benchmarks such as
ForecastBench \citep{karger2024forecastbench} resolve against genuine outcomes but provide no
latent-state ground truth and no alternative trajectories, so belief and dynamics errors cannot
be separated. ForecastBench-Sim \citep{lee2026forecastbenchsim} evaluates probabilistic
forecasts in Freeciv and demonstrates conditional forecasting with paired intervention rollouts.
Causal-reasoning benchmarks test inference from statistical descriptions
\citep{jin2024corr2cause} or interactive discovery in synthetic laboratories
\citep{yang2026causalab}, and SocietyBench forecasts social events on anonymized timelines
\citep{wang2026societybench}. Game-agent benchmarks such as CivRealm
and BALROG \citep{qi2024civrealm,paglieri2025balrog} score an acting policy, entangling world
knowledge with planning and execution. Isolating world-model capabilities therefore requires
prediction tasks with verifiable hidden states and intervention outcomes, plus controls for
observation limits and stochasticity.

We introduce \bench, a benchmark built on the Freeciv game
engine.\footnote{Freeciv is an open-source turn-based strategy game,
\url{https://www.freeciv.org}.} Civilizations expand, research, trade, cooperate, and compete under
recorded rules and latent behavioral parameters. \Cref{fig:overview} illustrates how textual
observation histories support three task levels: \Lone\ association forecasting (\QNT, \EVT),
\Ltwo\ belief inference (\TRT, \RUL), and \Lthree\ intervention prediction (\INT).
The benchmark contains \nWorlds\ worlds, \nAnchors\ decision points, and \nRows\ prompts.
Global, egocentric, and out-of-distribution settings separate observation limits from
configuration shifts. Saved checkpoints enable paired intervention, control, and placebo
rollouts, and re-seeded futures measure engine stochasticity. \floorline, \expert, and \ceilingline\ references contextualize performance, and the
Sandbox-World Forecasting Index (SWFI) aggregates reference-normalized scores across levels
and settings.

Our evaluation of \nLinesShort\ language models identifies distinct capability profiles.
SWFI correlates with real-world forecasting leaderboards on matched model panels more closely
than with a general-capability index, supporting its use as a forecasting-specific proxy.
Within the benchmark, association forecasting rankings do not
consistently reflect belief and intervention capabilities.
Most models predict intervention directions above chance while estimating effect magnitudes
poorly. Process verification identifies intermediate errors missed by final-answer scores, and
the strongest model in the process study improves effect prediction through structured reasoning.
Agent-play results show that useful advice is not identified by association forecasting
alone.

Our contributions are threefold. First, we enable \emph{attributable evaluation} through a
structured multi-agent benchmark with controlled observations, paired intervention outcomes,
and task-specific reference predictors. Second, we make answers and intermediate predictions
\emph{verifiable} against recorded engine states and rollouts, including Table and Code
reasoning protocols. Third, we assess whether these capabilities are \emph{usable} in agent
decision-making and how they relate to real-world forecasting. Engine patches and evaluation code are available at
\url{https://github.com/Leon-Francis/CivWorldBench}. The benchmark data release is in preparation.

\section{The \bench\ Environment}
\label{sec:env}

\bench\ uses Freeciv to combine multi-agent dynamics with recorded hidden states, controlled
observations, and executable interventions. Each world contains 3--10 AI civilizations that
expand, research, trade, form alliances, and fight over hundreds of turns.

\subsection{State, dynamics, and observations}
A world $w\sim\Wset$ is an engine instance with state $\st{t}=(\vis{t},z_t;\theta_w)$, where
$\vis{t}$ contains reportable map, city, unit, economic, technological, and diplomatic
fields, the hidden dynamic state $z_t$ includes attitudes and internals, and $\theta_w$
contains fixed personality traits and world parameters. Formally, each world is a partially
observable Markov decision process $(\Sset,\mathcal A,\trans,\Omega,\obsfn{v})$ without a
reward, since the benchmark evaluates prediction rather than control. Each built-in agent $i$
follows a policy $\pi_i$ that maps its observation $\obsfn{\text{ego}(i)}(\st{t})\in\Omega$ to
an action in $\mathcal A$, and the joint actions together with the random innovation
$\noise{t}$ induce the transition $\st{t+1}=\trans(\st{t},\{\pi_i\},\noise{t})$. The evaluated world
model is a map $f(\hist{t_1},\{\pi_i\})\mapsto\big(b(z_{t_1},\theta_w),\;\hat p(\st{t_1+h})\big)$
from the observation history and the agents' current policies to a belief over the hidden
state and a forecast of the future state under those policies.

The world model's input at $t_1$ is three-frame history
$\hist{t_1}=(\obs{t_1-2\Delta},\obs{t_1-\Delta},\obs{t_1},\text{narrative},\text{policies})$
with $\Delta=40$ turns, or 20 when $t_1<81$, where each frame $\obs{t}=\obsfn{v}(\st{t})$ is a
text report. The global view $v=\text{omni}$ renders reportable fields across civilizations,
and the egocentric view $v=\text{ego}(i)$ is the engine's own fog-of-war rendering for agent
$i$. The narrative summarizes observed events in five-turn blocks, and the policies give each
agent's research aim and course of action, such as its expansion or war
posture. The output is elicited as quantile forecasts, event probabilities, and hidden-state beliefs,
as \cref{sec:tasks} specifies. \Cref{app:formal} gives the formal definitions.

\subsection{Behavioral validity of hidden targets}
Hidden state includes personality traits that shape strategic preferences, diplomatic
attitudes that influence relations between civilizations, and world parameters governing
disaster frequency, city growth, and research speed. \Ltwo\ infers traits and world
parameters, and \Lthree\ additionally intervenes on attitudes. In Freeciv's default AI, several traits had little or no effect and war was rarely reversed. We apply \nPatches\ source patches to activate
these traits, initialize contact at peace, and permit ceasefires. \Cref{app:corpus,app:traps} document the modifications.

\subsection{World generation}
\begin{wraptable}{r}{0.53\linewidth}
\vspace{-\dimexpr3\baselineskip+0.4ex\relax}
\centering\scriptsize
\setlength{\tabcolsep}{3pt}
\caption{Corpus composition across \nWorlds\ worlds and \nAnchors\ decision points, with
\nRecords\ per-turn state records and \nArmsPT\ collected arms.}
\label{tab:corpus}
\begin{tabular}{@{}llrrrrr@{}}\toprule
Set & Class & Worlds & Points & Agents & Map & Turns \\
\midrule
\multirow{3}{*}{Basic} & small & 75 & 300 & 3 & $24{\times}32$ & 400 \\
 & medium & 75 & 300 & 6 & $32{\times}64$ & 800 \\
 & large & 75 & 300 & 8 & $48{\times}80$ & 1200 \\
\midrule
Gen. & joint & 250 & 1{,}000 & 3--10 & $24{\times}32$--$64{\times}85$ & 400--1200 \\
\bottomrule\end{tabular}
\end{wraptable}
\Cref{tab:corpus} summarizes the two world sets. The \emph{basic} set fixes the ruleset and
difficulty across three map classes while varying hidden parameters and generated worlds.
The \emph{generalization} set jointly varies the ruleset, AI difficulty, map size,
civilization count, hidden-parameter ranges, and team play: \nWorldsClassic\ worlds use the
classic ruleset, which changes the unit, building, and technology sets, five difficulty levels
appear, and \nWorldsTeam\ worlds play in teams, so the shift affects game dynamics rather than
only the map. Each world contributes four decision points, and collection runs beyond the reporting budget
to record forecast targets.
Median \QNT\ prompt lengths are \tokS, \tokM, and \tokL\ tokens for the basic map classes.

\subsection{Interventions}
An intervention applies an edit $m$ to a saved checkpoint before continuing the simulation.
Each eligible checkpoint produces four arms: an unedited \emph{control}, a repeated
\emph{recheck}, a real edit, and a matched \emph{placebo} $m_0$ with negligible or
mechanism-irrelevant impact. Real-arm outcomes are compared with the rerun control rather than the mainline, and placebos
measure responses to nominal edits. Reloads and edits can alter random-number consumption,
so branch differences retain stochastic variation, as \cref{app:branchcollection} documents.

\section{Tasks and Evaluation Settings}
\label{sec:tasks}

The three levels query different components of the process in \cref{sec:env}: \Lone\
predicts future observables $\vis{t+h}$, \Ltwo\ infers the static hidden parameters
$\theta_w$, and \Lthree\ predicts $\vis{t+h}$ after an external edit to $\st{t}$.
\Cref{tab:tasks} summarizes the five tasks.

\begin{table}[!t]\centering\scriptsize
\setlength{\tabcolsep}{3pt}
\caption{Task definitions and evaluation metrics. Prompt counts include all settings
and ablation conditions. Reference ceilings are for \omniID.}
\label{tab:tasks}
\begin{tabular}{@{}cc>{\raggedright\arraybackslash}m{3.6cm}c>{\raggedright\arraybackslash}m{3.2cm}c@{}}
\toprule
Level & Task & Prediction target & Ground truth & Metric (reference $\to$ ceiling) & Prompts \\
\midrule
\multirow{3}{*}{\Lone\unskip} & \QNT & Five quantiles for 11 quantities at two horizons & Mainline & CRPS skill over trend \mbox{($0 \to \ceilOneABasic$)} & \nRowsOneA \\
 & \EVT & Probabilities for \nTemplates\ event templates: ten binary with supplied base rates, one leader & Mainline & Brier gain over reference \mbox{($0 \to \ceilOneB$)} & \nRowsOneB \\
\midrule
\multirow{3}{*}{\Ltwo\unskip} & \TRT & Rank civilizations on four traits & Recorded $\theta_w$ & Adjusted pairwise accuracy \mbox{($0 \to \ceilTraits$)} & \multirow{3}{*}{\nRowsTwo} \\
 & \RUL & Infer three hidden world rules from observable evidence & Recorded $\theta_w$ & Adjusted accuracy \mbox{($0 \to \ceilRulesID$)} & \\
\midrule
\Lthree\unskip & \INT & Quantity forecasts for 11 edits and matched placebos & Paired rollouts & Capture, adjusted direction \mbox{($0 \to \ceilDirID$)}, specificity & \nRowsThree \\
\bottomrule
\end{tabular}
\end{table}

\paragraph{\Lone: Association.}
\QNT\ requests five quantiles for eleven quantities at two budget-dependent horizons,
listed in \cref{app:questions}. Six quantity families, namely economy, scale,
technology--culture, military, diplomacy, and overall score, receive equal weight. \EVT\ requests probabilities for ten binary event templates, including city capture,
wonder completion, and alliance formation, and a distribution over the future score leader.

\paragraph{\Ltwo: Belief.}
\TRT\ ranks civilizations on aggression, expansion, trade, and building preference.
\RUL\ classifies disaster frequency, growth speed, and research speed into three bins each,
subject to evidence requirements. Both tasks use recorded engine values as ground truth.

\paragraph{\Lthree: Intervention.}
Eleven operators modify treasury, research points, cities, government access, map visibility,
personality traits, attitudes, or disaster frequency. The model receives the history and edit description, then re-answers \QNT\ within the
operator's target and horizon range. \emph{Capture} compares the predicted shift from the model's own control
answer with the realized shift between engine branches. \emph{Direction} assesses sign
agreement, and \emph{specificity} measures response selectivity between real and placebo edits.

\paragraph{Settings.}
\omniID\ uses the basic worlds with global observations. \ego\ uses the same worlds through
one civilization's view, with questions scoped to contacted civilizations.
\omniOOD\ applies global observations to the generalization worlds, whose rulesets, AI
difficulty, map sizes, civilization counts, hidden-parameter ranges, and team structures differ
from the basic set, separating restricted information from shifts in world dynamics. \Cref{tab:abl} reports information ablations.

\section{Evaluation Protocol and Research Questions}
\label{sec:protocol}

\subsection{Scores}
\label{sub:scores}
\textbf{Quantity forecasting.}
For each cell $c$, indexed by world, decision point, civilization, quantity, and horizon, let $y$ be the engine
target and $\hat q=(\hat q_\tau)_{\tau\in\mathcal T}$ the predicted quantiles at
$\mathcal T=\{0.1,0.25,0.5,0.75,0.9\}$. Quantile loss (QL) approximates
the continuous ranked probability score (CRPS) \citep{gneiting2007strictly}:
\begin{equation}
\mathrm{QL}(\hat q,y)=\frac{2}{|\mathcal T|}\sum_{\tau\in\mathcal T}
\rho_\tau\!\big(\varphi(y)-\varphi(\hat q_\tau)\big),
\qquad \rho_\tau(u)=\big(\tau-\mathbb 1[u<0]\big)u,
\label{eq:crps}
\end{equation}
where $\rho_\tau$ is the pinball loss and $\mathbb 1[\cdot]$ the indicator.
The transform $\varphi(x)=\operatorname{sign}(x)\log(1+|x|)$ applies to
treasury and population levels and culture increments, and is the identity otherwise.
The normalized cell loss is $\ell_c=\mathrm{QL}(\hat q,y)/\sigma_v$, where $\sigma_v>0$
is the fixed corpus scale for quantity $v$, equal to one for log-transformed quantities.
Enemy and ally counts instead use the absolute error of the predicted median.

For eligible cells $\mathcal C_{wf}$ of family $f$ in world $w$, let $\ell_{c,\mathrm{ref}}$
denote trend-reference loss. World--family skill $S_{wf}$ and the Quantity score
$S_{\mathrm{QNT}}$ are
\begin{equation}
S_{wf}=\operatorname{clip}_{[-1,1]}\!\left(1-
\frac{\sum_{c\in\mathcal C_{wf}}\ell_c}{\sum_{c\in\mathcal C_{wf}}\ell_{c,\mathrm{ref}}}\right),
\qquad S_{\mathrm{QNT}}=\frac{1}{|\mathcal W|}\sum_{w\in\mathcal W}
\frac{1}{|\mathcal F_w|}\sum_{f\in\mathcal F_w}S_{wf},
\label{eq:skill}
\end{equation}
where $\mathcal F_w$ contains families with positive reference loss, $\mathcal W$ contains
worlds with at least one such family, and vertical bars denote cardinality.

\textbf{Event forecasting.}
\EVT\ averages Brier gain \citep{brier1950verification,murphy1988skill} $G=\mathrm{Brier}(p_{\mathrm{ref}},y)-\mathrm{Brier}(\hat p,y)$
for model probability $\hat p$, reference probability $p_{\mathrm{ref}}$, and outcome $y$.
Binary items use $\mathrm{Brier}(p,y)=(p-y)^2$ for event probability $p$ and $y\in\{0,1\}$,
referenced to the supplied task-cell rate. Leader items use multiclass loss
$\sum_j(p_j-\mathbb 1[y=j])^2$, where $p_j$ is the probability that civilization $j$ leads
and $y$ the realized leader, with current-leader persistence as reference.

\textbf{Hidden-state inference.}
\TRT\ uses pairwise ranking accuracy, and \RUL\ averages classification accuracy equally
across rule parameters.

\textbf{Intervention prediction.}
For cell $c$ of quantity $v$, let $\hat q_{m,0.5}$ and $y_m$ denote the predicted median
and engine outcome under edit $m$. Subscript $\mathrm{ctrl}$ denotes the unedited control.
Predicted and realized effects are
$\hat d_c=[\varphi(\hat q_{m,0.5})-\varphi(\hat q_{\mathrm{ctrl},0.5})]/\sigma_v$ and
$d_c=[\varphi(y_m)-\varphi(y_{\mathrm{ctrl}})]/\sigma_v$.
For an operator's eligible cells $\mathcal C$ with nonzero realized effect $d_c$, Capture is
\begin{equation}
C=1-\frac{\sum_{c\in\mathcal C}|\hat d_c-d_c|}{\sum_{c\in\mathcal C}|d_c|},
\label{eq:capture}
\end{equation}
clipped to $[-1,1]$ before averaging equally over operators with at least three eligible cells.
No-effect and exact predictions score 0 and 1.
Direction measures sign agreement among nonzero responses on these cells, counting missing
answers as failures and weighting operators by eligible moved-cell counts.

Let $\bar s_{\mathrm{real}}$ and $\bar s_{\mathrm{pl}}$ denote mean $|\hat d_c|$ under real edits
and matched placebos across valid cells and operators, including unchanged-truth cells.
Valid-response Specificity is
\begin{equation}
\mathrm{Spec}_{\mathrm{valid}}=\operatorname{clip}_{[-1,1]}\!\left(
1-\frac{\bar s_{\mathrm{pl}}}{\bar s_{\mathrm{real}}}\right),
\label{eq:spec}
\end{equation}
defined when $\bar s_{\mathrm{real}}>0$ and both arms have valid shifts. It measures
selectivity, not correctness, and should be interpreted alongside Capture.

\textbf{Reference adjustment.}
Trait, Rules, and Direction report $S_{\mathrm{adj}}=c(a-b)/(c-b)$ for raw accuracy $a$,
baseline $b$, and reference ceiling $c$, which maps the baseline to zero without changing the
ceiling or the model ordering. Missing or unparseable answers receive task-specific failure
penalties, as specified in \cref{app:metrics}.

\subsection{Reference lines and composite index}
\textbf{Reference lines.}
\floorline\ defines metric-specific origins, with Capture's clipped lower endpoint $-1$. \expert\ adds domain knowledge and a tool-assisted
harness under the same observations, without hidden test state, future outcomes, or random
seeds. The conceptual \ceilingline\ assumes
complete simulator state and engine knowledge, but not future random draws.

\textbf{Composite index.}
For model $m$, setting $s$, and metric $k$, let $x_{msk}$ be the score, with floor $f_{sk}$
and ceiling $c_{sk}$. Completion is $r_{msk}=(x_{msk}-f_{sk})/(c_{sk}-f_{sk})$, a fraction that
tables report as a percentage. To define the Sandbox-World Forecasting Index (SWFI), let
$\mathcal S_{\mathrm{eval}}=\{\omniID,\ego,\omniOOD\}$ and
$\mathcal K_\ell$ contain Quantity and Event at level $\ell=1$, Trait and Rules at $\ell=2$,
and Capture, Direction, and Specificity at $\ell=3$. Then
\begin{equation}
\operatorname{SWFI}(m)=\frac{100}{9}\sum_{s\in\mathcal S_{\mathrm{eval}}}
\sum_{\ell=1}^{3}\frac{1}{|\mathcal K_\ell|}\sum_{k\in\mathcal K_\ell}r_{msk}.
\label{eq:swfi}
\end{equation}
Metrics are weighted equally within levels, then across levels and settings. Weights and
reference scales are fixed independently of external leaderboards, and completion and SWFI
are not clipped.

\subsection{Research questions}
\textbf{RQ1, validity.} Do information controls, reference lines, and external forecasting
associations support the evaluation? \textbf{RQ2, limits.} Where do current models fall short
across task levels? \textbf{RQ3, process.} How does intermediate prediction quality relate to
final answers, and how do reasoning protocols affect intervention prediction?
\textbf{RQ4, utility.} Can world models improve agent decisions, and which capability
profiles indicate utility?

\subsection{Models and inference}
We evaluate \nLinesShort\ language models with thinking enabled at model-specific effort
settings recorded in \cref{app:repro}. Open-weight models, including Qwen3.8-Flash-Next,
hereafter Flash-Next, cover the full grid. Claude Opus~5 and GPT-5.6~Sol, hereafter Opus and Sol, use medium reasoning effort on
\nApiAnchors\ shared frozen decision points, \nApiSampleFraction\ of the benchmark, to bound
API cost. This subset is
stratified to match the full benchmark's composition by split, world size, budget, ruleset,
decision time, and operator, one anchor per world, so subset scores estimate the same quantities
as the full grid. \expert\ uses Sol on the same anchors with a manual combining
engine rules and regularities from \nManualWorlds\ disjoint worlds. The Artificial Analysis (AA) Intelligence Index v4.3 \citep{artificialanalysis2026index}
provides a general-capability reference. The reported evaluations consumed about 3,600 H20-equivalent GPU-hours of open-weight
inference, 5,000 CPU core-hours of engine simulation, and \$2,500 of closed-model API calls.

\section{Benchmark Validity and Model Capabilities}
\label{sec:results}

\begin{table}[!b]\centering\scriptsize
\setlength{\tabcolsep}{1.5pt}\renewcommand{\arraystretch}{1.12}
\caption{Model performance on \bench: three-setting SWFI and \omniID\ diagnostics. Bold denotes the best model score, excluding the expert reference. Percentages indicate reference-normalized completion. Full three-setting results are reported in \cref{tab:main_full}.}
\label{tab:main}
\begin{tabular}{@{}>{\raggedright\arraybackslash}p{0.16\linewidth}>{\centering\arraybackslash}p{0.045\linewidth}>{\raggedleft\arraybackslash}p{0.075\linewidth}>{\raggedleft\arraybackslash}p{\dimexpr(0.72\linewidth-32.1pt)/7\relax}>{\raggedleft\arraybackslash}p{\dimexpr(0.72\linewidth-32.1pt)/7+3.6pt\relax}*{4}{>{\raggedleft\arraybackslash}p{\dimexpr(0.72\linewidth-32.1pt)/7\relax}}>{\raggedleft\arraybackslash}p{\dimexpr(0.72\linewidth-32.1pt)/7+1.5pt\relax}@{}}
\toprule
 & & & \multicolumn{2}{c}{\Lone\ association} & \multicolumn{2}{c}{\Ltwo\ belief} & \multicolumn{3}{c}{\Lthree\ intervention} \\
\cmidrule(lr){4-5}\cmidrule(lr){6-7}\cmidrule(lr){8-10}
Model & AA & SWFI & \mbox{Quantity\,$\uparrow$} & \mbox{Event\,$\uparrow$} & \mbox{Trait\,$\uparrow$} & \mbox{Rules\,$\uparrow$} & \mbox{Capture\,$\uparrow$} & \mbox{Direction\,$\uparrow$} & \mbox{Specificity\,$\uparrow$} \\
\midrule
\floorline & \textemdash & 0.00 & +0.00 & +0.000 & 0.00 & 0.00 & $-$1.00 & 0.00 & +0.00 \\
\tabledash
gpt-oss-120b & 12 & $-$3.89 & \mbox{+0.03\headroompct{tableTeal}{4}} & \mbox{$-$0.031\headroompct{tableRose}{-29}} & \mbox{0.01\headroompct{tableTeal}{2}} & \mbox{0.04\headroompct{tableTeal}{5}} & \mbox{$-$0.65\headroompct{tableTeal}{25}} & \mbox{0.15\headroompct{tableTeal}{16}} & \mbox{+0.05\headroompct{tableTeal}{5}} \\
Gemma~4 31B & 15 & 16.52 & \mbox{+0.21\headroompct{tableTeal}{31}} & \mbox{+0.012\headroompct{tableTeal}{11}} & \mbox{0.08\headroompct{tableTeal}{12}} & \mbox{0.11\headroompct{tableTeal}{12}} & \mbox{$-$0.61\headroompct{tableTeal}{28}} & \mbox{0.30\headroompct{tableTeal}{33}} & \mbox{\textbf{+0.19}\headroompct{tableTeal}{19}} \\
Nemotron~3 Ultra & 23 & $-$0.98 & \mbox{+0.08\headroompct{tableTeal}{12}} & \mbox{+0.003\headroompct{tableTeal}{3}} & \mbox{0.08\headroompct{tableTeal}{11}} & \mbox{0.12\headroompct{tableTeal}{13}} & \mbox{$-$0.63\headroompct{tableTeal}{27}} & \mbox{$-$0.16\headroompct{tableRose}{-17}} & \mbox{$-$0.33\headroompct{tableRose}{-33}} \\
Inkling & 26 & 15.73 & \mbox{+0.22\headroompct{tableTeal}{32}} & \mbox{+0.012\headroompct{tableTeal}{11}} & \mbox{0.13\headroompct{tableTeal}{20}} & \mbox{0.11\headroompct{tableTeal}{12}} & \mbox{$-$0.31\headroompct{tableTeal}{49}} & \mbox{0.09\headroompct{tableTeal}{10}} & \mbox{$-$0.11\headroompct{tableRose}{-11}} \\
Qwen3.8-27B & 34 & 12.47 & \mbox{+0.25\headroompct{tableTeal}{36}} & \mbox{+0.007\headroompct{tableTeal}{7}} & \mbox{0.10\headroompct{tableTeal}{16}} & \mbox{$-$0.11\headroompct{tableRose}{-12}} & \mbox{$-$0.34\headroompct{tableTeal}{48}} & \mbox{0.24\headroompct{tableTeal}{26}} & \mbox{$-$0.03\headroompct{tableRose}{-3}} \\
GPT-5.6~Sol & 39 & 27.68 & \mbox{+0.27\headroompct{tableTeal}{39}} & \mbox{+0.022\headroompct{tableTeal}{20}} & \mbox{\textbf{0.23}\headroompct{tableTeal}{34}} & \mbox{0.16\headroompct{tableTeal}{17}} & \mbox{$-$0.60\headroompct{tableTeal}{29}} & \mbox{0.28\headroompct{tableTeal}{30}} & \mbox{$-$0.02\headroompct{tableRose}{-2}} \\
DeepSeek-V4.1-Flash & 40 & 20.48 & \mbox{+0.19\headroompct{tableTeal}{27}} & \mbox{+0.017\headroompct{tableTeal}{16}} & \mbox{0.14\headroompct{tableTeal}{22}} & \mbox{\textbf{0.17}\headroompct{tableTeal}{18}} & \mbox{$-$0.58\headroompct{tableTeal}{31}} & \mbox{0.30\headroompct{tableTeal}{32}} & \mbox{+0.07\headroompct{tableTeal}{7}} \\
Qwen3.8-Flash-Next & 40 & 16.93 & \mbox{+0.26\headroompct{tableTeal}{38}} & \mbox{+0.011\headroompct{tableTeal}{10}} & \mbox{0.10\headroompct{tableTeal}{15}} & \mbox{0.08\headroompct{tableTeal}{9}} & \mbox{$-$0.57\headroompct{tableTeal}{31}} & \mbox{0.22\headroompct{tableTeal}{24}} & \mbox{+0.05\headroompct{tableTeal}{5}} \\
Kimi~K3 & 44 & 21.94 & \mbox{+0.34\headroompct{tableTeal}{50}} & \mbox{\textbf{+0.025}\headroompct{tableTeal}{23}} & \mbox{$-$0.00\headroompct{tableRose}{-0}} & \mbox{0.05\headroompct{tableTeal}{6}} & \mbox{$-$0.35\headroompct{tableTeal}{47}} & \mbox{0.23\headroompct{tableTeal}{25}} & \mbox{+0.08\headroompct{tableTeal}{8}} \\
GLM-5.3 & 45 & 19.19 & \mbox{\textbf{+0.37}\headroompct{tableTeal}{53}} & \mbox{+0.023\headroompct{tableTeal}{21}} & \mbox{$-$0.01\headroompct{tableRose}{-2}} & \mbox{0.07\headroompct{tableTeal}{8}} & \mbox{$-$0.25\headroompct{tableTeal}{54}} & \mbox{0.16\headroompct{tableTeal}{17}} & \mbox{$-$0.05\headroompct{tableRose}{-5}} \\
Claude Opus~5 & 45 & \textbf{29.47} & \mbox{+0.33\headroompct{tableTeal}{48}} & \mbox{+0.021\headroompct{tableTeal}{19}} & \mbox{0.17\headroompct{tableTeal}{26}} & \mbox{0.13\headroompct{tableTeal}{14}} & \mbox{\textbf{$-$0.08}\headroompct{tableTeal}{66}} & \mbox{\textbf{0.33}\headroompct{tableTeal}{35}} & \mbox{+0.12\headroompct{tableTeal}{12}} \\
\tabledash
\expert & \textemdash & 51.18 & \mbox{+0.37\headroompct{tableTeal}{55}} & \mbox{+0.021\headroompct{tableTeal}{20}} & \mbox{0.26\headroompct{tableTeal}{39}} & \mbox{0.52\headroompct{tableTeal}{56}} & \mbox{+0.13\headroompct{tableTeal}{81}} & \mbox{0.58\headroompct{tableTeal}{63}} & \mbox{+0.86\headroompct{tableTeal}{86}} \\
\tabledash
\ceilingline & \textemdash & 100.00 & +0.68 & +0.108 & 0.66 & 0.93 & +0.39 & 0.92 & 1.00 \\
\bottomrule\end{tabular}
\end{table}

\Cref{tab:main} summarizes the main results, with full setting-wise results in \cref{tab:main_full}.
We examine evaluation validity (RQ1), then identify model limitations and directions for
improvement across task levels (RQ2).

\subsection{Do the tasks and metrics support valid evaluation? (RQ1)}
\label{sub:sound}

\begin{figure}[t]
\begin{minipage}[t]{0.52\linewidth}
\vspace{0pt}
\centering
\includegraphics[width=\linewidth]{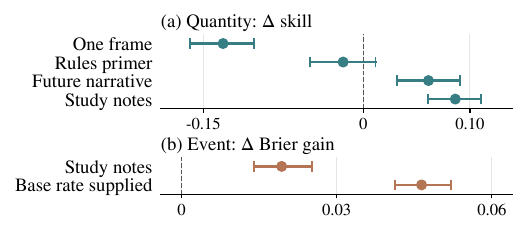}
\caption{Flash-Next's \omniID\ ablations with paired 95\% world-bootstrap intervals.
\Cref{tab:abl,app:study} report scores and protocols.}
\label{fig:information}
\end{minipage}\hfill
\begin{minipage}[t]{0.45\linewidth}
\vspace{0pt}
\centering
\includegraphics[width=\linewidth]{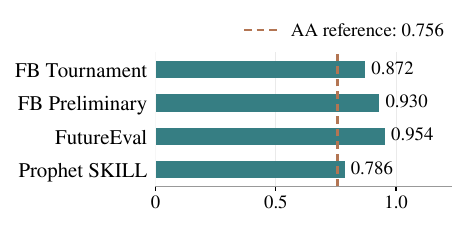}
\caption{Pearson correlations between SWFI and real-world forecasting scores.
\Cref{app:externalvalidity} provides details.}
\label{fig:externalvalidity}
\end{minipage}
\end{figure}

\textbf{Insight 1: Information controls and engine checks support validity.}
Re-seeded rollouts in \cref{app:repeatedfuture} retain uncertainty despite complete state
information, motivating distributional scoring.
\Cref{fig:information} shows that removing history reduces forecasting performance, whereas
study notes and privileged future narratives improve it, demonstrating evidence sensitivity.
Construction checks in \cref{app:validationchecks} detect no leakage or consistency
violations, and replay diagnostics in \cref{app:reloadchecks} justify factual-rerun controls.

\textbf{Insight 2: Reference lines locate the deficit in the models.}
\Cref{tab:main} shows that \expert\ reaches 51\% of \ceilingline, against 29\% for the best
model, so knowledge and harness support reveal substantial headroom under fixed observations, and
the headroom lies in the models rather than in noisy targets. Branch noise
lowers the Capture \ceilingline\ to $+0.39$ but cannot make a model negative, since predicting
no effect scores 0. Nine of eleven models nonetheless score below $-0.3$ in \omniID, while
\expert\ reaches $+0.13$ and study notes raise Flash-Next by $+0.24$ on paired cells in
\cref{app:study}. Models shift forecasts as much under placebo as under real edits in
\cref{tab:l3}, so their negative scores reflect over-reaction. For \TRT\ the limit is the
evidence: a privileged observer reaches only 0.59, so the task measures recovery of a weak but
real signal, on which \expert\ and Sol still separate from models at chance.

\textbf{Insight 3: Sandbox scores track real-world forecasting.}
\Cref{fig:externalvalidity} shows positive correlations between SWFI and forecasting scores on
ForecastBench \citep{karger2024forecastbench}, Metaculus FutureEval
\citep{metaculus2026futureeval}, and Prophet Arena \citep{prophet2026history}, with $r=0.79$--$0.95$ over panels of four to six matched models, and a model-level
permutation test pooling the panels gives $r=\pooledR$ with $p=\pooledP$. These exceed SWFI's correlation of $r=0.76$ with the AA general-capability index, and
\cref{tab:capabilitycorr} shows the belief components nearly uncorrelated with AA. SWFI thus tracks
forecasting-specific ability rather than general capability alone, which supports \bench\ as a
proxy for real-world forecasting capability.

\subsection{Where do current world models fall short? (RQ2)}
\label{sub:climb}

\textbf{Insight 1: Association forecasting strength does not reliably extend to belief inference and intervention prediction.}
\Cref{tab:main} shows that GLM-5.3 leads the model panel in quantity forecasting,
yet its trait score is below chance.
 Both GLM-5.3 and Kimi~K3 lag behind
Sol in belief inference and intervention direction. Such reversals recur across the panel, so association forecasting rankings do not
consistently reflect belief and intervention capabilities.

\begin{wrapfigure}[14]{r}{0.45\linewidth}
\vspace{-10pt}
\centering
\includegraphics[width=\linewidth]{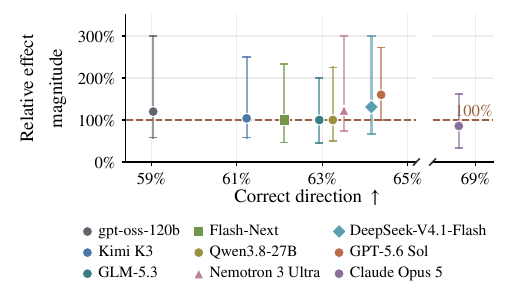}
\caption{Direction accuracy and effect magnitude for nine models in \omniID.
\Cref{app:directionmagnitude} adds the rest.}
\label{fig:intervention}
\end{wrapfigure}
The process analysis in \cref{sec:cot} traces this gap to intermediate states, which models
often mispredict even when their final answers are accurate and whose quality tracks
intervention performance. The usability study in \cref{sec:usability} further suggests that
agent performance is more closely associated with \Ltwo\ and \Lthree\ capabilities than
with \Lone\ forecasting alone, motivating evaluation beyond observation.

\textbf{Insight 2: Effect prediction remains a shared weakness across models.}
\Cref{tab:main} shows above-chance intervention direction for most models.
Yet nearly all models have negative capture across settings, below the zero-effect baseline.
\Cref{fig:intervention} shows informative direction predictions alongside widely dispersed
effect magnitudes. Among correct-direction predictions, some medians approach 100\%, while
interquartile ranges span underestimation and overestimation.
\Cref{app:directionmagnitude} reports coverage and conditional results across settings,
including \magnitudeOvershootRange\ of correct-direction predictions in \omniID, and up to
44\% in \omniOOD, exceeding twice the realized magnitude.
These errors arise from over-reaction to edits rather than from
target noise. A single global rescaling cannot correct errors that fall on both sides of the
truth, so effect magnitudes must be calibrated conditional on the world state.

\par

\section{Process Verification and Prediction Quality}
\label{sec:cot}
\newcommand{\cotAlignedAnchors}{190}
\newcommand{\cotSolAnchors}{30}
\newcommand{\cotMismatchRange}{4.7--9.5}
\newcommand{\cotEffectRhoRange}{0.85--0.94}
\newcommand{\cotProcessCapturePearson}{0.38}
\newcommand{\cotProcessCaptureSpearman}{0.54}
\newcommand{\cotProcessCaptureLow}{-0.27}
\newcommand{\cotProcessCaptureHigh}{+0.78}
\newcommand{\cotModelCapturePearson}{0.69}
\newcommand{\cotModelCaptureSpearman}{0.80}
\newcommand{\cotModelCaptureLow}{-0.15}
\newcommand{\cotModelCaptureHigh}{+1.00}
\newcommand{\cotSolFreeCapture}{-0.533}
\newcommand{\cotSolTableCapture}{-0.303}
\newcommand{\cotSolCodeCapture}{-0.240}
\newcommand{\cotSolCodeGain}{0.293}

Recorded trajectories and engine ground truth make intermediate
predictions verifiable, linking process quality to intervention prediction and
structured reasoning to process supervision (RQ3).

\subsection{Tasks and experimental settings}
We evaluate Qwen3.8-Flash-Next, Qwen3.8-27B, Gemma~4~31B, and Kimi~K3 on
\cotAlignedAnchors\ anchors, 10\% of the benchmark, and GPT-5.6~Sol on a \cotSolAnchors-anchor subset, each under three reasoning protocols.
\emph{Free} uses unrestricted reasoning without a prescribed process score.
\emph{Table} predicts prescribed intermediate states before answering, and \emph{Code}
generates a stochastic simulator, executed across 50 random seeds to yield intermediate
predictions and final-answer distributions. Both are scored against engine truth at the same prescribed intermediate cells, with process
quality defined as intermediate-state skill over the trend reference. \Cref{app:cot} details
the protocols.

\begin{figure}[!htbp]\centering
\includegraphics[width=\linewidth]{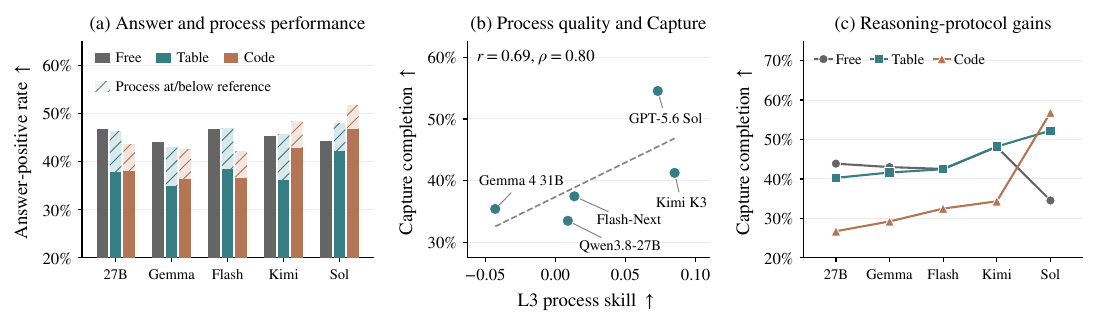}
\caption{Reasoning protocols and prediction quality. (a) Above-reference answers, with
hatching for intermediate forecasts at or below reference. (b) L3 process skill and Capture
completion, averaged over Table and Code. (c) Capture completion by protocol. Scores average
\omniID\ and \omniOOD\ equally, and \cref{app:cot} details scoring and coverage.}
\label{fig:cot}
\end{figure}

\subsection{How does process quality relate to prediction performance? (RQ3)}
\label{sub:cotfindings}

\textbf{Insight 1: Strong final-answer performance can accompany weak intermediate forecasts.}
\Cref{fig:cot}a shows that \cotMismatchRange\% of matched predictions exceed the final-answer
reference without exceeding the intermediate-state reference, across models and both
structured protocols. Final-answer evaluation therefore misses intermediate prediction
errors, motivating separate process verification.

\textbf{Insight 2: Process quality is positively associated with L3 intervention prediction performance.}
\Cref{fig:cot}b shows that higher L3 process quality accompanies greater Capture completion
across models. This association supports process verification as a complementary diagnostic
and intermediate-state forecasting as a route to stronger intervention prediction.
\Cref{app:cotmodelassociation} reports uncertainty.

\textbf{Insight 3: Structured reasoning can improve intervention prediction.}
\Cref{fig:cot}c shows Table gains over Free increasing with SWFI.
Sol benefits from both structured protocols and improves most with Code. Tables expose
intermediate states for verification, while code connects them through executable dynamics.
These results motivate world-model process supervision on states and intervention-induced
changes, for which \bench\ provides a multi-agent counterpart to the verified executable
worlds of \citet{wang2026codeasworlds}.

\section{World-Model Capabilities and Decision Utility}
\label{sec:usability}
We assess whether world-model forecasts and advice improve agent decisions and how these
gains relate to benchmark capability profiles (RQ4).

\subsection{Experimental setup and evaluation}
\label{sub:usability_setup}
A fixed Flash-Next agent plays turns 60--210 in 76 worlds, either without
world-model assistance or with forecasts and advice from Flash-Next or Kimi~K3 to inform
plan revisions.
A local study compares original, self-reviewed, and assisted plans at matched decision points
in 36 validation worlds, using Flash-Next, Kimi, Sol, and Opus as advisers. Each plan undergoes
four 20-turn executions without further strategic consultation. Both studies compare the
engine's civilization score within worlds, as \cref{app:agentplay} details.

\subsection{Do better world models support better decisions? (RQ4)}
\label{sub:usability_results}
\begin{figure}[!t]
\begin{minipage}[t]{0.48\linewidth}
\vspace{0pt}
\centering
\includegraphics[width=\linewidth]{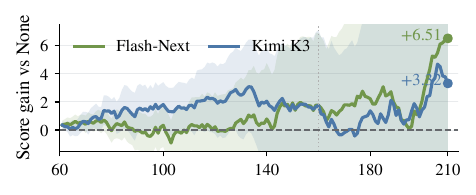}
\caption{Mean gains over no WM in 76 worlds with pointwise 95\% world-bootstrap intervals.}
\label{fig:ap_trajectories}
\end{minipage}\hfill
\begin{minipage}[t]{0.48\linewidth}
\vspace{0pt}
\centering
\captionof{table}{\omniID\ level completion and local gain over self-review.
Bold indicates the highest completion. \Cref{tab:ap_capability}
reports uncertainty.}
\label{tab:ap_capability_summary}
{\footnotesize
\setlength{\tabcolsep}{3pt}
\renewcommand{\arraystretch}{1.05}
\begin{tabular}{@{}lrrrr@{}}
\toprule
Model & \Lone & \Ltwo & \Lthree & $\Delta$ score \\
\midrule
Flash-Next & $24.1\%$ & $12.3\%$ & $19.7\%$ & $+0.19$ \\
Kimi K3 & $\mathbf{36.6}\%$ & $2.8\%$ & $26.7\%$ & $-0.66$ \\
Sol & $29.7\%$ & $\mathbf{26.0}\%$ & $19.0\%$ & $+1.40$ \\
Opus & $33.5\%$ & $19.7\%$ & $\mathbf{37.7}\%$ & $+1.28$ \\
\bottomrule
\end{tabular}
\par}
\end{minipage}
\end{figure}
\textbf{Insight 1: World models can support agent decision-making.}
\Cref{fig:ap_trajectories} traces score gains over turns 60--210, with positive mean gains for both assisted conditions at turn 210. In the local study, selecting plans on trial executions and evaluating them on held-out
executions, adding WM-generated plans improves selection by 2.36 score points over choosing
among original and self-reviewed plans only, so advice expands the set of useful alternatives.
The case studies show what useful advice consists of: Opus corrects the agent's reading of a
conditional loss probability as a certainty and redirects its remaining gold to one fortified
defender, saving two cities, whereas Sol's precautionary tax increase buys defenses against a
threat that never materializes and costs research. Useful advice therefore hinges on reading
the opponent's intentions and the state-specific cost of a precaution, which are belief and
intervention judgments rather than trend extrapolation.
\Cref{fig:ap_case_dialogue,fig:ap_bad_case_dialogue} illustrate how specific recommendations
change decisions.

\textbf{Insight 2: Decision-oriented evaluation should extend beyond association forecasting.}
\Cref{tab:ap_capability_summary} shows larger estimated decision gains for Opus and Sol than Kimi despite lower \Lone\ scores,
and these two advisers also lead the panel in belief inference and intervention direction.
Decision utility thus tracks the capabilities that association forecasting leaves unmeasured. This motivates evaluating hidden-state inference and action-conditioned effect prediction
alongside association forecasting.

\section{Conclusion}
\label{sec:conclusion}

\bench\ evaluates forecasting, hidden-state inference, and intervention prediction in
multi-agent worlds with engine-verifiable targets. Its composite score tracks real-world
forecasting more closely than general capability, while level diagnostics expose weaknesses
hidden by observational accuracy: intermediate-state errors, poorly calibrated effects, and
advice utility that forecasting alone does not explain. These findings prioritize state
representation, effect calibration, and verifiable process supervision. Taken together, the results characterize current language
models as competent trend extrapolators but weak causal simulators of multi-agent worlds, and
the benchmark supplies the hidden-state and intervention targets needed to train the latter.
\Cref{app:limitations} discusses limitations.

\subsection*{AI use statement}
Generative AI tools, including Claude and Codex, assisted implementation and manuscript
polishing, and were used for retrieval and discovery, such as finding related work and
checking bibliographic details, which the authors then verified against the original sources. Language models also produced the forecasts, reasoning, executable predictors,
and advice evaluated in the experiments. These outputs are distinct from the ground-truth
states and outcomes generated by the game engine. Expert-harness and mechanism-card uses
are documented with their experimental protocols. The authors are responsible for the
implementation, analyses, and manuscript writing.

\subsection*{Ethics statement}
The study involves no human participants. Benchmark trajectories use engine-controlled
agents, and the usability study introduces language-model-controlled decisions within the
same simulated environment. Freeciv is distributed under the GPL, whose requirements apply
to engine modifications. Benchmark performance does not validate policy recommendations
for human societies.
The icons used in our figures are sourced from Flaticon\footnote{\url{https://www.flaticon.com}}
and are free for personal and commercial purposes with attribution.

\subsection*{Reproducibility statement}
Scoring and analysis code, engine patches, and build instructions are available at
\url{https://github.com/Leon-Francis/CivWorldBench}. The benchmark data release, including frozen questions, targets, and model predictions,
is in preparation.
\Cref{app:repro} records evaluation configurations and result provenance.
\Cref{app:ceiling} specifies reference-line estimation, and \cref{app:corpus} documents the
collection pipeline and validation checks.

\bibliography{refs}
\bibliographystyle{unipat_report}
\clearpage
\appendix
\section{Related Work}
\label{app:related}
\label{sec:related}

\paragraph{World-model evaluation and intervention prediction.}
WorldPrediction evaluates action and action-sequence selection from video states
\citep{chen2025worldprediction}, while \citet{li2026twm} survey text world models for agents.
Corr2Cause tests causal inference from statistical descriptions \citep{jin2024corr2cause},
while CausaLab supports interactive causal discovery in synthetic laboratories
\citep{yang2026causalab}. These studies examine dynamics prediction and mechanism recovery
under different task formulations. ForecastBench-Sim \citep{lee2026forecastbenchsim} evaluates
binary event probabilities and continuous predictive distributions from fixed Freeciv reports.
Its intervention pilots score conditional forecasts against outcomes from edited savegames.
\bench\ uses observation histories to evaluate forecasting, hidden-state inference, and
intervention prediction within the same multi-agent dynamics. Recorded trait and rule values
serve as belief targets, and paired intervention and control rollouts provide targets for
effect direction and magnitude.

\paragraph{Real-world and social forecasting.}
ForecastBench evaluates forecasts of future events \citep{karger2024forecastbench}.
Prophet Arena separates event collection, prediction, and evaluation in a live forecasting
pipeline \citep{yang2025prophet}, while \citet{lu2025forecasting} compares model
forecasts with expert predictions on Metaculus questions. SocietyBench evaluates social-event
forecasting on anonymized, date-shifted timelines \citep{wang2026societybench}, and SWM-Bench uses prediction-market
trajectories to evaluate collective belief transitions \citep{yu2026swmbench}.
\bench\ complements these outcome-based evaluations with recorded hidden states and
repeated stochastic futures. \Cref{app:externalvalidity} relates its composite scores
to real-world forecasting through matched-model comparisons.

\paragraph{Multi-agent environments and decision utility.}
SOTOPIA evaluates social interaction \citep{zhou2024sotopia}, while social world models
explicitly represent other agents' mental states \citep{zhou2025swm}.
CivRealm and BALROG assess decision-making agents through game interaction
\citep{qi2024civrealm,paglieri2025balrog}. End-to-end performance combines dynamics knowledge,
planning, and action execution. \bench\ separately evaluates an observer of engine-controlled
civilizations, then holds the acting model fixed when comparing world-model advisers.
This design connects prediction diagnostics to decision utility while controlling
the agent's execution capabilities.

\paragraph{Process verification and executable worlds.}
\citet{rahmani2026debugging} distinguishes action-generation errors from state-propagation
errors in code world models. Verifier-driven executable models also support planning in
ARC-AGI-3 by checking simulated transitions against previous observations
\citep{rodionov2026executable}. Code as Worlds uses verified executable worlds to derive
physical-reasoning supervision \citep{wang2026codeasworlds}. \bench\ extends process
evaluation to stochastic multi-agent trajectories, checking Table and Code predictions
at shared intermediate states and relating their quality to final intervention forecasts.
These targets support studying supervision for both state prediction and intervention-induced
changes.

\section{Limitations}\label{app:limitations}
\paragraph{Scope.}
All dynamics come from one engine and its built-in AI. The generalization set changes the
ruleset, difficulty, map, civilization count, hidden-parameter ranges, and team structure, but
remains within that engine. The benchmark does not test transfer to other simulators or to
human societies.

\paragraph{References and targets.}
The Trait and Direction ceilings are computed on identifiable subsets, namely trait gaps of
at least 60 and cells with $|d_c|>\sigma_c$, so completion in these columns is conservative, and the Capture ceiling in \cref{app:referencelines} assumes Gaussian branch divergence.
Intervention targets include the reload-path divergence measured in \cref{app:reloadchecks}.
This divergence bounds attainable Capture but cannot produce negative scores, which
\cref{sec:results} attributes to the models.
Traits are only weakly identifiable from 80 turns of behavior, even for a privileged observer, whose accuracy is 0.59, so \TRT\ scores should be read against that observer. \expert\ is a language model
with a manual and tools, not a human baseline.

\paragraph{External and downstream evidence.}
The matched forecasting panels contain four to six models, and a general-capability index is
also correlated with the same leaderboards, so proxy validity rests on the component-level
evidence in \cref{tab:capabilitycorr} as much as on the composite. The process study covers
five models, and agent play uses one actor in benchmark worlds. Their model-level association
and full-game gain intervals include zero, as \cref{app:cot,app:agentplay} report.

\section{Formal definition}\label{app:formal}
\paragraph{World and observation.}
In the process $(\Sset,\mathcal A,\trans,\Omega,\obsfn{v})$ of \cref{sec:env}, a world
$w\sim\Wset$ specifies a ruleset, map, agent population, difficulty, hidden parameters, and
team layout. The action set $\mathcal A$ collects the engine commands available to an agent,
and $\Omega$ is the space of rendered reports. Determinism conditional on an engine random
stream does not imply that an edited reload consumes that stream identically to an unedited
run. The rendered policies $\{\pi_i\}$ describe current aims rather than future commitments.
\paragraph{Queries.} The world model $f(\hist{t_1},\{\pi_i\})$ answers four query types. With
$g$ a visible quantity of one agent, $e$ an event predicate on a trajectory segment, and $m$ an
operator on states, they are
\begin{align*}
\QNT:&\quad P\big(g(\st{t_1+h}) \mid \hist{t_1},\{\pi_i\}\big) \text{ reported as five quantiles,}\\
\EVT:&\quad P\big(e(\st{t_1:t_1+h}) = 1 \mid \hist{t_1},\{\pi_i\}\big) \text{ for binary items,}\\
\TRT,\RUL:&\quad P\big(\hid_k \mid \hist{t_1}\big) \text{ as a ranking or a bin, policies withheld,}\\
\INT:&\quad P\big(g(\st{t_1+h}') \mid \hist{t_1},\{\pi_i\},\ \dox(\st{t_1}\leftarrow m(\st{t_1}))\big),
\end{align*}
where $\st{}'$ follows $\trans$ from $m(\st{t_1})$ under the unchanged policies, and $\hid_k$
denotes a queried trait or world parameter in $\theta_w$. Mainline trajectories provide forecast targets, recorded
hidden fields provide belief targets, and rerun arms provide intervention targets.
Realized branch differences are not noise-free conditional average treatment effects.
Matched placebos measure response to negligible or mechanism-irrelevant edits, while
the model's own no-intervention forecast defines its predicted effect.
\paragraph{References.} A reference predictor is a fixed map $r(\hist{t_1})$. Persistence uses the current value, while the trend reference extrapolates
the two most recent frames. The conceptual ceiling knows $\trans$ and $\st{t_1}$ but not
future $\noise{}$. Re-seeded reruns approximate its forecasting loss.
The information sets of the ceiling and expert references are
\begin{align*}
\hat y_{\mathrm{ceil}} &\in \arg\min_{\hat y}
  \mathbb E[\ell(\hat y,Y)\mid \st{t_1},\trans],\\
\hat y_{\mathrm{expert}} &= h_{\mathcal H}(\hist{t_1};\mathcal K),
\end{align*}
where $\mathcal K$ denotes expert knowledge and $\mathcal H$ the harness. The ceiling has
complete state information, while the expert retains the model's rendered history and adds
knowledge and computation. The process evidence in \cref{sec:cot} consists of claims about
$g(\st{\tau})$, $t_1<\tau<t_1+h$, each scored against the trajectory.

\section{Corpus construction and engine patches}\label{app:corpus}
\subsection{World generation}
The evaluation corpus contains \nWorlds\ worlds with four nominal decision points per world.
The basic set fixes the ruleset to \texttt{civ2civ3} and difficulty to hard, with 75 worlds
in each of three map and population classes. The generalization set jointly varies ruleset,
difficulty, map, civilization count, hidden parameters, and team structure.
\Cref{tab:corpus} summarizes the map, population, and budget ranges.
Hidden-parameter ranges are wider than in the development pool. The resulting shift
does not require every OOD world to be harder than every ID world.

Mainline collection saves every turn and runs beyond the reporting budget to provide future
targets. The reporting budgets are 400, 800, and 1,200 turns. Decision points are distributed
from turn 60 to the reporting endpoint. Budget-specific forecast horizons and the final
post-filtering prompt inventory are defined separately in \Cref{app:questions}.

\subsection{Engine and ruleset modifications}
The benchmark uses Freeciv 3.1.6 with five source patches that make selected behavioral
parameters effective and permit changes in diplomatic relations. \Cref{tab:enginepatches}
summarizes their roles. Ruleset modifications widen trait ranges and enable domestic and
intercontinental trade. Illness variants alter generation but their suffixes are not shown
in model observations. The corpus and all intervention arms were collected with the resulting image, tagged \texttt{freeciv-server:3.1.6-v2}. The optional alliance-threshold patch is
unapplied, and the later task-persistence patch enters only the agent-play image of \cref{app:agentplay}, not the benchmark engine.

\begin{table}[t]\centering\small
\caption{Source patches used to construct the benchmark engine.}
\label{tab:enginepatches}
\begin{tabular}{@{}p{.26\linewidth}p{.68\linewidth}@{}}\toprule
Patch & Modification \\ \midrule
Trait initialization & Corrects trait arithmetic and initializes diplomatic contact at peace. \\
Diplomatic thresholds & Revises ceasefire, peace, and war decision thresholds. \\
Alliance dynamics & Reduces persistent alliance-bloc warfare. \\
Aggression response & Connects aggression to ceasefire acceptance and war persistence. \\
Production preference & Connects building preference to the active production decision. \\
\bottomrule\end{tabular}
\end{table}

Development checks tested diplomatic transitions, trait sensitivity, rule identifiability,
team-game completion, and intervention persistence before model evaluation. They establish
behavioral variation, while the final evidence gates define which inference questions are asked.

\subsection{Replacements and branch collection}
\label{app:branchcollection}
Three worlds were replaced after crashes or premature termination prevented target collection.
Only their replacements enter the final question set.

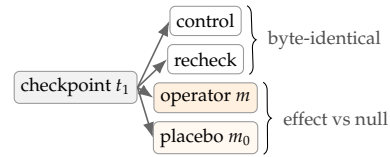
\begin{wrapfigure}{r}{0.46\linewidth}
\centering
\begin{tikzpicture}[font=\scriptsize, >=Latex, node distance=3pt and 12pt,
  n/.style={draw=black!40, rounded corners=2pt, inner sep=2.5pt, fill=white, align=center, minimum height=11pt},
  arm/.style={->, semithick, black!60}]
\node[n, fill=black!5] (ckpt) {checkpoint $t_1$};
\node[n, right=of ckpt, yshift=25pt] (ctrl) {control};
\node[n, below=of ctrl] (chk) {recheck};
\node[n, below=of chk, fill=orange!12] (real) {operator $m$};
\node[n, below=of real, fill=orange!5] (plac) {placebo $m_0$};
\foreach \a in {ctrl,chk,real,plac} \draw[arm] (ckpt.east) -- (\a.west);
\draw[decorate, decoration={brace, amplitude=3pt}] ($(ctrl.north east)+(2pt,0)$) -- ($(chk.south east)+(2pt,0)$)
  node[midway, right=4pt, align=left, text=black!70] {byte-identical};
\draw[decorate, decoration={brace, amplitude=3pt}] ($(real.north east)+(2pt,0)$) -- ($(plac.south east)+(2pt,0)$)
  node[midway, right=4pt, align=left, text=black!70] {effect vs null};
\end{tikzpicture}
\caption{Four branches from a shared checkpoint provide replay checks and intervention controls.}
\label{fig:arms}
\end{wrapfigure}

\Cref{fig:arms} shows the four branches collected from each eligible checkpoint.
Collection produces an unedited factual rerun, a factual
recheck, a real intervention, and a matched placebo. Branches are saved every turn.
Validation checks the applied edit, save continuity, and factual/recheck agreement over
comparable saved states. Successful short runs contribute only horizons with recorded truth.
Failed engine states are retried under recorded settings, including anchor-level reseeding
where needed, and their control arms are regenerated consistently.
The branch metadata records these deviations. No model prediction is used to select a retry.

Matching factual and recheck digests verifies replay, not shared randomness after an edit.
End-turn settings, save frequency, reload behavior, and interventions can change subsequent
random-number consumption. \Cref{app:reloadchecks} reports placebo and reseeding diagnostics.

\subsection{Integrity and provenance}
Each world and branch record identifies the engine configuration, initial state, intervention,
observation window, and available targets.
The question verifier checks scoped leakage and consistency constraints before scoring.
\Cref{app:validationchecks} reports these checks. Versioned checksums distinguish changes
to experimental inputs from changes to scoring or presentation.

\section{Question construction}\label{app:questions}
\subsection{Observation windows and question scopes}
Each question is associated with a world, decision turn, task, and observation condition.
With the frame spacing of \cref{sec:env}, an anchor at turn 60 uses turns 20, 40, and 60. A
narrative summarizes intervening events without including events after the anchor.
The egocentric viewer is the median-scoring living civilization at that decision point.
Its question scope includes itself and living civilizations it has contacted.

Belief questions omit the rendered policies because policy fields can reveal their hidden
targets, and truth fields remain outside the model-facing prompt.

\subsection{Numerical forecasts}
\QNT\ requests the 10th, 25th, 50th, 75th, and 90th percentiles of each eligible quantity.
Treasury, population, city count, total city size, military and civilian unit counts, and
enemy and ally counts are levels. Score, culture, and technology targets are increments
from the decision point. Repeated Future Technology advances do not increase the technology
count. Forecast horizons are 60/120, 120/250, and 200/300 turns for reporting budgets
of 400, 800, and 1,200 turns, respectively.

Unavailable engine targets are not imputed. Constant culture in classic-rule worlds and
technology increments after all technologies are known are excluded by the scorer's
eligibility rules. These exclusions are distinct from missing model answers, which receive
the penalties specified in \Cref{app:metrics}.

\subsection{Events and supplied base rates}
Binary templates describe occurrences within a horizon or comparisons between the initial
and final state. Leader items instead request a categorical distribution over civilizations.
The first filter retains binary template--budget--horizon--decision-turn cells with empirical
rates between 0.1 and 0.9. A second pass checks the reference cells actually used by each
evaluation setting. Cells with at least 200 items use the setting-specific empirical rate.
Sparse cells fall back to the corresponding corpus cell, which must also contain at least
200 items and satisfy the same rate filter.
The final reference is therefore determined before model scoring, not by its valid outputs.

Each retained binary question prints its reference rate. Leader items retain their original
categorical format and current-leader persistence reference. Intervention event prompts use
the same reference-cell machinery. These rates are in-sample benchmark statistics supplied
equally to all models, not estimates from an independent training set.

\subsection{Hidden traits and world rules}
Trait questions rank civilizations separately on aggression, expansion, trade, and building
preference. Each ranking includes every rankable civilization exactly once and permits
predicted ties. True ties are excluded from pairwise scoring. A scope with fewer than two
rankable civilizations has no trait-ranking question.

World-rule questions classify disaster frequency, growth cost, and research cost into
three bins. Disaster bins are at most 12, 13--40, and above 40. Growth- and research-cost
bins are at most 82, 83--117, and above 117. Lower costs correspond to faster growth or
research. Growth questions require at least one observed city-size change, and research
questions require at least two completed technologies within the observation window.
Egocentric evidence gates use the viewer's evidence rather than inaccessible foreign events.
Disaster questions have no additional event-count gate. Illness is not a benchmark target.

\subsection{Intervention prompts and eligibility}
An intervention prompt retains the history and describes the edit. Visible-state edits also
update the affected current observation. Numerical and event subquestions are separate
prompts, and a no-intervention control prompt is added when the required horizon or question
set differs from the ordinary forecast. Each intervention is paired with this explicit
control or the ordinary forecast, matching the target variables and horizons.

\Cref{tab:operators} lists the numerical scoring targets. Standard budget horizons are used
when they fall inside an operator's declared band. Otherwise the longest resolved arm
horizon in that band is selected. The available arm horizons are 30, 60, 90, 120, 150,
200, 250, and 300 turns. The target mask is fixed before inspecting a model's errors.

\begin{table}[t]\centering\small
\caption{Benchmark intervention families and numerical scoring targets. Bands are in turns.}
\label{tab:operators}
\begin{tabular}{@{}p{.26\linewidth}p{.55\linewidth}r@{}}\toprule
Intervention & Scored quantities & Band \\ \midrule
Treasury change & Treasury & 5--30 \\
Research points & Technologies learned & 5--60 \\
City destruction & Total city size, population & 5--150 \\
Government-enabling technology & Score increment & 5--60 \\
Map reveal & City count, score increment & 30--300 \\
Aggression trait & Enemy count, military units & 50--300 \\
Expansion trait & City count, total city size & 50--300 \\
Trade trait & Treasury, technologies learned & 50--300 \\
Building trait & Total city size, military units & 50--300 \\
Attitude change & Enemy and ally counts & 50--300 \\
Disaster frequency & Population, total city size & 30--300 \\
\bottomrule\end{tabular}
\end{table}

Real and placebo edits share their operator family but differ in magnitude or mechanism.
For example, city-destruction arms remove large versus small cities. City count is excluded
from their numerical target mask because both edits remove the same number of cities.
Likewise, the government-enabling and placebo technology grants each add one technology,
so their numerical target is a downstream score change rather than the identical count
increment. Four single-trait operators replace the earlier composite-trait edit.
Their collected modifiers are +90 for real edits and +3 for placebos. An accepted edit need
not produce a favorable or nonzero downstream effect in every state. Main capture uses
truth-moved cells, whereas specificity retains both moved and unmoved response cells.

\section{Metrics}\label{app:metrics}
\paragraph{Quantity aggregation.}
The headline \QNT\ score pools model and reference losses within each world and quantity
family before taking the ratio in \cref{eq:skill}. Families with zero reference loss are
omitted. Valid families are averaged within each world, followed by equal weighting of worlds.
Enemy and ally counts use the absolute error of the predicted median rather than the
five-quantile loss in \cref{eq:crps}. The appendix's family breakdown additionally
equalizes variables within each family before averaging worlds. It is a diagnostic and its
columns need not average to the headline \QNT\ score. The constants and variable transforms
are shared across compared conditions. Pooling reduces sensitivity to small per-cell
denominators but does not eliminate sensitivity when aggregate reference loss is small.

\paragraph{Events and hidden attributes.}
Event gain combines calibration and discrimination, without identifying them separately.
The template plot uses a Brier \emph{skill ratio} only as a diagnostic. \TRT\ averages pairwise ordering accuracy
over four traits. Predicted ties receive half credit and true ties are excluded. \RUL\
averages accuracies over the three rule parameters equally. Its majority reference and observer
reference use the same parameter weights, rather than weights proportional to question counts.

\paragraph{Baselines for the reference adjustment.}
The adjustment in \cref{sec:protocol} uses $b=0.5$ for trait and direction accuracy and, for
rules, the equal-parameter majority accuracy on the complete truth set, independent of a
model's missing or unparseable answers. \Cref{tab:rawaccuracies} reports the underlying
accuracies.
\begin{table}[t]\centering\small
\caption{Underlying accuracies before the main table's reference adjustment.
Triples are \omniID/\ego/\omniOOD. Raw references are 0.50 for trait and
direction, and the equal-parameter majority accuracy for rules.}
\label{tab:rawaccuracies}
\begin{tabular}{@{}lccc@{}}\toprule
Line & Trait accuracy & Rule accuracy & Direction accuracy \\ \midrule
Raw reference & 0.50 / 0.50 / 0.50 & 0.40 / 0.40 / 0.41 & 0.50 / 0.50 / 0.50 \\ \midrule
gpt-oss-120b & 0.50 / 0.49 / 0.52 & 0.43 / 0.44 / 0.46 & 0.57 / 0.56 / 0.54 \\
Gemma~4 31B & 0.52 / 0.51 / 0.54 & 0.47 / 0.46 / 0.46 & 0.64 / 0.63 / 0.59 \\
Nemotron~3 Ultra & 0.52 / 0.50 / 0.51 & 0.47 / 0.46 / 0.40 & 0.43 / 0.42 / 0.41 \\
Inkling & 0.53 / 0.51 / 0.55 & 0.47 / 0.48 / 0.44 & 0.54 / 0.56 / 0.48 \\
Qwen3.8-27B & 0.52 / 0.51 / 0.54 & 0.34 / 0.30 / 0.29 & 0.61 / 0.61 / 0.57 \\
GPT-5.6~Sol & 0.56 / 0.52 / 0.57 & 0.50 / 0.48 / 0.51 & 0.63 / 0.65 / 0.69 \\
DeepSeek-V4.1-Flash & 0.53 / 0.51 / 0.56 & 0.50 / 0.47 / 0.51 & 0.64 / 0.60 / 0.59 \\
Qwen3.8-Flash-Next & 0.52 / 0.51 / 0.54 & 0.45 / 0.47 / 0.39 & 0.60 / 0.62 / 0.58 \\
Kimi~K3 & 0.50 / 0.50 / 0.52 & 0.43 / 0.45 / 0.48 & 0.61 / 0.61 / 0.61 \\
GLM-5.3 & 0.50 / 0.50 / 0.49 & 0.44 / 0.46 / 0.43 & 0.57 / 0.60 / 0.55 \\
Claude Opus~5 & 0.54 / 0.51 / 0.56 & 0.48 / 0.48 / 0.51 & 0.65 / 0.72 / 0.59 \\
\expert & 0.56 / 0.53 / 0.57 & 0.70 / 0.61 / 0.71 & 0.76 / 0.79 / 0.65 \\
\bottomrule\end{tabular}
\end{table}

\paragraph{Missing or unparseable outputs.}
We retain required answer cells that a model omits or renders unparseably and count them as
failed predictions, using the same policy for every model. For \QNT, a missing cell
contributes twice its reference loss to the model-loss accumulator before the existing
family aggregation and clipping. A wholly missing family therefore receives skill $-1$.
The same rule uses the MAE reference for diplomacy. Families with zero reference loss remain
undefined. Missing binary event probabilities receive Brier loss 1, and missing leader
distributions receive multiclass Brier loss 2. Missing trait orderings and categorical labels
receive zero accuracy. These penalties do not impute a forecast.
For intervention capture, a missing arm or control prediction in cell $c$ contributes $2|d_c|$
to the effect-error numerator, and its direction is counted as incorrect. For specificity, let $r$
be the fraction of required response cells with valid arm and control predictions. We report
$r\,\mathrm{Spec}_{\mathrm{valid}}-(1-r)$, with score $-1$ when no responses are valid.
Here $\mathrm{Spec}_{\mathrm{valid}}$ is the valid-response selectivity score in \cref{eq:spec}.
An all-valid set retains the original statistic. Raw loss and calibration diagnostics still
require actual predictions. Incomplete experiments are reported at their available evaluation
scope. Entirely unrun anchors are not counted as failed model attempts.

\paragraph{Interventions.}
Capture uses truth-moved cells, $d_c\ne0$, of the target civilization within each
operator's declared variable set and horizon band, applying the failure convention above.
Zero predicted responses are excluded from the direction denominator. Correctness and
selectivity use different denominators, so a specificity of 1 can accompany inaccurate
real-arm predictions, and negative capture can accompany above-chance direction accuracy.
For reference ceiling $c_{\mathrm{Capture}}$, the completion percentage is
$100(C+1)/(c_{\mathrm{Capture}}+1)$, not the improvement over a no-effect prediction, which
attains $C=0$ inside this display interval.

\paragraph{Paired analyses and uncertainty.}
The paired analyses in \cref{sec:results} intersect valid cells within each comparison and retain
the original aggregation, truth, and reference losses. A cell is indexed by world, decision
point, operator and arm when relevant, civilization, target, and horizon. Repeated event items
with the same scored key retain their source-order occurrence index, assigned before validity
filtering. The ablation builders preserve this order. Paired cells must have identical ground
truth. Missing predictions are excluded jointly, so the paired estimate
is conditional on common valid outputs and is not a missingness-adjusted full-run estimate.
\QNT\ and \EVT\ intervals resample worlds jointly between conditions with 1,000 replicates and seed 2026. For capture, exponential world weights give a paired Bayesian bootstrap that keeps
the eligible operator set fixed. If any included operator has only one world, no between-world
interval is reported. These intervals do not quantify variation across repeated model calls,
and the exploratory comparisons are not adjusted for multiple testing.

\section{Validation, reference lines, and evaluation noise}\label{app:ceiling}
\subsection{Repeated-future reference protocol}
\label{app:repeatedfuture}
We sample \nCeilAnchors\ decision points stratified by evaluation set and turn budget.
Each contributes \nCeilReruns\ re-seeded reruns saved every turn, combined with a historical
factual rerun to form the nine-member reference ensemble.
The mainline's rank among the nine trajectories has mean probability integral transform
\ceilPIT, compared with 0.5 under exchangeability. Fair CRPS and fair Brier losses correct
for finite ensemble size. We report comparisons against mainline truth, leave-one-out
comparisons against rerun truth, and an ensemble evaluated with the benchmark scorer.
Leave-one-out estimates are the primary event reference because reload behavior can change
the event distribution relative to the uninterrupted mainline.

Reload triggers AI government re-evaluation, changing the government-event distribution.
Ten-turn saves also miss brief treasury-to-zero events, shifting their reference from
\ceilBankruptTen\ to \ceilBankruptPT\ with per-turn recording. The historical factual member
retains ten-turn evaluation, leaving a residual resolution mismatch. These template-specific
distribution shifts remain despite the near-central aggregate rank. In \ego, the full-checkpoint
observer remains a privileged reference rather than the Bayes limit of the rendered history.
\subsection{Reference definitions and interpretation}
\label{app:referencelines}
\begin{table}[t]\centering\small
\caption{Validation evidence and reference assumptions. Triples denote
\omniID/\ego/\omniOOD. Numerical references differ in information and estimation procedure.}
\label{tab:validation}
\begin{tabular}{@{}p{.18\linewidth}p{.25\linewidth}p{.50\linewidth}@{}}\toprule
Check & Observation & Interpretation \\ \midrule

Repeated futures & \QNT\ $\simeq+0.68$, \EVT\ $\simeq+0.11$ &
Predictable signal remains under engine randomness with full-checkpoint information. \\
Trait observer & 0.59 on all pairs, 0.63 at gap $\ge$40, 0.66 at gap $\ge$60 &
Privileged behavioral features provide partial recovery of trait orderings. \\
Rule observer & \ceilRules &
Formula inversion and disaster statistics support parameter-specific identifiability. \\
Direction reference & all moved cells \ceilDirAll &
The displayed 0.92 concerns only effects above estimated placebo divergence. \\
Capture reference & \ceilCapture &
A reference derived under a Gaussian noise decomposition. \\
Specificity endpoint & 1.00 &
An algebraic endpoint for real-only responses. \\
Expert reference & \QNT\ \humOneA &
A positive control on \nHumanAnchors\ anchors with additional knowledge and harness support. \\

\bottomrule\end{tabular}
\end{table}

\subsubsection{Information sets of the numerical ceiling}
Repeated futures approximate the forecast loss of the conceptual \ceilingline, while the
other columns use privileged observers, noise models, or an algebraic endpoint. Trait and
rule observers infer labels from privileged behavioral evidence with target fields withheld,
so their scores characterize empirical recoverability rather than proven information limits.

\subsubsection{Trait and rule references}
The trait observer is leave-world-out ridge regression on 33 privileged behavior statistics
from the 80 turns preceding the decision point. Its accuracy ranges from \ceilTraitsRange\
across traits, averaging \ceilTraitsAll\ over all unequal pairs. Restricting true trait gaps
to at least 40 raises the four-trait mean to \ceilTraitsGapForty, with a per-trait range of 0.59--0.66, and gaps of at least 60 raise it to 0.66. The displayed 0.66 is this
well-separated-pair reference, whereas the model scores include all eligible unequal pairs. The accuracies in this paragraph precede the
main table's reference adjustment. The improvement with larger trait gaps indicates that
behavioral evidence is more informative for strongly separated dispositions.
The rule observer inverts growth and research-cost
formulas where evidence is available and uses a leave-world-out classifier on disaster counts
per city-turn for the remaining parameter. Disaster accuracy is \ceilDisasters. Averaging
the three parameter references equally yields \ceilRules.

\subsubsection{Intervention noise model}
Let $d_c$ be the observed real-arm effect and $\sigma_c$ the root-mean-square placebo effect
for the corresponding operator and quantity. The direction reference uses
$\Phi(\sqrt{\max(d_c^2-\sigma_c^2,0)}/\sigma_c)$, where $\Phi$ is the standard normal
cumulative distribution function. It is \ceilDirAll\ over all moved cells and about
\ceilDirIdentifiable\ on the \fracDirIdentifiable\ of moved cells with $|d_c|>\sigma_c$, whereas models are scored on all moved cells.
The capture reference assumes centered Gaussian divergence and exact prediction of the
underlying mechanism, so expected absolute residual error is $\sigma_c\sqrt{2/\pi}$.
Substituting this into the operator-level capture ratio gives
$1-\sum_c\sigma_c\sqrt{2/\pi}/\sum_c|d_c|$, then the operator macro-average
\ceilCapture. Cell pooling gives \ceilCapturePooled, illustrating the effect of operator
weighting. These estimates assume Gaussian divergence and transferable placebo variance.
Completion expresses progress along the resulting reference scale.

\subsubsection{Construction of the expert reference}
\label{app:expertprotocol}
The published \expert\ reference combines GPT-5.6~Sol with the frozen v3.3 manual and a
tool-assisted harness on the shared \nApiAnchors-anchor sample.

\paragraph{Additional expert knowledge.}
The v3.3 manual summarizes game rules, default-AI logic, engine patches, and empirical
regularities from non-benchmark worlds. It covers growth and research formulas, behavioral
cues for hidden targets, forecast intervals, and intervention reasoning, without target-world
answers or model scores.

\paragraph{Harness and information boundary.}
An independent agent handles each anchor--condition pair using the manual, canonical
observations, task prompts, and a fixed answer inventory. It maintains working notes,
uses Python for calculations, and answers sequentially to coordinate factual and intervention
predictions. Access is restricted to this input pack, excluding external data, prior
predictions, scores, hidden state, and future outcomes. The harness adds computation and
multi-step reasoning without changing the evaluated world's information boundary.

\paragraph{Validation and interpretation.}
\Cref{tab:main_full} reports the combined effect of expert knowledge, tools, and reasoning
resources under the same observations. Evaluation uses the fixed v3.3 manual and input packs,
with answer inventories checked before scoring and standard penalties for missing or
unparseable cells. Later manual revisions and expanded-information pilots are excluded.

\subsection{Association with real-world forecasting performance}
\label{app:externalvalidity}

\begin{table}[t]\centering\small
\caption{Associations between fixed-weight SWFI and external model scores.
$n$ counts matched models. Ranges span eligible external configurations.
The two ForecastBench tracks overlap. AA provides a general-capability reference.
\Cref{app:externalvalidity} describes the protocols and sensitivity analyses.}
\label{tab:externalvalidity}
\begin{tabular}{@{}lrrr@{}}\toprule
External reference & $n$ & Pearson $r$ & Spearman $\rho$ \\ \midrule
ForecastBench Tournament (Overall) & 4 & 0.864--0.879 & 0.800 \\
ForecastBench Preliminary (Dataset) & 5 & 0.930 & 0.900 \\
Metaculus FutureEval (Avg Score) & 5 & 0.954 & 0.975 \\
Prophet Fixed Context (SKILL) & 6 & 0.786 & 0.886 \\
\midrule
AA Intelligence Index v4.3 (reference) & 11 & 0.756 & 0.795 \\
\bottomrule\end{tabular}
\end{table}

\paragraph{External associations.}
\Cref{tab:externalvalidity} shows positive Pearson and Spearman correlations between SWFI
and matched-model forecasting scores across all reported tracks. Agreement in both score
variation and ranking supports the use of \bench\ as a proxy for real-world forecasting
capability.

\paragraph{Evaluation scope.}
All comparisons use the same scores, missing-answer penalties, and reference scales as
\cref{tab:main,tab:main_full}. The analysis uses 1,900 anchors for open models
and 190 for Opus~5 and GPT-5.6~Sol. Gemini~3.7 Flash, GPT-6 Astra, GLM-5.3 Flash, and
Hunyuan Hy3 add evaluations on the same 60 anchors under the same scoring policy.
Both subsets are stratified to match the full benchmark's composition, as described in
\cref{sec:protocol}, so the three sample sizes estimate the same quantities at different precision. As a check,
SWFI recomputed on the 60 anchors is 27.5 for Sol against 27.7 on its 190 anchors, and 25.1
for Kimi~K3 against 21.9 on the full grid. Neither difference changes either model's rank
among the matched models on any panel.
The assisted expert is excluded as a non-independent model.

\paragraph{Expansion and matching protocol.}
Weights and scales were fixed before evaluating the four additional models, selected
for external overlap and API access. Leaderboard eligibility was refined during exploratory
analysis after inspecting existing results. \Cref{tab:externalmatches} lists exact model
identities, local scores, and eligible matches. Each model contributes one observation,
with multiple external configurations handled as described below.

\begin{table}[t]\centering\small
\caption{SWFI and external scores for exact model matches. T/P denotes ForecastBench
Tournament/Preliminary and FE denotes FutureEval. SKILL uses Prophet's Fixed Context track.
Dashes indicate unavailable matches. External scores retain their native units.}
\label{tab:externalmatches}
\setlength{\tabcolsep}{4pt}
\begin{tabular}{@{}lrrrrrr@{}}\toprule
Model & SWFI & FB T & FB P & FE & SKILL & AA \\ \midrule
GPT-6 Astra & 36.59 & -- & -- & -- & -0.98 & -- \\
Claude Opus~5 & 29.47 & 66.10--66.30 & 61.70 & 15.78 & -- & 45 \\
Gemini 3.7 Flash & 28.75 & -- & 62.00 & -- & -0.91 & -- \\
GPT-5.6~Sol & 27.68 & 65.40--65.50 & 61.50 & 12.93 & -1.05 & 39 \\
GLM-5.3 Flash & 22.57 & -- & -- & -- & -- & -- \\
Kimi~K3 & 21.94 & 65.60 & 60.90 & 12.93 & -1.14 & 44 \\
DeepSeek-V4.1-Flash & 20.48 & -- & -- & -- & -- & 40 \\
GLM-5.3 & 19.19 & -- & -- & -- & -1.08 & 45 \\
Qwen3.8-Flash-Next & 16.93 & -- & -- & -- & -- & 40 \\
Gemma~4 31B & 16.52 & 61.50 & 57.40 & -- & -- & 15 \\
Hunyuan Hy3 & 15.83 & -- & -- & -- & -1.17 & -- \\
Inkling & 15.73 & -- & -- & -- & -- & 26 \\
Qwen3.8-27B & 12.47 & -- & -- & -- & -- & 34 \\
Nemotron~3 Ultra & -0.98 & -- & -- & 5.83 & -- & 23 \\
gpt-oss-120b & -3.89 & -- & -- & -0.26 & -- & 12 \\
\bottomrule\end{tabular}
\end{table}

\paragraph{External sources and configurations.}
ForecastBench \citep{karger2024forecastbench,forecastbench2026leaderboard} uses the
September~22, 2026 organizer tables for Tournament Overall and Preliminary Dataset.
Preliminary remains provisional, and the two tracks overlap.
Eligible rows retain their search/code and reasoning configurations. For multiple eligible
variants, \cref{tab:externalvalidity} reports all correlation combinations. The main-text
figure correlates each model's median external score across configurations.

Metaculus FutureEval \citep{metaculus2026futureeval} uses the archived leaderboard PDF
dated September~23, 2026 and its log-score-based Avg Score. All five matches use AskNews
variants, although Opus and GPT reasoning levels differ from the local medium settings.
Matching distinguishes Gemma~26B from Gemma~31B.

Prophet Arena \citep{prophet2026history} uses the Fixed Context SKILL snapshot retrieved
September~23, 2026 at 16:17 China Standard Time, with six exact model matches. SKILL is
the fitted market-anchored parameter $\theta$. Retired models retain their last scores.
Evidence and inference budgets can differ within the shared track.
FutureX is excluded because a common-framework comparison could not be established.

\paragraph{Sensitivity and uncertainty.}
Pearson correlation measures linear association, while Spearman correlation measures rank
agreement. Each comparison uses models with observed scores on both benchmarks.

\begin{table}[t]\centering\small
\caption{External-validity sensitivity. Drop-one ranges omit each model in turn and include all
eligible configurations. Exact two-sided permutation tests are exploratory and unadjusted.}
\label{tab:externalsensitivity}
\setlength{\tabcolsep}{4pt}
\begin{tabular}{@{}lrr@{}}\toprule
External reference & Drop-one $r$ & Permutation $p$ \\ \midrule
ForecastBench Tournament (Overall) & 0.455--1.000 & 0.0833--0.1250 \\
ForecastBench Preliminary (Dataset) & 0.917--0.994 & 0.0250 \\
Metaculus FutureEval (Avg Score) & 0.937--0.987 & 0.0167 \\
Prophet Fixed Context (SKILL) & 0.673--0.913 & 0.0667 \\
\bottomrule\end{tabular}
\end{table}

\begin{table}[t]\centering\small
\caption{Same-model comparison with AA: each predictor is correlated with the external outcome.
This differs from the SWFI--AA association in \cref{fig:externalvalidity}.
Only panels with at least three recorded AA values are included.}
\label{tab:externalAA}
\begin{tabular}{@{}lrrr@{}}\toprule
External reference & $n$ & SWFI $r$ & AA $r$ \\ \midrule
ForecastBench Tournament (Overall) & 4 & 0.864--0.879 & 0.991--0.994 \\
ForecastBench Preliminary (Dataset) & 4 & 0.917 & 0.964 \\
Metaculus FutureEval (Avg Score) & 5 & 0.954 & 0.987 \\
Prophet Fixed Context (SKILL) & 3 & 0.510 & -0.645 \\
\bottomrule\end{tabular}
\end{table}

\Cref{tab:externalsensitivity} shows that Preliminary and FutureEval retain strong
correlations after each model is removed, while Tournament is more sensitive to model
removal. The reported ranges cover leave-one-model-out estimates and eligible configurations.
Exact two-sided permutation tests are exploratory and unadjusted for multiple comparisons.
With four matched models the smallest attainable two-sided $p$ is $0.083$, so the Tournament
panel cannot reach significance at any correlation. The pooled test in \cref{sec:results}
standardizes each panel's external scores, pools the \pooledPairs\ model--panel pairs over
\pooledModels\ distinct models, and permutes SWFI across models so that a model's pairs move
together, with 200,000 draws. It gives $r=\pooledR$ and $p=\pooledP$, and $r=\pooledRnoT$
with $p=\pooledPnoT$ after dropping the Tournament panel, which overlaps Preliminary. The
SWFI--AA correlation over eleven models has permutation $p=\aaPermP$.
These analyses condition on observed scores and model panels, leaving external score
uncertainty, decoding variation, and model-family dependence outside their uncertainty estimates.

The AA Intelligence Index v4.3 \citep{artificialanalysis2026index} provides a
general-capability reference for the eleven main-table models, excluding the assisted expert.
\Cref{tab:capabilitycorr} reports its association with each metric, where the forecasting and
Capture columns track the index and the belief columns do not. \Cref{tab:externalAA}
separately compares SWFI and AA against the same forecasting outcomes on matched models.
\begin{table}[t]\centering\small
\caption{Spearman rank correlations between the external capability index and
the unrounded main-table scores across 11 unassisted model lines in each setting.
All seven metrics are included; tied values receive average ranks. The assisted
\expert\ line is excluded. These are descriptive model-level associations, not
matched-anchor comparisons or significance tests.}
\label{tab:capabilitycorr}
\begin{tabular}{@{}lrrr@{}}\toprule
Metric & \omniID & \ego & \omniOOD \\ \midrule
\QNT & $+0.84$ & $+0.67$ & $+0.92$ \\
\EVT & $+0.75$ & $+0.86$ & $+0.77$ \\
\TRT & $+0.07$ & $-0.03$ & $+0.11$ \\
\RUL & $+0.15$ & $+0.31$ & $+0.23$ \\
Capture & $+0.76$ & $+0.75$ & $+0.58$ \\
Direction & $+0.34$ & $+0.39$ & $+0.34$ \\
Specificity & $+0.15$ & $+0.55$ & $+0.65$ \\
\bottomrule\end{tabular}
\end{table}

\subsection{Evidence sensitivity and construction checks}\label{app:validationchecks}
Question targets are computed from recorded engine states and trajectories.
Automated validation covers 1,900 decision points and 22,763 prompts, including 4,264
intervention-event prompts excluded from the reported task inventory. No violations were
detected in the tested constraints. The checks cover hidden-key
and future-turn leakage, observer scope, event-cell eligibility, hidden-rule evidence gates,
intervention horizon bands and arm pairing, and the task-specific presence of policy
summaries.
\Cref{app:study} reports the paired evidence-sensitivity ablations.

\subsection{Noise sources and reload checks}
\label{app:reloadchecks}
Uncertainty arises from future engine randomness, ambiguity in partially observed state,
reload and temporal-resolution effects, and finite evaluation samples. Repeated futures
address the first source, privileged observers characterize the second under their stated
information sets, and replay checks address the third. World bootstraps estimate sampling
uncertainty across worlds conditional on the recorded model outputs.

All intervention truths use rerun factual arms from the same saved checkpoint rather than the
uninterrupted mainline. Saves are taken every turn so short-lived events are not missed.
The RNG-sharing diagnostic compares 141,250 placebo cells with 11,091 re-seeded cells:
mean normalized absolute divergence is 0.116 versus 0.259, and 56.1\% of placebo cells have
zero divergence. Only 3,752 of 28,968 compared saved turns, or 13.0\%, have identical digests.
The low digest agreement shows that edited and unedited runs can follow different random
streams. The divergence means describe different cell populations, so the comparison is
descriptive. Placebo rollouts characterize variation introduced along the edit and reload
path. Intervention errors therefore reflect both effect-estimation error and stochastic
branch divergence.

\section{Full results}\label{app:results}
\subsection{Complete three-setting results}
\Cref{tab:main_full} reports all seven metrics in \omniID, \ego, and \omniOOD,
using the score sources and references of \cref{tab:main}. SWFI averages the three
settings. The expert uses the shared \nApiAnchors-anchor sample and the knowledge and
harness specified in \cref{app:expertprotocol}, with the same failure penalties as other models.
\begin{table}[t]\centering\scriptsize
\setlength{\tabcolsep}{1.5pt}\renewcommand{\arraystretch}{1.12}
\caption{Complete results across three settings. Bold denotes the best model score, excluding the expert reference. Percentages report reference-normalized completion. Missing or unparseable answers receive the failure penalties in \cref{app:metrics}. \Cref{app:referencelines} specifies the numerical reference assumptions.}
\label{tab:main_full}
\begin{tabular}{@{}p{0.195\linewidth}>{\centering\arraybackslash}p{0.045\linewidth}*{6}{>{\raggedleft\arraybackslash}p{\dimexpr(0.76\linewidth-27pt)/7\relax}}>{\raggedleft\arraybackslash}p{\dimexpr(0.76\linewidth-27pt)/7+3pt\relax}@{}}
\toprule
 & & \multicolumn{2}{c}{\Lone\ association} & \multicolumn{2}{c}{\Ltwo\ belief} & \multicolumn{3}{c}{\Lthree\ intervention} \\
\cmidrule(lr){3-4}\cmidrule(lr){5-6}\cmidrule(lr){7-9}
Line & \shortstack{AA\\Index} & \QNT$\uparrow$ & \EVT$\uparrow$ & \TRT$\uparrow$ & \RUL$\uparrow$ & \textsc{Capture}$\uparrow$ & \textsc{Direction}$\uparrow$ & \textsc{Specificity}$\uparrow$ \\
\midrule
\rowcolor{tableTeal!12} \multicolumn{9}{@{}l}{\textcolor{tableTeal}{\omniID}} \\
\rowcolor{black!3} \floorline & \textemdash & +0.00 & +0.000 & 0.00 & 0.00 & $-$1.00 & 0.00 & +0.00 \\
\tabledash
gpt-oss-120b & 12 & +0.03\headroompct{tableTeal}{4} & $-$0.031\headroompct{tableRose}{-29} & 0.01\headroompct{tableTeal}{2} & 0.04\headroompct{tableTeal}{5} & $-$0.65\headroompct{tableTeal}{25} & 0.15\headroompct{tableTeal}{16} & +0.05\headroompct{tableTeal}{5} \\
\rowcolor{black!2}
Gemma~4 31B & 15 & +0.21\headroompct{tableTeal}{31} & +0.012\headroompct{tableTeal}{11} & 0.08\headroompct{tableTeal}{12} & 0.11\headroompct{tableTeal}{12} & $-$0.61\headroompct{tableTeal}{28} & 0.30\headroompct{tableTeal}{33} & \textbf{+0.19}\headroompct{tableTeal}{19} \\
Nemotron~3 Ultra & 23 & +0.08\headroompct{tableTeal}{12} & +0.003\headroompct{tableTeal}{3} & 0.08\headroompct{tableTeal}{11} & 0.12\headroompct{tableTeal}{13} & $-$0.63\headroompct{tableTeal}{27} & $-$0.16\headroompct{tableRose}{-17} & $-$0.33\headroompct{tableRose}{-33} \\
\rowcolor{black!2}
Inkling & 26 & +0.22\headroompct{tableTeal}{32} & +0.012\headroompct{tableTeal}{11} & 0.13\headroompct{tableTeal}{20} & 0.11\headroompct{tableTeal}{12} & $-$0.31\headroompct{tableTeal}{49} & 0.09\headroompct{tableTeal}{10} & $-$0.11\headroompct{tableRose}{-11} \\
Qwen3.8-27B & 34 & +0.25\headroompct{tableTeal}{36} & +0.007\headroompct{tableTeal}{7} & 0.10\headroompct{tableTeal}{16} & $-$0.11\headroompct{tableRose}{-12} & $-$0.34\headroompct{tableTeal}{48} & 0.24\headroompct{tableTeal}{26} & $-$0.03\headroompct{tableRose}{-3} \\
\rowcolor{black!2}
GPT-5.6~Sol & 39 & +0.27\headroompct{tableTeal}{39} & +0.022\headroompct{tableTeal}{20} & \textbf{0.23}\headroompct{tableTeal}{34} & 0.16\headroompct{tableTeal}{17} & $-$0.60\headroompct{tableTeal}{29} & 0.28\headroompct{tableTeal}{30} & $-$0.02\headroompct{tableRose}{-2} \\
DeepSeek-V4.1-Flash & 40 & +0.19\headroompct{tableTeal}{27} & +0.017\headroompct{tableTeal}{16} & 0.14\headroompct{tableTeal}{22} & \textbf{0.17}\headroompct{tableTeal}{18} & $-$0.58\headroompct{tableTeal}{31} & 0.30\headroompct{tableTeal}{32} & +0.07\headroompct{tableTeal}{7} \\
\rowcolor{black!2}
Qwen3.8-Flash-Next & 40 & +0.26\headroompct{tableTeal}{38} & +0.011\headroompct{tableTeal}{10} & 0.10\headroompct{tableTeal}{15} & 0.08\headroompct{tableTeal}{9} & $-$0.57\headroompct{tableTeal}{31} & 0.22\headroompct{tableTeal}{24} & +0.05\headroompct{tableTeal}{5} \\
Kimi~K3 & 44 & +0.34\headroompct{tableTeal}{50} & \textbf{+0.025}\headroompct{tableTeal}{23} & $-$0.00\headroompct{tableRose}{-0} & 0.05\headroompct{tableTeal}{6} & $-$0.35\headroompct{tableTeal}{47} & 0.23\headroompct{tableTeal}{25} & +0.08\headroompct{tableTeal}{8} \\
\rowcolor{black!2}
GLM-5.3 & 45 & \textbf{+0.37}\headroompct{tableTeal}{53} & +0.023\headroompct{tableTeal}{21} & $-$0.01\headroompct{tableRose}{-2} & 0.07\headroompct{tableTeal}{8} & $-$0.25\headroompct{tableTeal}{54} & 0.16\headroompct{tableTeal}{17} & $-$0.05\headroompct{tableRose}{-5} \\
Claude Opus~5 & 45 & +0.33\headroompct{tableTeal}{48} & +0.021\headroompct{tableTeal}{19} & 0.17\headroompct{tableTeal}{26} & 0.13\headroompct{tableTeal}{14} & \textbf{$-$0.08}\headroompct{tableTeal}{66} & \textbf{0.33}\headroompct{tableTeal}{35} & +0.12\headroompct{tableTeal}{12} \\
\tabledash
\rowcolor{tableTeal!7}
\expert & \textemdash & +0.37\headroompct{tableTeal}{55} & +0.021\headroompct{tableTeal}{20} & 0.26\headroompct{tableTeal}{39} & 0.52\headroompct{tableTeal}{56} & +0.13\headroompct{tableTeal}{81} & 0.58\headroompct{tableTeal}{63} & +0.86\headroompct{tableTeal}{86} \\
\tabledash
\rowcolor{tableTeal!4} \ceilingline & \textemdash & +0.68 & +0.108 & 0.66 & 0.93 & +0.39 & 0.92 & 1.00 \\
\addlinespace[2pt]\midrule
\rowcolor{tableTeal!12} \multicolumn{9}{@{}l}{\textcolor{tableTeal}{\ego}} \\
\rowcolor{black!3} \floorline & \textemdash & +0.00 & +0.000 & 0.00 & 0.00 & $-$1.00 & 0.00 & +0.00 \\
\tabledash
gpt-oss-120b & 12 & $-$0.20\headroompct{tableRose}{-29} & $-$0.035\headroompct{tableRose}{-32} & $-$0.02\headroompct{tableRose}{-4} & 0.06\headroompct{tableTeal}{7} & $-$0.73\headroompct{tableTeal}{20} & 0.14\headroompct{tableTeal}{15} & $-$0.82\headroompct{tableRose}{-82} \\
\rowcolor{black!2}
Gemma~4 31B & 15 & +0.05\headroompct{tableTeal}{7} & +0.003\headroompct{tableTeal}{3} & 0.06\headroompct{tableTeal}{9} & 0.11\headroompct{tableTeal}{12} & $-$0.51\headroompct{tableTeal}{35} & 0.28\headroompct{tableTeal}{30} & +0.08\headroompct{tableTeal}{8} \\
Nemotron~3 Ultra & 23 & $-$0.16\headroompct{tableRose}{-24} & +0.006\headroompct{tableTeal}{6} & $-$0.02\headroompct{tableRose}{-3} & 0.11\headroompct{tableTeal}{12} & $-$0.68\headroompct{tableTeal}{23} & $-$0.17\headroompct{tableRose}{-19} & $-$0.39\headroompct{tableRose}{-39} \\
\rowcolor{black!2}
Inkling & 26 & +0.07\headroompct{tableTeal}{10} & +0.008\headroompct{tableTeal}{8} & 0.04\headroompct{tableTeal}{6} & 0.13\headroompct{tableTeal}{16} & $-$0.31\headroompct{tableTeal}{50} & 0.12\headroompct{tableTeal}{14} & $-$0.03\headroompct{tableRose}{-3} \\
Qwen3.8-27B & 34 & +0.06\headroompct{tableTeal}{9} & +0.006\headroompct{tableTeal}{5} & 0.04\headroompct{tableTeal}{5} & $-$0.20\headroompct{tableRose}{-23} & $-$0.33\headroompct{tableTeal}{48} & 0.24\headroompct{tableTeal}{26} & +0.01\headroompct{tableTeal}{1} \\
\rowcolor{black!2}
GPT-5.6~Sol & 39 & +0.16\headroompct{tableTeal}{23} & +0.019\headroompct{tableTeal}{17} & \textbf{0.08}\headroompct{tableTeal}{12} & \textbf{0.15}\headroompct{tableTeal}{18} & $-$0.27\headroompct{tableTeal}{53} & 0.32\headroompct{tableTeal}{35} & \textbf{+0.16}\headroompct{tableTeal}{16} \\
DeepSeek-V4.1-Flash & 40 & +0.04\headroompct{tableTeal}{6} & +0.014\headroompct{tableTeal}{13} & 0.03\headroompct{tableTeal}{5} & 0.12\headroompct{tableTeal}{14} & $-$0.69\headroompct{tableTeal}{23} & 0.22\headroompct{tableTeal}{24} & +0.12\headroompct{tableTeal}{12} \\
\rowcolor{black!2}
Qwen3.8-Flash-Next & 40 & +0.05\headroompct{tableTeal}{7} & +0.009\headroompct{tableTeal}{9} & 0.02\headroompct{tableTeal}{3} & 0.13\headroompct{tableTeal}{15} & $-$0.39\headroompct{tableTeal}{44} & 0.26\headroompct{tableTeal}{28} & +0.09\headroompct{tableTeal}{9} \\
Kimi~K3 & 44 & \textbf{+0.22}\headroompct{tableTeal}{32} & +0.018\headroompct{tableTeal}{17} & 0.01\headroompct{tableTeal}{2} & 0.09\headroompct{tableTeal}{11} & $-$0.24\headroompct{tableTeal}{54} & 0.24\headroompct{tableTeal}{26} & +0.03\headroompct{tableTeal}{3} \\
\rowcolor{black!2}
GLM-5.3 & 45 & +0.12\headroompct{tableTeal}{18} & +0.016\headroompct{tableTeal}{15} & $-$0.02\headroompct{tableRose}{-2} & 0.11\headroompct{tableTeal}{12} & $-$0.21\headroompct{tableTeal}{57} & 0.22\headroompct{tableTeal}{24} & +0.04\headroompct{tableTeal}{4} \\
Claude Opus~5 & 45 & +0.19\headroompct{tableTeal}{28} & \textbf{+0.021}\headroompct{tableTeal}{20} & 0.03\headroompct{tableTeal}{5} & 0.14\headroompct{tableTeal}{16} & \textbf{+0.01}\headroompct{tableTeal}{73} & \textbf{0.48}\headroompct{tableTeal}{52} & +0.10\headroompct{tableTeal}{10} \\
\tabledash
\rowcolor{tableTeal!7}
\expert & \textemdash & +0.29\headroompct{tableTeal}{42} & +0.017\headroompct{tableTeal}{16} & 0.14\headroompct{tableTeal}{22} & 0.38\headroompct{tableTeal}{44} & +0.15\headroompct{tableTeal}{83} & 0.64\headroompct{tableTeal}{69} & +0.84\headroompct{tableTeal}{84} \\
\tabledash
\rowcolor{tableTeal!4} \ceilingline & \textemdash & +0.68 & +0.106 & 0.66 & 0.86 & +0.39 & 0.92 & 1.00 \\
\addlinespace[2pt]\midrule
\rowcolor{tableTeal!12} \multicolumn{9}{@{}l}{\textcolor{tableTeal}{\omniOOD}} \\
\rowcolor{black!3} \floorline & \textemdash & +0.00 & +0.000 & 0.00 & 0.00 & $-$1.00 & 0.00 & +0.00 \\
\tabledash
gpt-oss-120b & 12 & $-$0.01\headroompct{tableRose}{-1} & $-$0.030\headroompct{tableRose}{-28} & 0.07\headroompct{tableTeal}{10} & 0.10\headroompct{tableTeal}{11} & $-$0.69\headroompct{tableTeal}{24} & 0.08\headroompct{tableTeal}{9} & $-$0.10\headroompct{tableRose}{-10} \\
\rowcolor{black!2}
Gemma~4 31B & 15 & +0.13\headroompct{tableTeal}{19} & +0.008\headroompct{tableTeal}{8} & 0.16\headroompct{tableTeal}{24} & 0.10\headroompct{tableTeal}{11} & $-$0.61\headroompct{tableTeal}{30} & 0.20\headroompct{tableTeal}{22} & +0.02\headroompct{tableTeal}{2} \\
Nemotron~3 Ultra & 23 & +0.05\headroompct{tableTeal}{7} & +0.000\headroompct{tableTeal}{0} & 0.04\headroompct{tableTeal}{7} & $-$0.01\headroompct{tableRose}{-1} & $-$0.80\headroompct{tableTeal}{16} & $-$0.19\headroompct{tableRose}{-21} & $-$0.29\headroompct{tableRose}{-29} \\
\rowcolor{black!2}
Inkling & 26 & +0.20\headroompct{tableTeal}{30} & +0.012\headroompct{tableTeal}{11} & 0.19\headroompct{tableTeal}{29} & 0.05\headroompct{tableTeal}{6} & $-$0.42\headroompct{tableTeal}{45} & $-$0.04\headroompct{tableRose}{-4} & $-$0.12\headroompct{tableRose}{-12} \\
Qwen3.8-27B & 34 & +0.27\headroompct{tableTeal}{40} & +0.002\headroompct{tableTeal}{2} & 0.15\headroompct{tableTeal}{23} & $-$0.22\headroompct{tableRose}{-24} & $-$0.41\headroompct{tableTeal}{45} & 0.16\headroompct{tableTeal}{18} & +0.01\headroompct{tableTeal}{1} \\
\rowcolor{black!2}
GPT-5.6~Sol & 39 & +0.29\headroompct{tableTeal}{43} & +0.021\headroompct{tableTeal}{20} & \textbf{0.30}\headroompct{tableTeal}{45} & 0.18\headroompct{tableTeal}{20} & $-$0.24\headroompct{tableTeal}{58} & \textbf{0.41}\headroompct{tableTeal}{44} & +0.20\headroompct{tableTeal}{20} \\
DeepSeek-V4.1-Flash & 40 & +0.21\headroompct{tableTeal}{30} & +0.016\headroompct{tableTeal}{16} & 0.24\headroompct{tableTeal}{36} & \textbf{0.19}\headroompct{tableTeal}{21} & $-$0.69\headroompct{tableTeal}{24} & 0.19\headroompct{tableTeal}{21} & \textbf{+0.44}\headroompct{tableTeal}{44} \\
\rowcolor{black!2}
Qwen3.8-Flash-Next & 40 & +0.28\headroompct{tableTeal}{41} & +0.007\headroompct{tableTeal}{6} & 0.17\headroompct{tableTeal}{25} & $-$0.02\headroompct{tableRose}{-3} & $-$0.63\headroompct{tableTeal}{29} & 0.17\headroompct{tableTeal}{18} & +0.04\headroompct{tableTeal}{4} \\
Kimi~K3 & 44 & +0.36\headroompct{tableTeal}{53} & \textbf{+0.021}\headroompct{tableTeal}{20} & 0.06\headroompct{tableTeal}{9} & 0.14\headroompct{tableTeal}{15} & $-$0.45\headroompct{tableTeal}{42} & 0.23\headroompct{tableTeal}{25} & +0.04\headroompct{tableTeal}{4} \\
\rowcolor{black!2}
GLM-5.3 & 45 & \textbf{+0.41}\headroompct{tableTeal}{60} & +0.021\headroompct{tableTeal}{20} & $-$0.04\headroompct{tableRose}{-7} & 0.04\headroompct{tableTeal}{4} & $-$0.33\headroompct{tableTeal}{51} & 0.11\headroompct{tableTeal}{12} & +0.02\headroompct{tableTeal}{2} \\
Claude Opus~5 & 45 & +0.33\headroompct{tableTeal}{48} & +0.017\headroompct{tableTeal}{17} & 0.26\headroompct{tableTeal}{40} & 0.18\headroompct{tableTeal}{20} & \textbf{$-$0.10}\headroompct{tableTeal}{70} & 0.19\headroompct{tableTeal}{21} & +0.08\headroompct{tableTeal}{8} \\
\tabledash
\rowcolor{tableTeal!7}
\expert & \textemdash & +0.37\headroompct{tableTeal}{54} & +0.023\headroompct{tableTeal}{22} & 0.29\headroompct{tableTeal}{45} & 0.55\headroompct{tableTeal}{60} & +0.06\headroompct{tableTeal}{81} & 0.34\headroompct{tableTeal}{37} & +0.85\headroompct{tableTeal}{85} \\
\tabledash
\rowcolor{tableTeal!4} \ceilingline & \textemdash & +0.68 & +0.105 & 0.66 & 0.90 & +0.30 & 0.92 & 1.00 \\
\bottomrule\end{tabular}
\end{table}

\FloatBarrier

\subsection{Detailed model profiles}
\begin{figure}[t]\centering
\includegraphics[width=\linewidth]{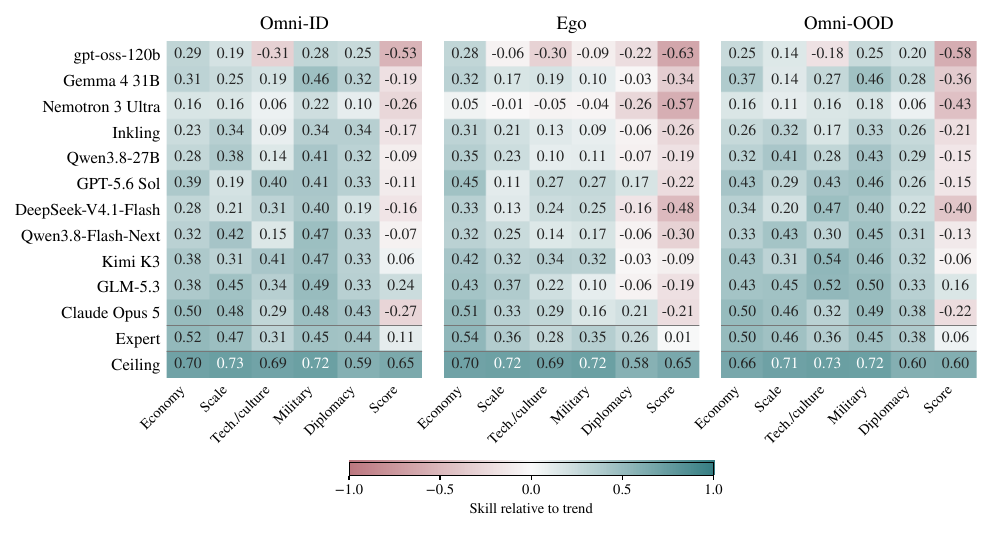}
\caption{Quantity-family skill across three settings. All panels share a color scale centered
on the trend reference at zero. Variables are equally weighted within each family.}
\label{fig:quantityfamilies}
\end{figure}

\Cref{fig:quantityfamilies} compares the six quantity families across all models and settings.
Overall-score increments remain difficult despite positive skill on many stock quantities.
The diagnostic equalizes variables within families as specified in \cref{app:metrics}.
\Cref{tab:traitsdetail,tab:rulesdetail} further show that hidden-target recovery varies
across attributes. \Cref{tab:main_full} shows model-dependent changes under OOD shift:
Flash-Next improves in quantity forecasting, Opus~5 changes little, and Gemma~4 declines.
The OOD set changes rulesets and trajectory budgets without matching ID difficulty or
selecting worlds to enforce a particular score ordering.
\begin{table}[t]\centering\scriptsize
\caption{Pairwise ranking accuracy by personality trait. Triples are Omni-ID / Ego / Omni-OOD, using the main-table samples and failure penalties.}
\label{tab:traitsdetail}
\begin{tabular}{@{}lcccc@{}}\toprule
Model & Aggressive & Expansionist & Trader & Builder \\ \midrule
gpt-oss-120b & 0.51 / 0.49 / 0.52 & 0.51 / 0.51 / 0.56 & 0.49 / 0.49 / 0.49 & 0.50 / 0.49 / 0.49 \\
Gemma~4 31B & 0.52 / 0.51 / 0.54 & 0.53 / 0.53 / 0.57 & 0.52 / 0.51 / 0.52 & 0.52 / 0.51 / 0.52 \\
Nemotron~3 Ultra & 0.51 / 0.48 / 0.50 & 0.52 / 0.52 / 0.55 & 0.53 / 0.50 / 0.52 & 0.51 / 0.49 / 0.48 \\
Inkling & 0.53 / 0.50 / 0.55 & 0.54 / 0.53 / 0.58 & 0.54 / 0.50 / 0.54 & 0.52 / 0.50 / 0.52 \\
Qwen3.8-27B & 0.52 / 0.50 / 0.53 & 0.53 / 0.53 / 0.57 & 0.52 / 0.50 / 0.52 & 0.53 / 0.51 / 0.52 \\
GPT-5.6~Sol & 0.54 / 0.52 / 0.54 & 0.55 / 0.49 / 0.61 & 0.59 / 0.54 / 0.58 & 0.55 / 0.53 / 0.56 \\
DeepSeek-V4.1-Flash & 0.51 / 0.49 / 0.54 & 0.54 / 0.53 / 0.59 & 0.56 / 0.51 / 0.56 & 0.54 / 0.50 / 0.54 \\
Qwen3.8-Flash-Next & 0.51 / 0.49 / 0.53 & 0.53 / 0.53 / 0.57 & 0.53 / 0.49 / 0.53 & 0.52 / 0.51 / 0.53 \\
Kimi~K3 & 0.48 / 0.49 / 0.50 & 0.50 / 0.52 / 0.55 & 0.52 / 0.50 / 0.52 & 0.50 / 0.50 / 0.49 \\
GLM-5.3 & 0.47 / 0.48 / 0.47 & 0.51 / 0.51 / 0.52 & 0.52 / 0.50 / 0.50 & 0.49 / 0.50 / 0.47 \\
Claude Opus~5 & 0.54 / 0.50 / 0.55 & 0.53 / 0.51 / 0.60 & 0.57 / 0.54 / 0.56 & 0.53 / 0.48 / 0.54 \\
\expert & 0.57 / 0.55 / 0.56 & 0.58 / 0.53 / 0.62 & 0.57 / 0.53 / 0.54 & 0.53 / 0.52 / 0.56 \\
\bottomrule\end{tabular}
\end{table}

\begin{table}[t]\centering\scriptsize
\caption{Classification accuracy by world rule. Triples are Omni-ID / Ego / Omni-OOD, using the main-table samples and failure penalties.}
\label{tab:rulesdetail}
\begin{tabular}{@{}lccc@{}}\toprule
Model & Disaster frequency & Growth cost & Research cost \\ \midrule
gpt-oss-120b & 0.63 / 0.60 / 0.61 & 0.34 / 0.35 / 0.37 & 0.32 / 0.36 / 0.41 \\
Gemma~4 31B & 0.65 / 0.64 / 0.62 & 0.39 / 0.38 / 0.34 & 0.36 / 0.37 / 0.41 \\
Nemotron~3 Ultra & 0.63 / 0.62 / 0.62 & 0.39 / 0.36 / 0.30 & 0.40 / 0.41 / 0.28 \\
Inkling & 0.65 / 0.64 / 0.62 & 0.36 / 0.39 / 0.35 & 0.39 / 0.40 / 0.34 \\
Qwen3.8-27B & 0.42 / 0.38 / 0.40 & 0.29 / 0.25 / 0.24 & 0.31 / 0.27 / 0.21 \\
GPT-5.6~Sol & 0.73 / 0.66 / 0.66 & 0.32 / 0.37 / 0.44 & 0.43 / 0.43 / 0.42 \\
DeepSeek-V4.1-Flash & 0.68 / 0.65 / 0.67 & 0.40 / 0.38 / 0.40 & 0.42 / 0.38 / 0.46 \\
Qwen3.8-Flash-Next & 0.56 / 0.61 / 0.55 & 0.41 / 0.40 / 0.32 & 0.39 / 0.41 / 0.31 \\
Kimi~K3 & 0.62 / 0.63 / 0.63 & 0.35 / 0.37 / 0.39 & 0.33 / 0.36 / 0.42 \\
GLM-5.3 & 0.58 / 0.61 / 0.54 & 0.37 / 0.36 / 0.34 & 0.38 / 0.41 / 0.40 \\
Claude Opus~5 & 0.71 / 0.64 / 0.63 & 0.37 / 0.33 / 0.39 & 0.36 / 0.46 / 0.50 \\
\expert & 0.77 / 0.76 / 0.72 & 0.66 / 0.46 / 0.77 & 0.68 / 0.61 / 0.63 \\
\bottomrule\end{tabular}
\end{table}

\begin{figure}[t]\centering
\includegraphics[width=\linewidth]{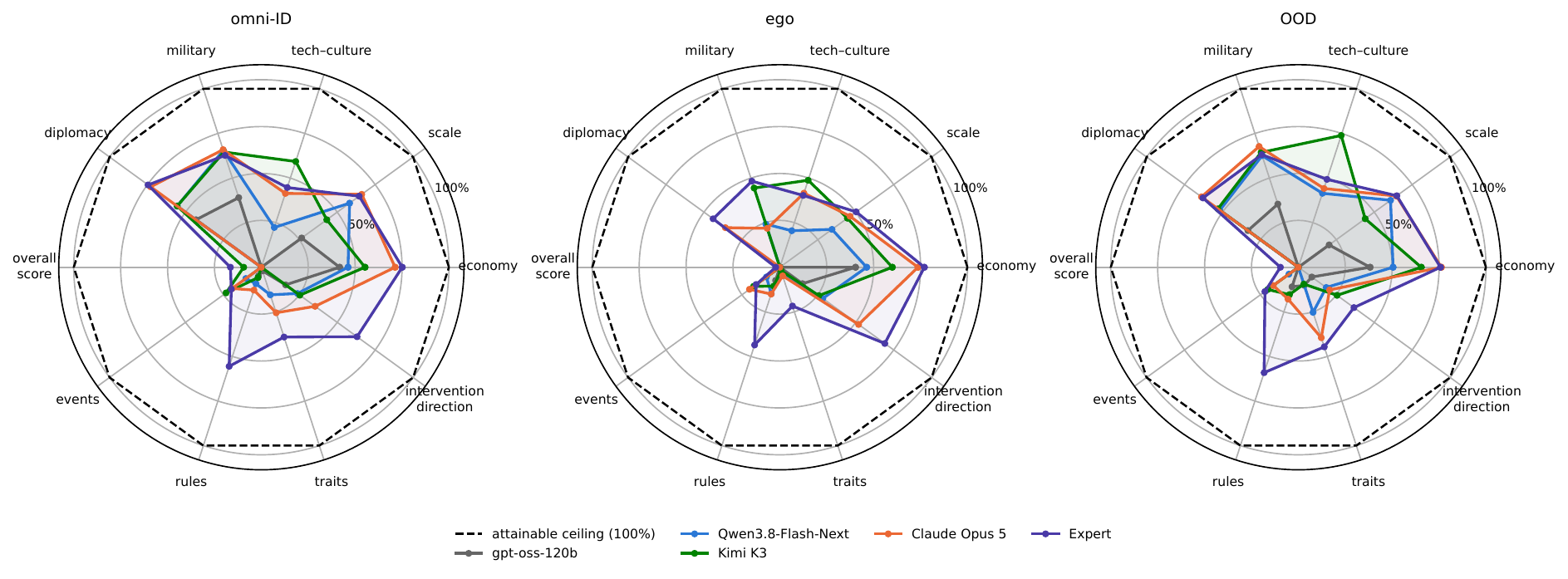}
\caption{Selected model profiles across three settings. Axes show reference-normalized
scores clipped to $[0,1]$ for display. The intervention axis uses direction, and radar
area is not an aggregate score.}
\label{fig:radar}
\end{figure}

\begin{figure}[t]\centering
\includegraphics[width=\linewidth]{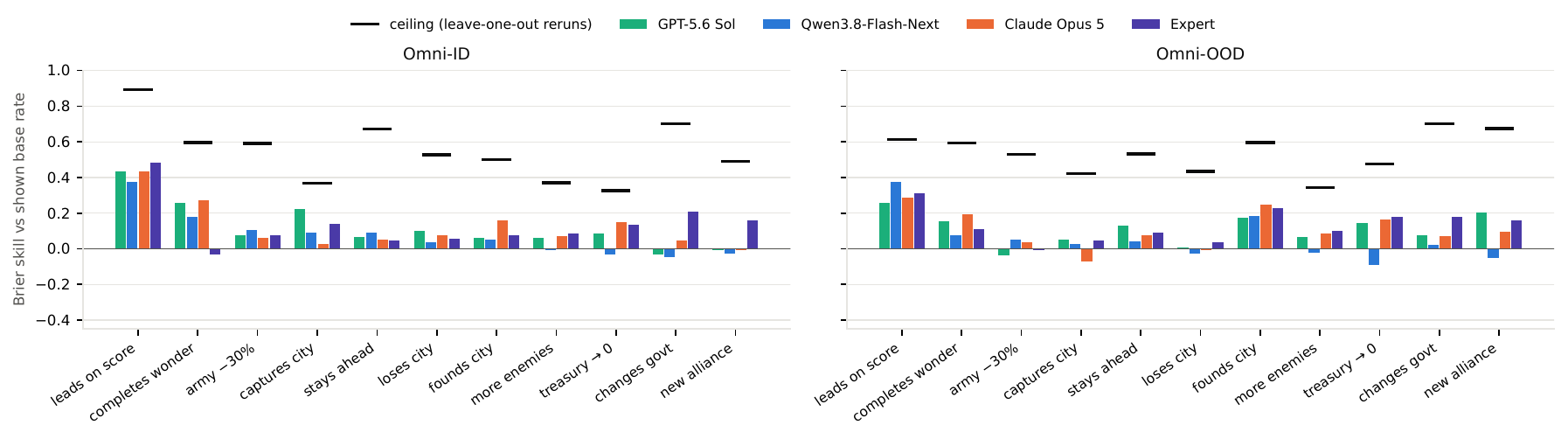}
\caption{Event-template Brier skill ratios in \omniID\ and \omniOOD, with leave-one-out
rerun references in black. This diagnostic uses $1-\sum\mathrm{Brier}/\sum\mathrm{Brier}_{\rm ref}$,
not the headline Brier gain.}
\label{fig:onebee}
\end{figure}
\FloatBarrier

\subsection{Information ablations and three-game study notes}\label{app:study}
\begin{table}[t]\centering\scriptsize
\caption{Paired information ablations of Flash-Next in Omni-ID. Each row uses its own common valid cells; intervals are paired world bootstraps. Event comparisons here use a no-base-rate baseline.}
\label{tab:abl}
\begin{tabular}{@{}llrrrrrl@{}}\toprule
Condition & Task & Worlds & Cells & Base & Added/ablated & Change & 95\% interval \\ \midrule
One frame & \QNT & 60 & 27251 & +0.269 & +0.137 & -0.132 & [-0.163, -0.103] \\
Future narrative & \QNT & 60 & 23085 & +0.285 & +0.346 & +0.061 & [+0.031, +0.091] \\
Rules primer & \QNT & 40 & 18027 & +0.266 & +0.247 & -0.019 & [-0.050, +0.011] \\
Study notes & \QNT & 60 & 26408 & +0.268 & +0.354 & +0.086 & [+0.061, +0.111] \\
Thinking off & \QNT & 60 & 27564 & +0.270 & +0.217 & -0.053 & [-0.078, -0.029] \\
Study notes & \EVT & 60 & 22018 & -0.034 & -0.015 & +0.019 & [+0.014, +0.025] \\
Base rate supplied & \EVT & 60 & 22018 & -0.034 & +0.012 & +0.046 & [+0.041, +0.052] \\
\bottomrule\end{tabular}
\end{table}

\paragraph{Paired design.}
The Flash-Next ablations use 60 basic worlds, or 240 anchors, in \omniID, except the
primer, which covers 40 worlds. Each row of \cref{tab:abl} uses its own common valid
cells and paired world-bootstrap interval, so baselines can differ across rows.
The conditions remove history or thinking, add generic rules, supply privileged future
narratives, or add study notes. Future narratives provide an information control rather
than a deployable forecast input.

\paragraph{Study input and notes.}
The one-shot study uses three independently seeded, 3-civilization, 400-turn games outside
the evaluation corpus. Each is summarized at turns 60, 145, 230,
315, and 400, with intervening narratives. The 411,688-character input produces an
11,221-character note reused across 1,624 evaluation prompts. No model parameters are
updated, and the note is not revised during evaluation.

\paragraph{Forecasting gains.}
\Cref{tab:abl} shows a paired quantity gain of \studyPairedGain\ with interval
\studyPairedInterval. Notes also improve event Brier gain from $-0.034$ to $-0.015$.
This event comparison uses the no-base-rate protocol, including eligible leader items.
Supplying base rates is a separate condition that reaches $+0.012$ from the same baseline.
\begin{table}[t]\centering\scriptsize
\caption{Study notes: paired CAPTURE on each arm's common valid moved cells in Omni-ID. Both arm and no-intervention predictions are note-conditioned. Intervals use paired world Bayesian bootstrap weights.}
\label{tab:studycapture}
\begin{tabular}{@{}lrrrrrrl@{}}\toprule
Arm & Cells & Worlds & Ops & Base & Notes & Change & 95\% interval \\ \midrule
real & 433 & 60 & 11 & -0.514 & -0.277 & +0.237 & [+0.070, +0.343] \\
placebo & 341 & 58 & 11 & -0.585 & -0.335 & +0.250 & [+0.053, +0.357] \\
\bottomrule\end{tabular}
\end{table}

\Cref{tab:studycapture} shows similar gains for real and placebo capture, both remaining
below zero. Notes also change the no-intervention forecast. Thus, learned context improves
forecasting broadly, while selective responses to intervention mechanisms remain difficult.

\subsection{Intervention breakdowns and mechanism cards}\label{app:mechanism}
\paragraph{Direction-conditioned magnitude distributions.}\label{app:directionmagnitude}
\Cref{fig:intervention} examines effect magnitudes among correct-sign predictions under the
main score's real-arm, target-variable, horizon-band, and truth-movement rules. Predicted and
realized effects use the transformed, normalized differences in \cref{sub:scores}.
Eligible cells exclude undefined control-relative loss ratios, including zero-loss control
forecasts, with counts in \cref{tab:magnitudesensitivity}. Diagnostic direction accuracy is
the cell-weighted fraction of correct signs among valid nonzero responses. The main score
additionally penalizes missing answers and weights operator scores by eligible-cell counts.
\Cref{tab:directionmagnitude0,tab:directionmagnitude1,tab:directionmagnitude2} report missing
and zero responses separately. Relative magnitude is $|\widehat{\delta}|/|\delta|$, where
$\widehat{\delta}$ and $\delta$ denote predicted and realized effects. Figure medians and
25th--75th percentile bars summarize its distribution across correct-sign cells, with
100\% denoting a magnitude match. Ratios are cell-weighted and unclipped, with linearly
interpolated quantiles. Additional bins classify predictions below half, within a factor
of two, or above twice the realized magnitude, using inequalities in the same scoring space.
The conditional score $S_{+}$ applies the clipped equal-operator Capture formula to
correct-sign cells, retaining operators with at least three such cells.

\paragraph{Conditional results.}
\Cref{tab:relativemagnitude} shows that near-matching medians coexist with broad magnitude
distributions across settings. In \omniID, six of eleven models have negative $S_{+}$
estimates, with intervals entirely below zero for gpt-oss, Gemma, and DeepSeek. For these
models, magnitude errors persist even when the direction is correct. GLM-5.3 has positive
$S_{+}$ with an interval above zero, while its main Capture remains negative and its direction
intervals exceed chance in all three settings. Its conditional improvement illustrates the
contribution of wrong-direction predictions to overall error. A correct-sign prediction
above twice the realized magnitude has greater absolute error than a zero-effect prediction.
\Cref{tab:magnitudesensitivity} shows that these overshoots persist, at lower frequencies,
after removing effects at or below each operator's lower quartile before conditioning on
correct signs. Underestimation and overestimation therefore remain relevant beyond the
smallest effects, motivating state-conditioned magnitude calibration.

\paragraph{Sampling and interpretation.}
The magnitude bars describe dispersion across predictions, and intervals follow
\cref{app:metrics}. Model-specific samples and correct-sign subsets differ, so these
comparisons diagnose within-model errors. Conditioning also changes the cells and operators
included in $S_{+}$ relative to main Capture. The targets are realized stochastic branch
differences, and future calibration studies should evaluate gains on held-out worlds.
\begin{table}[t]\centering\scriptsize
\setlength{\tabcolsep}{2pt}\renewcommand{\arraystretch}{1.12}
\caption{Direction-conditioned magnitude diagnostics in omni-ID. Counts give eligible, missing, zero-response, and correct-sign cells. Direction is the percentage correct among valid nonzero responses. Magnitude-bin percentages condition on a correct sign. Brackets give 95\% world-bootstrap intervals. $S_{+}$ is selected-subset capture, not a replacement main score. An asterisk marks an omitted interval due to a singleton-world operator.}
\label{tab:directionmagnitude0}
\begin{tabular}{@{}lrrrrrr@{}}\toprule
Model & Counts & Direction & $<0.5\times$ & $0.5$--$2\times$ & $>2\times$ & $S_{+}$ \\
\midrule
gpt-oss-120b & 1749/57/105/937 & 59 [56, 62] & 21 & 45 & 34 [30, 38] & -0.34 [-0.44, -0.16] \\
Gemma 4 31B & 1759/1/145/1033 & 64 [61, 67] & 18 & 48 & 34 [31, 38] & -0.31 [-0.41, -0.20] \\
Nemotron 3 Ultra & 1767/545/93/717 & 64 [60, 67] & 15 & 52 & 33 [29, 37] & -0.07 [-0.20, +0.03] \\
Inkling & 1772/169/178/871 & 61 [58, 65] & 27 & 51 & 21 [18, 25] & +0.20 [+0.10, +0.28] \\
Qwen3.8-27B & 1764/56/100/1017 & 63 [60, 66] & 23 & 50 & 27 [24, 30] & +0.03 [-0.07, +0.13] \\
GPT-5.6 Sol & 176/0/16/103 & 64 [54, 74] & 9 & 57 & 34 [25, 44] & -0.07 [-0.30, +0.18] \\
DeepSeek-V4.1-Flash & 1769/4/164/1027 & 64 [61, 67] & 17 & 48 & 35 [32, 38] & -0.27 [-0.38, -0.15] \\
Flash-Next & 1768/52/77/1018 & 62 [59, 65] & 26 & 47 & 27 [24, 31] & -0.11 [-0.18, +0.02] \\
Kimi K3 & 1774/8/123/1006 & 61 [58, 64] & 19 & 53 & 28 [25, 32] & +0.08 [-0.04, +0.17] \\
GLM-5.3 & 1773/142/142/937 & 63 [60, 66] & 26 & 52 & 22 [19, 25] & +0.16 [+0.10, +0.26] \\
Claude Opus 5 & 176/4/35/94 & 69 [58, 78] & 33 & 50 & 17 [10, 26] & +0.32 [+0.06, +0.46] \\
\midrule
Expert & 176/0/59/90 & 77 [65, 87] & 36 & 57 & 8 [3, 14] & +0.33 [+0.23, +0.41] \\
\bottomrule\end{tabular}
\end{table}

\begin{table}[t]\centering\scriptsize
\setlength{\tabcolsep}{2pt}\renewcommand{\arraystretch}{1.12}
\caption{Direction-conditioned magnitude diagnostics in ego. Counts give eligible, missing, zero-response, and correct-sign cells. Direction is the percentage correct among valid nonzero responses. Magnitude-bin percentages condition on a correct sign. Brackets give 95\% world-bootstrap intervals. $S_{+}$ is selected-subset capture, not a replacement main score. An asterisk marks an omitted interval due to a singleton-world operator.}
\label{tab:directionmagnitude1}
\begin{tabular}{@{}lrrrrrr@{}}\toprule
Model & Counts & Direction & $<0.5\times$ & $0.5$--$2\times$ & $>2\times$ & $S_{+}$ \\
\midrule
gpt-oss-120b & 1753/25/132/917 & 57 [54, 61] & 20 & 48 & 32 [28, 36] & -0.33 [-0.44, -0.20] \\
Gemma 4 31B & 1767/2/185/995 & 63 [60, 66] & 24 & 48 & 27 [24, 30] & -0.10 [-0.20, +0.02] \\
Nemotron 3 Ultra & 1767/480/104/700 & 59 [55, 63] & 16 & 47 & 37 [33, 41] & -0.24 [-0.36, -0.13] \\
Inkling & 1774/135/218/885 & 62 [59, 66] & 27 & 49 & 24 [20, 27] & +0.12 [+0.04, +0.23] \\
Qwen3.8-27B & 1770/11/116/1014 & 62 [59, 65] & 22 & 50 & 29 [25, 32] & +0.04 [-0.07, +0.16] \\
GPT-5.6 Sol & 176/0/14/107 & 66 [56, 75] & 17 & 60 & 23 [15, 35] & +0.11 [-0.10, +0.28] \\
DeepSeek-V4.1-Flash & 1772/20/157/977 & 61 [58, 64] & 15 & 48 & 37 [33, 40] & -0.32 [-0.43, -0.18] \\
Flash-Next & 1774/13/119/1028 & 63 [60, 66] & 25 & 48 & 27 [23, 30] & -0.09 [-0.16, +0.03] \\
Kimi K3 & 1774/4/123/1007 & 61 [58, 64] & 22 & 53 & 26 [22, 29] & +0.14 [+0.06, +0.20] \\
GLM-5.3 & 1774/60/130/992 & 63 [60, 66] & 25 & 54 & 21 [18, 24] & +0.18 [+0.11, +0.26] \\
Claude Opus 5 & 176/0/28/108 & 73 [63, 81] & 30 & 58 & 12 [6, 19] & +0.37 [+0.26, +0.45] \\
\midrule
Expert & 176/0/51/97 & 78 [66, 87] & 35 & 56 & 9 [4, 15] & +0.32 [+0.21, +0.38] \\
\bottomrule\end{tabular}
\end{table}

\begin{table}[t]\centering\scriptsize
\setlength{\tabcolsep}{2pt}\renewcommand{\arraystretch}{1.12}
\caption{Direction-conditioned magnitude diagnostics in OOD. Counts give eligible, missing, zero-response, and correct-sign cells. Direction is the percentage correct among valid nonzero responses. Magnitude-bin percentages condition on a correct sign. Brackets give 95\% world-bootstrap intervals. $S_{+}$ is selected-subset capture, not a replacement main score. An asterisk marks an omitted interval due to a singleton-world operator.}
\label{tab:directionmagnitude2}
\begin{tabular}{@{}lrrrrrr@{}}\toprule
Model & Counts & Direction & $<0.5\times$ & $0.5$--$2\times$ & $>2\times$ & $S_{+}$ \\
\midrule
gpt-oss-120b & 1781/118/111/903 & 58 [55, 61] & 16 & 42 & 42 [38, 46] & -0.42 [-0.52, -0.28] \\
Gemma 4 31B & 1793/2/163/971 & 60 [57, 62] & 16 & 40 & 44 [41, 48] & -0.31 [-0.40, -0.24] \\
Nemotron 3 Ultra & 1802/592/66/716 & 63 [59, 67] & 15 & 41 & 44 [40, 49] & -0.29 [-0.44, -0.15] \\
Inkling & 1804/228/137/812 & 56 [53, 59] & 21 & 52 & 28 [24, 31] & +0.04 [-0.09, +0.16] \\
Qwen3.8-27B & 1802/80/87/991 & 61 [58, 64] & 16 & 46 & 39 [35, 42] & -0.13 [-0.22, -0.00] \\
GPT-5.6 Sol & 169/0/12/108 & 69 [59, 79] & 10 & 46 & 44 [33, 56] & -0.07 [-0.27, +0.11] \\
DeepSeek-V4.1-Flash & 1803/25/167/967 & 60 [57, 63] & 14 & 43 & 43 [39, 47] & -0.36 [-0.48, -0.23] \\
Flash-Next & 1802/69/84/998 & 61 [58, 64] & 19 & 43 & 38 [35, 42] & -0.21 [-0.33, -0.05] \\
Kimi K3 & 1802/12/114/1030 & 61 [59, 65] & 16 & 46 & 37 [34, 41] & -0.06 [-0.15, +0.03] \\
GLM-5.3 & 1807/101/134/929 & 59 [56, 62] & 22 & 50 & 28 [25, 31] & +0.08 [-0.01, +0.20] \\
Claude Opus 5 & 169/0/18/90 & 60 [49, 70] & 27 & 50 & 23 [13, 36] & +0.25 [+0.11, +0.37] \\
\midrule
Expert & 169/0/59/73 & 66 [53, 79] & 48 & 37 & 15 [8, 27] & +0.23 [+0.13, +0.33] \\
\bottomrule\end{tabular}
\end{table}

\begin{table}[t]\centering\scriptsize
\setlength{\tabcolsep}{3pt}
\caption{Magnitude sensitivity in \omniID. All percentages condition on a correct sign. The sensitivity filter removes effects at or below each operator's lower quartile before selecting correct-sign predictions. Excluded counts refer instead to valid predictions omitted by the inherited main scoring mask.}
\label{tab:magnitudesensitivity}
\begin{tabular}{@{}lrrrrrr@{}}\toprule
Model & $<1\times$ & $>1\times$ & $>2\times$ & Filtered $>2\times$ & Filtered $n_{+}$ & Excluded \\
\midrule
gpt-oss-120b & 38.1 & 52.7 & 33.7 & 27.6 & 646 & 25 \\
Gemma 4 31B & 38.7 & 54.1 & 34.1 & 25.6 & 738 & 15 \\
Nemotron 3 Ultra & 33.5 & 57.3 & 33.3 & 25.7 & 501 & 7 \\
Inkling & 50.9 & 40.4 & 21.4 & 12.5 & 615 & 2 \\
Qwen3.8-27B & 42.1 & 49.7 & 26.8 & 17.5 & 718 & 10 \\
GPT-5.6 Sol & 25.2 & 68.0 & 34.0 & 23.5 & 68 & 0 \\
DeepSeek-V4.1-Flash & 35.6 & 57.9 & 35.0 & 27.1 & 730 & 5 \\
Flash-Next & 44.5 & 47.6 & 27.2 & 18.6 & 714 & 6 \\
Kimi K3 & 40.5 & 50.6 & 28.3 & 19.9 & 695 & 0 \\
GLM-5.3 & 47.5 & 43.3 & 22.1 & 13.6 & 664 & 1 \\
Claude Opus 5 & 52.1 & 39.4 & 17.0 & 7.7 & 65 & 0 \\
Expert & 58.9 & 36.7 & 7.8 & 3.1 & 65 & 0 \\
\bottomrule\end{tabular}
\end{table}

\begin{table}[t]\centering\scriptsize
\setlength{\tabcolsep}{5pt}
\caption{Relative effect magnitudes conditional on correct direction. Entries give the median and [25th, 75th] percentiles, all in percentages. The ratio uses absolute predicted and realized effects in the scoring space. 100\% denotes a magnitude match. Brackets describe sample dispersion, not confidence intervals.}
\label{tab:relativemagnitude}
\begin{tabular}{@{}lrrr@{}}\toprule
Model & \omniID & \ego & \omniOOD \\
\midrule
gpt-oss-120b & 120.0 [58.3, 300.0] & 106.2 [55.7, 300.0] & 158.0 [68.4, 455.4] \\
Gemma 4 31B & 122.2 [62.5, 300.0] & 100.0 [50.0, 229.7] & 172.8 [82.1, 566.6] \\
Nemotron 3 Ultra & 123.3 [73.8, 300.0] & 140.0 [68.6, 327.5] & 183.3 [83.3, 407.8] \\
Inkling & 96.4 [43.4, 180.0] & 100.0 [45.1, 200.0] & 105.3 [53.1, 246.1] \\
Qwen3.8-27B & 100.0 [50.0, 225.0] & 108.5 [52.1, 244.0] & 150.0 [82.0, 349.0] \\
GPT-5.6 Sol & 160.0 [99.7, 272.1] & 133.3 [69.8, 200.0] & 176.0 [99.0, 406.6] \\
DeepSeek-V4.1-Flash & 130.8 [66.7, 300.0] & 136.7 [72.2, 333.3] & 162.5 [81.8, 468.3] \\
Flash-Next & 100.0 [46.9, 233.3] & 100.0 [50.0, 233.3] & 140.0 [63.8, 328.4] \\
Kimi K3 & 103.9 [57.8, 250.0] & 100.0 [50.5, 210.6] & 140.8 [66.7, 329.0] \\
GLM-5.3 & 100.0 [45.5, 200.0] & 100.0 [47.1, 175.9] & 100.0 [54.5, 235.3] \\
Claude Opus 5 & 85.7 [33.3, 161.4] & 100.0 [40.0, 150.0] & 100.0 [40.2, 200.0] \\
Expert & 75.7 [22.2, 157.5] & 74.3 [23.5, 157.5] & 50.0 [18.2, 130.8] \\
\bottomrule\end{tabular}
\end{table}

\begin{table}[t]\centering\scriptsize
\setlength{\tabcolsep}{5pt}
\caption{Intervention diagnostics across three settings. Capture and shift are reported for real arms with matched placebos in parentheses. Shift is the mean absolute change from the model's no-intervention forecast in the normalized units of \cref{sub:scores}. Specificity applies the missing-response penalty to the clipped shift ratio defined in \cref{eq:spec,app:metrics}. Triples are \omniID\ / \ego\ / \omniOOD.}
\label{tab:l3}
\begin{tabular}{@{}lccc@{}}
\toprule
Line & capture real (placebo) & direction & shift real (placebo) \\
\midrule
gpt-oss-120b & $-$0.65 ($-$0.74) / $-$0.73 ($-$0.77) / $-$0.69 ($-$0.87) & 0.57 / 0.56 / 0.54 & 0.56 (0.50) / 0.63 (1.15) / 0.58 (0.60) \\
Gemma~4 31B & $-$0.61 ($-$0.62) / $-$0.51 ($-$0.56) / $-$0.61 ($-$0.69) & 0.64 / 0.63 / 0.59 & 0.38 (0.30) / 0.29 (0.27) / 0.41 (0.40) \\
Nemotron~3 Ultra & $-$0.63 ($-$0.74) / $-$0.68 ($-$0.83) / $-$0.80 ($-$0.87) & 0.43 / 0.42 / 0.41 & 0.28 (0.28) / 0.31 (0.29) / 0.28 (0.24) \\
Inkling & $-$0.31 ($-$0.43) / $-$0.31 ($-$0.44) / $-$0.42 ($-$0.55) & 0.54 / 0.56 / 0.48 & 0.15 (0.15) / 0.18 (0.17) / 0.14 (0.14) \\
Qwen3.8-27B & $-$0.34 ($-$0.50) / $-$0.33 ($-$0.47) / $-$0.41 ($-$0.66) & 0.61 / 0.61 / 0.57 & 0.20 (0.20) / 0.27 (0.26) / 0.18 (0.17) \\
GPT-5.6~Sol & $-$0.60 ($-$0.70) / $-$0.27 ($-$0.60) / $-$0.24 ($-$0.50) & 0.63 / 0.65 / 0.69 & 0.17 (0.17) / 0.17 (0.15) / 0.15 (0.12) \\
DeepSeek-V4.1-Flash & $-$0.58 ($-$0.66) / $-$0.69 ($-$0.69) / $-$0.69 ($-$0.86) & 0.64 / 0.60 / 0.59 & 0.26 (0.24) / 0.31 (0.27) / 0.39 (0.21) \\
Qwen3.8-Flash-Next & $-$0.57 ($-$0.56) / $-$0.39 ($-$0.57) / $-$0.63 ($-$0.74) & 0.60 / 0.62 / 0.58 & 0.25 (0.23) / 0.31 (0.28) / 0.22 (0.21) \\
Kimi~K3 & $-$0.35 ($-$0.52) / $-$0.24 ($-$0.50) / $-$0.45 ($-$0.68) & 0.61 / 0.61 / 0.61 & 0.18 (0.16) / 0.18 (0.17) / 0.14 (0.14) \\
GLM-5.3 & $-$0.25 ($-$0.47) / $-$0.21 ($-$0.39) / $-$0.33 ($-$0.52) & 0.57 / 0.60 / 0.55 & 0.13 (0.12) / 0.16 (0.15) / 0.11 (0.10) \\
Claude Opus~5 & $-$0.08 ($-$0.26) / +0.01 ($-$0.04) / $-$0.10 ($-$0.00) & 0.65 / 0.72 / 0.59 & 0.07 (0.06) / 0.07 (0.07) / 0.06 (0.05) \\
\expert & +0.13 (+0.00) / +0.15 (+0.01) / +0.06 (+0.01) & 0.76 / 0.79 / 0.65 & 0.00 (0.00) / 0.00 (0.00) / 0.00 (0.00) \\
\bottomrule\end{tabular}
\end{table}

\begin{table}[t]\centering\scriptsize
\caption{Paired Capture with mechanism cards and a shared no-intervention prediction. Counts denote cells/operators. Real and placebo arms use separate moved-cell sets.}
\label{tab:mechanism}
\begin{tabular}{@{}llllrrr@{}}\toprule
Model & Setting & Arm & Cells/ops & Base & Card & Change \\ \midrule
Flash-Next & omni-ID & real & 48/7 & -0.581 & -0.678 & -0.097 \\
Flash-Next & omni-ID & placebo & 44/8 & -0.436 & -0.626 & -0.190 \\
Flash-Next & ego & real & 52/8 & -0.406 & -0.231 & +0.175 \\
Flash-Next & ego & placebo & 44/8 & -0.349 & -0.237 & +0.112 \\
Flash-Next & OOD & real & 39/9 & -0.482 & -0.592 & -0.110 \\
Flash-Next & OOD & placebo & 15/4 & -0.251 & -0.372 & -0.121 \\
Qwen3.8-27B & omni-ID & real & 48/7 & -0.327 & -0.231 & +0.096 \\
Qwen3.8-27B & omni-ID & placebo & 44/8 & -0.431 & -0.458 & -0.027 \\
Qwen3.8-27B & ego & real & 52/8 & -0.436 & -0.214 & +0.222 \\
Qwen3.8-27B & ego & placebo & 44/8 & -0.614 & -0.435 & +0.179 \\
Qwen3.8-27B & OOD & real & 38/9 & -0.300 & -0.462 & -0.162 \\
Qwen3.8-27B & OOD & placebo & 21/4 & -0.772 & -0.743 & +0.029 \\
Gemma 4 31B & omni-ID & real & 52/8 & -0.521 & -0.586 & -0.065 \\
Gemma 4 31B & omni-ID & placebo & 44/8 & -0.547 & -0.639 & -0.092 \\
Gemma 4 31B & ego & real & 52/8 & -0.450 & -0.463 & -0.014 \\
Gemma 4 31B & ego & placebo & 44/8 & -0.396 & -0.408 & -0.012 \\
Gemma 4 31B & OOD & real & 46/9 & -0.595 & -0.595 & -0.000 \\
Gemma 4 31B & OOD & placebo & 24/4 & -0.663 & -0.800 & -0.137 \\
\bottomrule\end{tabular}
\end{table}

\begin{figure}[tbp]
\begin{minipage}[t]{0.49\linewidth}
\vspace{0pt}
\centering
\includegraphics[width=\linewidth]{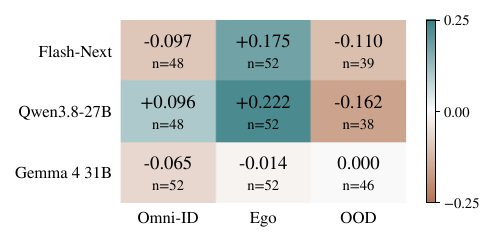}
\caption{Mechanism-card effects with fixed control forecasts.
Colors show Capture changes and counts give matched cells.}
\label{fig:mechanismcards}
\end{minipage}\hfill
\begin{minipage}[t]{0.49\linewidth}
\vspace{0pt}
\centering
\includegraphics[height=0.50\linewidth]{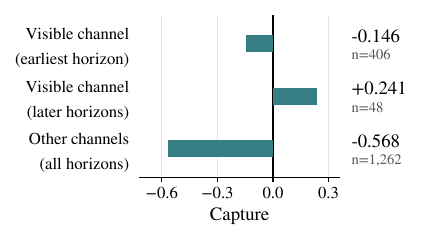}
\caption{Flash-Next's \omniID\ Capture by channel and horizon group.
Operator compositions differ across the three groups.}
\label{fig:interventionchannels}
\end{minipage}
\end{figure}

\paragraph{Mechanism-card design.}
The mechanism-card study contains 60 selected anchors and 310 intervention prompts, with a note on 260
real/placebo rows across 11 operators. Each note states generic engine mechanics, is identical
across worlds for its operator, and is inserted immediately before the task block. The same
note is used for real and placebo edits. No future arm outcome is supplied.
For the paired analysis, both conditions use the corresponding model's main-experiment
no-intervention median. Holding this reference fixed attributes paired changes to the
edited-arm forecasts.
\Cref{fig:mechanismcards} reports these common-control comparisons.

\paragraph{Mechanism-card results.}
\Cref{fig:mechanismcards} shows that Qwen3.8-27B improves in \omniID, while Flash-Next and Gemma
decline. All three remain below the zero-effect reference. Flash-Next and Qwen3.8-27B improve
in \ego, whereas OOD gains do not persist. Responses to mechanism information thus depend
on both the model and evaluation setting. In \omniID, common valid real-arm cells number
48, 48, and 52 for Flash-Next, Qwen3.8-27B, and Gemma, covering 7, 7, and 8 operators,
respectively. Pairing is performed within each model. Only Flash-Next's \omniID\ real-arm
comparison has at least two worlds per operator, giving a 95\% paired world-bootstrap
interval of $[-0.170,+0.028]$. Each of the other 17 comparisons includes a singleton-world
operator. \Cref{tab:mechanism} reports their point estimates, with between-world intervals
omitted under the same sampling rule.

\paragraph{Channel and horizon diagnostic.}
\Cref{fig:interventionchannels} keeps the formal target variables, horizon bands, and moved-cell mask.
Its shortest horizon is determined from asked target cells before filtering for truth movement
or valid model output. Visible-channel variables are treasury for the gold edit, technologies
learned for research points or a technology grant, and city count/total city size for city
destruction, subject to each operator's formal scored-variable mask. We partition the eligible
cells into three disjoint groups: the designated visible channels at the earliest eligible horizon, covering 406 cells from
three operators, the same channels at later eligible horizons, covering 48 cells from city
destruction only, and all other channels at any eligible horizon, covering 1,262 cells from
nine operators. Earliest/later is relative to each world, anchor, and operator, not a global
short/long cutoff. If the earliest prediction is missing or its truth is unmoved, a later
prediction does not become an earliest-horizon cell. The technology-grant operator has no
eligible moved cell in either visible group and receives no score there.
Other channels include remaining variables under visible-state edits as well as responses to
hidden-state edits. All three groups contain observable quantity forecasts in the global-view setting.

The positive later-horizon score concerns city destruction alone. Differences across these
groups therefore combine horizon and operator composition. The direct-effect sign check
in \cref{app:cards} complements this analysis with a broader target set, including city count
outside the main Capture mask.

\section{Process verification protocols and diagnostics}\label{app:cot}
\subsection{Scope and output protocols}
The aligned process study evaluates Flash-Next, Qwen3.8-27B, Gemma~4~31B, and Kimi~K3
on the same 190 anchors, of which 90 are Omni-ID and 100 OOD, and GPT-5.6~Sol on 30 of
these anchors, 17 Omni-ID and 13 OOD. Sol reuses compatible Free responses on 22 overlapping anchors
and adds responses on eight selected by budget and phase metadata. Table and Code cover
all 30. Each model contributes its full evaluated sample.
Free and Table also run the egocentric condition. Code has no masked structured-input
implementation. Main-text comparisons average Omni-ID and OOD equally.
\begin{table}[t]\centering\scriptsize
\caption{Completed output census and diagnostic coverage. F/T/C counts include all executed settings. Missing state/effect units are counted before protocol matching.}
\label{tab:cotalignedcoverage}
\begin{tabular}{@{}lrrrrr@{}}\toprule
Model & Anchors & Outputs F/T/C & State missing & Effect missing & Runnable \\ \midrule
Qwen3.8-27B & 190 & 1537/1537/1034 & 200/4272 & 142/2320 & 1034/1034 \\
Gemma 4 31B & 190 & 1537/1537/1034 & 205/4272 & 157/2320 & 1033/1034 \\
Flash-Next & 190 & 1537/1537/1034 & 205/4272 & 159/2320 & 1033/1034 \\
Kimi K3 & 190 & 1537/1537/1034 & 238/4272 & 180/2320 & 1033/1034 \\
GPT-5.6 Sol & 30 & 269/269/169 & 48/728 & 29/400 & 169/169 \\
\bottomrule\end{tabular}
\end{table}

\paragraph{Free and Table.}
Free produces ordinary answers with unrestricted reasoning and has no prescribed process
score. Table adds explicit intermediate predictions before answering. For quantitative
questions, the aligned protocol prescribes cells near one-quarter, one-half, and
three-quarters of each horizon, excluding all answer times. The target and quantities
are selected from question metadata before inference. Table must include these cells
and may add up to eight other claims. Its claim types cover quantities, diplomatic states,
governments, technologies, city ownership, and wonders. Other questions retain the
general 8--40-line requirement. Numeric process scores use the prescribed intermediate cells.

\paragraph{Code and feedback.}
Code receives structured observation frames and the ruleset name, with the intervention
applied when appropriate. It writes \texttt{predict(world, rng)}, whose 50 executions yield
answer quantiles, event frequencies, and intermediate predictions. Up to four generations
allow execution repairs and back-testing from an earlier observed frame to the last
observed frame. Future evaluation truth is never returned as feedback. Back-testing stops
when at most the larger of two cells or 15\% of checked cells disagree, or after two
rounds without improvement. The retained runnable round minimizes disagreement.
The comparison evaluates complete reasoning protocols, including execution and feedback.

\subsection{Scoring units and aggregation}\label{app:cotcriteria}
A state unit specifies an anchor, question, civilization, quantity, and answer horizon.
Intermediate errors average transformed, scale-normalized absolute errors at the
prescribed times. Table supplies point predictions, while Code supplies the mean of its
sampled intermediate states. Final point errors use the answer median. Table and Code are paired
on common valid units within each model and scope. L1 uses ordinary quantity questions.
L3-state uses real-intervention branch states. This paired diagnostic includes only units
with available truth and valid predictions in both protocols.

Answer-positive (A+) means final point error is strictly below its trend-reference error.
Process-positive (P+) applies the same criterion to intermediate errors. P- denotes an
intermediate error at or above the reference. The joint A+P- fraction is the number of
above-reference answers with intermediate forecasts at or below reference performance,
divided by all eligible units.

Intermediate-state skill is computed within each world and quantity family as
$\operatorname{clip}(1-\sum e/\sum e_{\mathrm{trend}},-1,1)$ for positive reference
denominators, averaged over families and then worlds. Both this skill and quadrant
shares are averaged equally over Omni-ID and OOD. Table and Code use the same valid
units within a model, although eligible sets differ across models. These diagnostics
evaluate intermediate states and final point forecasts. Formal Quantity and Capture scores
separately assess forecast distributions and intervention effects.
\begin{table}[t]\centering\scriptsize
\caption{Intermediate-state skill and joint answer-positive/process-negative shares. Percentages use all eligible matched units. L3-state evaluates intervention-branch states.}
\label{tab:cotalignedprocess}
\begin{tabular}{@{}llrrrr@{}}\toprule
Model & Protocol & L1 skill & L3-state skill & L1 A+P- (\%) & L3-state A+P- (\%) \\ \midrule
Qwen3.8-27B & Table & $-0.027$ & $+0.067$ & 10.55 & 5.66 \\
Qwen3.8-27B & Code & $-0.006$ & $-0.049$ & 5.84 & 4.82 \\
Gemma 4 31B & Table & $-0.058$ & $-0.036$ & 10.01 & 5.92 \\
Gemma 4 31B & Code & $-0.013$ & $-0.050$ & 7.38 & 4.49 \\
Flash-Next & Table & $-0.023$ & $+0.028$ & 7.63 & 9.16 \\
Flash-Next & Code & $-0.065$ & $-0.001$ & 6.49 & 4.54 \\
Kimi K3 & Table & $+0.021$ & $+0.017$ & 10.32 & 8.53 \\
Kimi K3 & Code & $+0.000$ & $+0.153$ & 5.65 & 5.08 \\
GPT-5.6 Sol & Table & $+0.064$ & $+0.115$ & 3.08 & 8.17 \\
GPT-5.6 Sol & Code & $+0.035$ & $+0.032$ & 7.35 & 2.08 \\
\bottomrule\end{tabular}
\end{table}

\Cref{fig:cot}a additionally matches Free final answers to these Table/Code units.
Within each model, all three formats use the same eligible cells, excluding missing
Free answers and nonfinite errors. Bar heights give the fraction of final median
errors strictly below the trend reference, averaged equally across L1 and L3 branch-state
scopes and then Omni-ID and OOD. Hatched segments give the joint A+P- fraction on the same
denominator. The combined diagnostic concerns numerical
point forecasts, with event forecasting evaluated separately.

Formal final-answer scores retain the benchmark's existing scorer and reference cells.
L3 Capture averages operator-level effect skills under the frozen eligibility,
clipping, and control definitions. Its reported value is the equal Omni-ID/OOD mean.
Raw Capture measures improvement over predicting no effect and can be negative.
\begin{table}[t]\centering\scriptsize
\caption{Final-answer scores averaged equally over Omni-ID and OOD. Each model uses 190 anchors, and GPT-5.6 Sol uses 30. Protocols use the same anchors within each model.}
\label{tab:cotalignedanswers}
\begin{tabular}{@{}llrrr@{}}\toprule
Model & Protocol & \QNT & \EVT & Capture \\ \midrule
Qwen3.8-27B & Free & $+0.206$ & $+0.012$ & $-0.408$ \\
Qwen3.8-27B & Table & $+0.171$ & $+0.009$ & $-0.456$ \\
Qwen3.8-27B & Code & $+0.127$ & $-0.087$ & $-0.642$ \\
Gemma 4 31B & Free & $+0.153$ & $+0.009$ & $-0.423$ \\
Gemma 4 31B & Table & $+0.120$ & $+0.007$ & $-0.443$ \\
Gemma 4 31B & Code & $+0.015$ & $-0.129$ & $-0.605$ \\
Flash-Next & Free & $+0.204$ & $+0.015$ & $-0.428$ \\
Flash-Next & Table & $+0.150$ & $+0.008$ & $-0.427$ \\
Flash-Next & Code & $+0.061$ & $-0.089$ & $-0.563$ \\
Kimi K3 & Free & $+0.283$ & $+0.012$ & $-0.355$ \\
Kimi K3 & Table & $+0.298$ & $+0.021$ & $-0.354$ \\
Kimi K3 & Code & $+0.213$ & $-0.029$ & $-0.535$ \\
GPT-5.6 Sol & Free & $+0.335$ & $+0.013$ & $-0.533$ \\
GPT-5.6 Sol & Table & $+0.345$ & $+0.013$ & $-0.303$ \\
GPT-5.6 Sol & Code & $+0.256$ & $-0.039$ & $-0.240$ \\
\bottomrule\end{tabular}
\end{table}

\subsection{Model-level process quality and Capture}\label{app:cotmodelassociation}
\Cref{fig:cot}b shows a positive model-level association between L3 branch-state process
skill and final Capture completion. Each score averages Omni-ID and OOD equally, then
Table and Code equally, yielding one point per model, with Free excluded. \Cref{tab:cotprocesscapture} compares model-averaged, pooled, and
protocol-specific associations. The L1 comparison pairs L1 process skill with equally
weighted, reference-normalized Quantity and Event scores.
\begin{table}[t]\centering\scriptsize
\caption{Associations between process and final-answer scores. Rows use L3 state skill and Capture unless marked L1. Model means average Table and Code before computing correlations. Completion follows the frozen main-table normalization. These are descriptive associations.}
\label{tab:cotprocesscapture}
\begin{tabular}{@{}lrrr@{}}\toprule
Aggregation & Points & Pearson $r$ & Spearman $\rho$ \\ \midrule
Model--protocol pairs & 10 & $+0.384$ & $+0.539$ \\
Table only & 5 & $+0.556$ & $+0.300$ \\
Code only & 5 & $+0.308$ & $+0.800$ \\
Model means, raw Capture & 5 & $+0.699$ & $+0.800$ \\
Model means, Capture completion & 5 & $+0.693$ & $+0.800$ \\
Model means, L1 process and L1 completion & 5 & $+0.872$ & $+0.800$ \\
\bottomrule\end{tabular}
\end{table}

The model-mean Pearson correlation is $\cotModelCapturePearson$ and Spearman correlation
is $\cotModelCaptureSpearman$. These associations support intermediate-state prediction
quality as a complementary diagnostic of intervention forecasting.

A 20,000-draw model bootstrap gives a Pearson
95\% interval of $[\cotModelCaptureLow,\cotModelCaptureHigh]$, excluding degenerate resamples.
The pooled ten model--format pairs yield Pearson $\cotProcessCapturePearson$ with a
model-cluster interval of $[\cotProcessCaptureLow,\cotProcessCaptureHigh]$.
Both intervals include zero, reflecting uncertainty in this five-model comparison.

\Cref{fig:cot}c expresses Capture using the main table's frozen reference normalization.
For each setting, completion is $100(s-f)/(c-f)$ with $f=-1$ and that setting's Capture
ceiling $c$. The two settings are then averaged equally, without additional clipping.
Models follow increasing main-table SWFI, with connecting lines between categorical
labels. \Cref{fig:cot}b uses the same normalization before averaging formats.

\paragraph{Protocol-dependent gains.}
Beyond the gains in \cref{sec:cot}, weaker models often lose Capture under structured
protocols, so process supervision should be adapted to model capability. Comparisons use the
same anchors within each model, with the smaller Sol sample and Code's execution and feedback
resources described above.

State accuracy and effect accuracy can move differently. Sol's Table exceeds Code on
the measured intermediate-state skill, while Code achieves better final Capture.
Kimi exhibits the reverse pattern on intervention branch-state skill and Capture.
Thus, process supervision should target both states and intervention differences,
with downstream Capture evaluated on held-out worlds.

\section{Negative design results}\label{app:neg}
\paragraph{Report reconstruction and short horizons.}
Generating an entire future report conflates dynamics with serialization and copying
unchanged fields. The benchmark instead scores explicit quantities and events. Very short
observational horizons also leave many quantities unchanged, favoring persistence.
Longer budget-specific horizons were retained for L1, while L3 retains short horizons when
an operator's immediate effect would otherwise fall outside its declared scoring band.

\paragraph{Hidden-target identifiability.}
Early absolute trait bins were weakly recoverable even from privileged behavioral features,
motivating within-world rankings. Several world parameters were removed as inference targets
because their effects were absent or ambiguous in the rendered window. Illness illustrates
this distinction: it affects engine dynamics but is neither explicitly rendered nor narrated,
so city-size changes alone cannot identify its cause. Growth and research questions now require
relevant observed changes. These exclusions concern the available evidence, not causal inactivity.

\paragraph{Intervention persistence and observability.}
Some savegame fields are recomputed by the AI after reload, so a successful file edit does
not establish a persistent intervention. Government changes are therefore induced indirectly
through technology access. Wonder-grant and composite-trait candidates were removed or replaced
when their scored effects were weak or their descriptions did not identify the complete edit.
The retained single-trait operators modify one dimension at a time. Attitude interventions
remain difficult because changed attitudes need not alter the scored war or alliance counts.
This is a limitation of downstream observability, not evidence that attitudes are irrelevant.

\Cref{app:questions,app:ceiling} specify the resulting rules and references.

\section{Engine-field semantics and validation}\label{app:traps}
\paragraph{Unit counts.}
Total unit counts include civilian units. Military and civilian counts are therefore
reconstructed separately from unit rosters and ruleset attack values.

\paragraph{Research accounting.}
Research progress records accumulated points, not one-turn research production.
Team games further index research by team rather than player slot. The renderer and truth
builder use this mapping and compute technologies learned from known-technology counts,
excluding repeated Future Technology advances.

\paragraph{War preparation.}
The diplomatic countdown is not uniformly a war timer. The renderer accepts values from
1 to 12 for active preparations and verifies the relationship and target before reporting
a war plan.

\paragraph{Event time and identity.}
Engine events and state-difference events use different timestamps. A difference attached
to state $t$ describes the transition during turn $t-1$, whereas an engine event at $t$
occurs during turn $t$. The builder normalizes this convention before constructing histories
and targets. It also normalizes recoverable name-encoding errors, excludes destroyed players
from relationship counts, and removes illness suffixes from displayed ruleset names.

\section{Prompt examples}\label{app:prompts}
\subsection{Shared context and forecast format}
Prompts combine observation frames, interval narratives, the agents' current policies where
permitted, and a task block. The excerpts below use turn-60 questions from a three-civilization world with
a $24\times32$ map, the civ2civ3 ruleset, and hard AI difficulty. Report rows are omitted
for space, and ground-truth metadata are excluded from model inputs.

\begin{quote}\small\ttfamily
===== OBSERVATION FRAME 1/3 turn 20 (with map) =====\\
\#\# WORLD REPORT\\
turn 20, year -3050\\
ruleset civ2civ3\\
{[map, civilization, city, unit, and relationship rows omitted]}\\
{[subsequent narratives and frames at turns 40 and 60 omitted]}
\end{quote}

The quantitative output grammar is unchanged across viewpoints. For this global-view example, three civilizations, eleven quantities, and two horizons require 66 lines.

\begin{quote}\small\ttfamily
<civilization> | <quantity> | +<horizon> | p10 | p25 | p50 | p75 | p90\\
Civilizations (3): Hannibal Barca, Leonidas, Hatshepsut\\
Horizons (2): +60, +120
\end{quote}

Under the egocentric condition, the report is headed
\texttt{WORLD REPORT (as Hannibal Barca believes it)} and unexplored map tiles are marked
\texttt{?}. At this anchor the viewer has not contacted the other civilizations, so the
question scope contains Hannibal Barca alone. Hidden truth is retained for scoring but
is not appended to the prompt.

\subsection{Events, hidden targets, and interventions}
Binary event items state their empirical reference probability. One retained question asks
whether Leonidas forms a new alliance between turns 60 and 180 and supplies a base rate of
0.26. Leader questions instead request a distribution over candidate civilizations.
Trait responses use a ranking grammar, such as
\texttt{Aggressive: <civilization> > <civilization> = <civilization>}.
Rule responses use \texttt{<setting> | <label>}, with only evidence-eligible settings requested.
These strings illustrate syntax, not example correct answers.

The real government-enabling intervention at this anchor states:
\begin{quote}\small
Hannibal Barca acquires the technology The Republic.
\end{quote}
The numerical question then requests the same forecast format at the operator's eligible
horizon. The corresponding no-intervention prompt supplies the model reference, while
the recorded factual rerun supplies the outcome reference. A technology grant creates
the opportunity for an AI government change rather than guaranteeing that change.

\subsection{Verifiable Table and Code outputs}
Table prepends prescribed intermediate predictions to the final answer. Numeric claims use
\texttt{V | <civilization> | <quantity> | <turn> | <prediction>}.
Other supported claims concern diplomatic relations, governments, technologies, city ownership,
and wonders. The aligned study specifies required numeric cells before inference, so a model
cannot choose only easy claims for its process score.

Code defines \texttt{predict(world, rng)} and returns
\texttt{\{"trajectory": \{turn: \{civilization: state\}\}\}}.
The following schematic shows the execution interface, not a fitted model or a supplied answer.
\begin{quote}\small\ttfamily
def predict(world, rng):\\
\hspace*{1em}state = initialize(world["frames"])\\
\hspace*{1em}trajectory = \{\}\\
\hspace*{1em}for turn in forecast\_turns(world):\\
\hspace*{2em}state = transition(state, world, rng)\\
\hspace*{2em}trajectory[turn] = snapshot(state)\\
\hspace*{1em}return \{"trajectory": trajectory\}
\end{quote}
All helper functions and dynamics in a submitted program are generated by the model.
The harness executes 50 sampled trajectories, derives final quantiles or event probabilities,
and checks intermediate predictions against reserved engine states. Repair feedback uses
execution errors and observed-history back-tests, never future evaluation truth.
\Cref{app:cot} specifies process scoring, and \cref{app:cards} reproduces the mechanism cards.

\section{Reproducibility}\label{app:repro}
Reproduction uses the fixed prompts and reference values, with \nAnchors\ decision points
for each open-weight model, a shared stratified \nApiAnchors-anchor sample for Opus~5,
GPT-5.6~Sol, and \expert, and 60 shared anchors for additional external-validation models.
Run open-weight models through vLLM with native chat templates and thinking enabled, using
temperature 1.0, top-$p$ 0.95, and top-$k$ 20. Reserve 16,384 reasoning tokens, or 65,536 for DeepSeek, plus the larger of 1,024 tokens and the task-specific answer estimate.
The context window is 262,144 tokens, except gpt-oss at 131,072, with the output allowance
reduced when needed to fit. Qwen models use medium effort, Kimi uses high effort, and Inkling
uses an NVFP4 checkpoint. Both main API models use medium effort, with adaptive thinking
for Opus. \Cref{app:expertprotocol} specifies the expert manual and harness.
Nemotron~3 Ultra returned unparseable intervention answers on 33--43\% of its required cells, and Inkling on 7--14\%. The failure penalties in \cref{app:metrics} produce Nemotron's
negative Direction and Specificity scores. The benchmark engine image is
\texttt{freeciv-server:3.1.6-v2}, and the agent-play image adds the task-persistence patch.

Keep reserved engine states and future outcomes outside model inputs. Separate reasoning
from final answers, parse each required prediction, and apply the eligibility rules and
missing-answer penalties in \cref{app:metrics}. Compute component scores and SWFI using
the fixed reference scales and weights in \cref{sec:protocol}. Retain raw responses and
model configurations, and regenerate tables and figures from these outputs without new
inference. \Cref{app:cot,app:study,app:agentplay} specify the process, ablation, and decision
experiments, including their sampling and uncertainty procedures.

\section{Agent-play details}\label{app:agentplay}
\subsection{Scope and protocol separation}
Agent play measures decision utility separately from observer-only prediction. The full-game
experiment uses 76 selected worlds from turn 60 to 210, with checkpoints at turns 160 and 210.
Comparisons share the planning protocol and endpoint, without pooling earlier development
variants. The sample includes benchmark worlds and therefore does not test transfer to an
independent world distribution.

\subsection{Actions and shared information}
The actor controls one civilization through structured unit, city, research, budget, and
diplomacy commands validated against the engine interface. Observations contain the actor's
available information. Refused legal actions, such as treaties, count as outcomes, whereas
interface and model-call failures are recorded separately.

All arms share the actor, observations, rules reference, planning opportunities, and execution
feedback. The no-WM control retains its own plan. Assisted arms additionally receive forecasts
and advice before revision. The adopted plan and discussion remain available during execution,
separating WM assistance from ordinary planning support.

\subsection{Planning and verification}
Planning occurs initially and at intervals of at most ten turns, with earlier replanning
allowed after material changes. The agent proposes one to three plans and receives at most
three forecasting rounds. Forecasts include score and loss distributions, explicit
assumptions, and possible opponent responses. Finalization requires a complete adopted plan.
If a final revision differs from the last forecast plan, the logged prediction is not treated
as a verified forecast of the revision.

Execution records track positions, city ownership, and garrisons. Resumption retains the
saved state and protocol revision, while schema repairs and validation failures are logged.
These checks verify execution rather than every natural-language assumption.

\subsection{Outcome analysis}
Comparisons pair worlds at the same endpoint and report score differences, wins, coverage,
and world-level uncertainty. Repeated executions estimate stochastic variation, while
forecast calibration and execution logs distinguish prediction from implementation errors.
Only completion-validated runs enter the reported analyses.

\subsection{Full-game and local outcomes}
\Cref{tab:ap_main} reports positive mean gains at turn 210 across 76 worlds.
All 228 runs across the three conditions passed completion validation. Approximate paired $t$ intervals
are $[-4.2,17.3]$ for Flash-Next and $[-8.5,15.1]$ for Kimi, indicating substantial
between-world uncertainty. Runtime repairs and resumption revisions are recorded with the manifest.

\begin{table}[t]\centering\small
\caption{Turn-210 values underlying \cref{fig:ap_trajectories}. Differences are paired
against the no-WM planning control. $\pm$ denotes standard error across 76 worlds.}
\label{tab:ap_main}
\begin{tabular}{lrrr}
\toprule
World model & Mean score & $\Delta$ score & Wins / 76 \\
\midrule
None & 157.49 & --- & --- \\
Flash-Next & 164.00 & $+6.51\pm5.39$ & 44 \\
Kimi K3 & 160.80 & $+3.32\pm5.93$ & 39 \\
\bottomrule
\end{tabular}
\end{table}

The local study samples one decision point per world from control trajectories using
seed 20260924. The first 12 worlds are exploratory. The next 36 are disjoint from them,
with 12 near each of turns 80, 120, and 170. All conditions begin from the same saved
state and original proposal within a world. If the historical proposal contains
alternatives, the first is retained by a fixed rule. Each self-review or WM condition
receives one review/advice opportunity and one agent revision, with the option to retain
the original. Revision opportunities are matched, while token budgets may differ.
The six finalized plans are executed twice, followed by two additional executions on the
36 validation worlds, totaling 1,008 local branches and 20,160 turns. All passed completion
validation. One malformed planning response was retried and retained.

The four WMs forecast the same six finalized plans in two batches of three, with shuffled
labels and no author identity or execution results. Forecasts precede the first
executions. Flash-Next and Kimi also repeat them before the additional executions. Repeated
executions evaluate a fixed plan using stochastic actor calls from the same engine checkpoint
without independent engine reseeding. The world remains the independent sampling unit.
We use 5,000 world-bootstrap samples with a fixed seed and report individual 95\% intervals
without multiple-comparison adjustment.

\begin{table}[t]\centering\small
\caption{Local gain relative to self-review, mean $\pm$ SE. The 36-world replication
re-executes the same finalized plans. The final column pools four
executions per plan. The exploratory 12 worlds are excluded from these columns.}
\label{tab:ap_replication}
\begin{tabular}{lrrr}
\toprule
WM & First two executions & Next two executions & Four executions \\
\midrule
Flash-Next & $+0.39\pm1.17$ & $-0.01\pm0.78$ & $+0.19\pm0.79$ \\
Kimi K3 & $-0.15\pm0.93$ & $-1.17\pm1.65$ & $-0.66\pm1.01$ \\
Sol & $+1.82\pm0.99$ & $+0.97\pm1.16$ & $+1.40\pm0.84$ \\
Opus & $+0.96\pm1.00$ & $+1.61\pm1.15$ & $+1.28\pm0.91$ \\
\bottomrule
\end{tabular}
\end{table}

\Cref{tab:ap_replication} reports the 36-world validation sample separately from the
12 exploratory worlds. Four-execution 95\% intervals versus self-review are
$[-1.38,1.69]$, $[-2.70,1.33]$, $[-0.26,3.00]$, and $[-0.47,3.11]$ for Flash-Next,
Kimi, Sol, and Opus, respectively.

\paragraph{Outcome-informed selection.}
For each world, we select the plan with the highest mean score in executions 1--2 and
evaluate it on executions 3--4, then reverse the roles and average. Tied plans receive
their mean held-out score. Expanding the pool from original and self-reviewed plans to
all six gains $2.36\pm0.90$, with interval $[0.80,4.19]$. This post-hoc pool comparison shows that
WM-generated plans supply useful alternatives for execution-based selection. The six-plan
selector also gains $2.90\pm0.86$ over the original plan and $2.44\pm0.63$, interval $[1.29,3.78]$,
over uniform selection from the same pool. Selection uses actual rollout outcomes, and
the expanded pool requires additional generation compute. Forecast-only selection yields
gain intervals containing zero for all four WMs. Thus, useful plan generation and reliable
forecast-based selection remain distinct capabilities.

\paragraph{Capability comparisons.}
The paired Opus-minus-Kimi contrast on four executions is $1.94\pm0.76$ with interval $[0.55,3.54]$, and Sol-minus-Kimi is $2.06\pm1.15$ with
interval $[-0.11,4.33]$.
This exploratory
four-model comparison relates \omniID\ diagnostics to fog-of-war utility. Estimating
independent level contributions or fitting index weights requires a larger model panel.

\Cref{tab:ap_capability_summary} reports level completion on \omniID. Each metric is
normalized by its frozen floor and ceiling, then averaged equally within its level
and multiplied by 100, following \cref{eq:swfi}. \Lone\ combines quantity and event,
\Ltwo\ combines trait and rules, and \Lthree\ combines capture, direction, and specificity.
Completion percentages are reported without clipping.

\begin{table}[t]\centering\small
\setlength{\tabcolsep}{4pt}
\caption{Component diagnostics and utility accompanying \cref{tab:ap_capability_summary}.
Diagnostics are \omniID\ scores
from \cref{tab:main}, while SWFI spans three settings. Utility averages four executions
per plan in 36 worlds and is reported relative to self-review, with world-level SE.}
\label{tab:ap_capability}
\begin{tabular}{lrrrrr}
\toprule
Model & \Lone\ quantity & \Ltwo\ trait & \Lthree\ capture & SWFI & $\Delta$ score \\
\midrule
Flash-Next & 0.26 & 0.10 & $-0.57$ & 16.93 & $+0.19\pm0.79$ \\
Kimi K3 & 0.34 & 0.00 & $-0.35$ & 21.94 & $-0.66\pm1.01$ \\
Sol & 0.27 & 0.23 & $-0.60$ & 27.68 & $+1.40\pm0.84$ \\
Opus & 0.33 & 0.17 & $-0.08$ & 29.47 & $+1.28\pm0.91$ \\
\bottomrule
\end{tabular}
\end{table}

\subsection{Case study: from accepting collapse to targeted defense}
\label{app:agentplay_case}
The turn-173 case is selected retrospectively for strategic disagreement
and improvement in both execution batches. Tupi has nine cities, score 161, and treasury 83,
under threat from Barbarossa's Frigates and Riflemen. \Cref{fig:ap_case_dialogue} traces
the original plan, one Opus consultation, one agent revision, and recorded engine feedback.

\begin{figure*}[ht]
\begin{promptbox}[A decision reversal: accept city losses or attempt targeted defense?]
\textbf{Checkpoint.} Turn 173: score 161, 9 cities, 83 gold. One consultation with Opus, followed by a Flash-Next revision and 20 turns of execution.
\par\textbf{Agent (Flash-Next): original plan.}\par
``Icara\'{i} and Siri Jibe have empty garrisons and P=1.00 regardless of actions; accept their loss.''\par ``Do NOT build anything at Icara\'{i} or Siri Jibe (likely lost).'' ``Do NOT buy anything.''
\par\textbf{World model (Opus): forecast and advice.}\par
Original-plan forecast at T193: score $[p_{10},p_{50},p_{90}]=[132, 155, 174]$; $P(\text{any city lost})=0.82$.\par
``'P=1.00' is P(loss GIVEN assault), not P(assault)'' $[\ldots]$ ``Realistic near-term capture risk is therefore moderate, not certain.''\par
``an empty city is captured by any single unit, a walled city with even one defender forces a fight.''\par
``Reconsider the Capuan triage: it is the most likely city to be assaulted NEXT turn (adjacent Riflemen), and one more fortified Musketeer takes P(loss\textbar{}assault) from 0.81 to 0.03 per the calculator $[\ldots]$ consider buying it if gold allows (\textasciitilde{}60-90 gold) rather than hoarding 83 gold that will drip away anyway.''
\par\textbf{Agent (Flash-Next): revised plan.}\par
``T173 BUY: Buy Capuan Musketeers (20/40 shields remaining).''\par ``T173 DEFENSE: Order Musketeer 520 from Par\'{a}yba (23,2) to Icara\'{i} (21,1).''\par
Siri Jibe: ``switch from Colosseum [45/70 stored] to Musketeers [40].''
\par\textbf{Execution feedback (first execution, T173).}\par
{\footnotesize\ttfamily OK bought Musketeers in Capuan for 60 gold (completes at turn change)}\par
{\footnotesize\ttfamily OK Siri Jibe now builds Musketeers (half the stored shields lost: category change)}\par
{\footnotesize\ttfamily moved unit 520 Musketeers to 21,1}\par
\textbf{Measured score trajectory (mean of four executions per frozen plan).}\par
{\centering\setlength{\tabcolsep}{8pt}
\begin{tabular}{lrrrrr}
\toprule
Plan & T173 & T178 & T183 & T188 & T193 \\
\midrule
Original & 161.00 & 147.75 & 127.00 & 113.50 & 102.25 \\
Self-review & 161.00 & 149.00 & 125.00 & 113.00 & 109.00 \\
Opus-assisted & 161.00 & 147.25 & 137.25 & 123.50 & 124.25 \\
\bottomrule\end{tabular}\par}
At T193, the revised plan reduces mean score loss by \textbf{22.00 points} relative to the original.
\end{promptbox}
\caption{Recorded agent--WM exchange and outcomes in the targeted-defense case. Quotations preserve model outputs, with $[\ldots]$ marking omissions. The forecast concerns the original plan, and 0.81/0.03 are estimates from the shared siege calculator. Recorded executions provide outcome evidence.}
\label{fig:ap_case_dialogue}
\end{figure*}

\begin{table}[t]\centering\small
\caption{Complete six-condition results for \cref{fig:ap_case_dialogue} after 20 turns.
Each pair independently executes the same finalized plan. The last column averages
remaining city counts across all four executions.}
\label{tab:ap_case}
\begin{tabular}{lrrrr}
\toprule
Plan & First pair & Second pair & Mean score & Mean cities \\
\midrule
Original & 106, 83 & 109, 111 & 102.25 & 1.75 \\
Self-review & 86, 147 & 84, 119 & 109.00 & 1.75 \\
Flash-assisted & 95, 95 & 117, 108 & 103.75 & 2.25 \\
Kimi-assisted & 96, 116 & 117, 106 & 108.75 & 1.50 \\
Sol-assisted & 118, 115 & 127, 139 & 124.75 & 3.50 \\
Opus-assisted & 134, 114 & 115, 134 & 124.25 & 3.75 \\
\bottomrule
\end{tabular}
\end{table}

\paragraph{Decision and outcome.}
\Cref{tab:ap_case} shows a 22-point mean gain and two additional surviving cities for
the Opus-assisted plan. Icara\'{i} survives three of four executions, versus none under
the original or self-reviewed plan. The agent translates advice into a Musketeer relocation.
The plan still loses most cities, and the original-plan forecast $[132,155,174]$ exceeds
the realized 83--111. The benefit comes from the revised plan as a whole and illustrates
useful strategic advice despite forecast error.

\subsection{Failure case: precaution without realized benefit}
\label{app:agentplay_bad_case}
The turn-172 case supplies a contrasting example under the same local
protocol. The original plan emphasizes research. Sol forecasts only a 0.07 probability
of losing any city within 20 turns, but recommends earlier tax increases and additional
defenses. The agent adopts those precautions and raises the treasury trigger from the
adviser's 300--350 gold to roughly 400--450 in its revised plan.
\Cref{fig:ap_bad_case_dialogue} shows the recorded exchange and executed changes.

\begin{figure*}[ht]
\begin{promptbox}[A failed revision: precaution without a realized score benefit]
\textbf{Checkpoint.} Turn 172: score 148, 8 cities, 665 gold. One consultation with Sol, followed by a Flash-Next revision and 20 turns of execution.
\par\textbf{Agent (Flash-Next): original plan.}\par
``PRIMARY LEVER: tech at 69 bulbs/turn.''\par ``Treasury 665, -13/turn, ends \textasciitilde{}T210 at \textasciitilde{}160''
\par\textbf{World model (Sol): forecast and advice.}\par
Original-plan forecast at T192: score $[p_{10},p_{50},p_{90}]=[158, 168, 178]$; $P(\text{any city lost})=0.07$.\par
``Begin raising taxes if the treasury falls below roughly 300-350 or if the observed deficit remains above 10 gold/turn; waiting for 200 risks forced building sales.''\par ``Oviedo and Santander particularly benefit from City Walls or a second defender''
\par\textbf{Agent (Flash-Next): revised plan.}\par
``Must raise taxes to 60/30/10 by \textasciitilde{}T182-185 when treasury nears 400 to prevent forced building sales.''\par ``DEFENSE: City Walls (0 upkeep, +100\% defense) on Oviedo, Santander, Valencia''
\par\textbf{Execution feedback (first execution).}\par
T176: {\footnotesize\ttfamily OK rates tax 60\% science 30\% luxury 10\%}\par
T177: {\footnotesize\ttfamily OK Valencia now builds City Walls}\par
T179: {\footnotesize\ttfamily OK Oviedo now builds City Walls}\par
\textbf{Measured score trajectory (mean of four executions per frozen plan).}\par
{\centering\setlength{\tabcolsep}{8pt}
\begin{tabular}{lrrrrr}
\toprule
Plan & T172 & T177 & T182 & T187 & T192 \\
\midrule
Original & 148.00 & 153.50 & 159.50 & 172.00 & 179.50 \\
Self-review & 148.00 & 153.00 & 162.75 & 170.00 & 177.75 \\
Sol-assisted & 148.00 & 152.50 & 156.25 & 164.75 & 171.50 \\
\bottomrule\end{tabular}\par}
The revision yields \textbf{8.00 fewer score points} at T192. All three conditions retain all 8 cities throughout. Mean science output is 56.4 versus 46.3 bulbs/turn (original versus Sol-assisted); final treasury is 386.0 versus 423.5. The original executions never end a turn below 358 gold.
\end{promptbox}
\caption{The precautionary revision retained more gold but reduced research output and final score. Quotations reproduce logged advice and plans. The forecast concerns the original plan, while outcomes reflect the complete revision.}
\label{fig:ap_bad_case_dialogue}
\end{figure*}

At T192, the four original executions score 175, 194, 175, and 174, compared with
173, 171, 170, and 172 under Sol assistance. The mean difference is $-8.00$ points,
with $-12.5$ in the first batch and $-3.5$ in the replication. The self-review mean is 177.75,
also above the assisted mean of 171.50. All three conditions retain eight cities
throughout. Relative to the original, the assisted plan produces fewer research bulbs per turn, 46.3 versus 56.4, finishes with
fewer technologies, 35.25 versus 36.50, and retains more gold, 423.5 versus 386.0. The original executions never end a turn below
358 gold.

The agent follows and strengthens the precautions, incurring a resource cost without a
realized survival benefit. The score gap reflects the combined plan revision, including
taxation and defense changes. Together, the retrospectively selected cases motivate state-specific
advice that balances strategic risks against opportunity costs.

\paragraph{Evidence verification.}
Quoted advice and plans are checked against recorded consultations, and outcomes against
engine trajectories and execution feedback.
The main gain plot clips confidence bands to $[-1.5,7.5]$ for readability, while the
data retain their full bounds. These are pointwise, not simultaneous, intervals.

\section{Mechanism cards and the direct-effect check}\label{app:cards}
\paragraph{Mechanism-card protocol.}
The eleven cards derive from engine source, patches, and the expert manual. Checks exclude
civilization names from the evaluated anchor and verify consistency with the engine.
Each operator uses the same text across worlds and real/placebo arms, with edit magnitudes
and outcomes withheld. The following paragraphs reproduce the intervention-specific prompt
text verbatim.
\Cref{app:mechanism} specifies the paired comparison and coverage.

\paragraph{Independent direct-effect check.}
The diagnostic examines the target civilization's directly affected quantity at the shortest
asked horizon for four visible-channel operators. It compares the sign of the model median's
shift from its own control with the realized arm-minus-factual sign, and separately records
zero responses and the absolute effect-magnitude ratio. Flash-Next's sign agreement is
\directEffectSign\ across the three settings, with zero responses on \directEffectNoMove\
of cells and median magnitude ratios of \directEffectRatio. These are conditional summaries,
not Capture scores. The diagnostic includes city count outside the main scoring mask,
and median ratios respond differently to outliers than Capture. It therefore complements
the channel analysis in \cref{app:mechanism} without replacing it.

\begingroup
\small
\paragraph{Treasury (gold stock).} The treasury is the civilization's stock of gold. Every turn the engine adds income (trade times the tax rate, plus trade-route revenue and the Palace bonus) and subtracts gold upkeep for buildings and, under Despotism, Monarchy, Fundamentalism, Democracy and Federation, for units beyond the free ones per city (Tribal, Republic and Communism pay unit upkeep in shields instead). When the treasury cannot cover upkeep the engine sells a building or disbands units, and the AI reacts within a few turns by moving rate points from science to tax and by stopping rush-buys; a large treasury lets it keep science high, rush-buy defenders in threatened cities and pay for bribes and incitements. The edit changes the stock once; income and upkeep rates are unchanged, so the difference is worked off only through the AI's own responses (rate points moved between tax and science, rush-buying or not), typically over some tens of turns, during which the treasury drifts back toward its trend while the number of units and buildings moves slightly in the same direction as the edit. It does not change technologies known, city sizes, terrain, diplomatic relations or personality.

\paragraph{Research points (bulbs).} Research points accumulate toward the current research target; when the stock reaches the target's cost the technology is learned, the surplus carries over, and the AI picks the next step toward its research goal. If the added points reach the current target's remaining cost the target is learned at once (and possibly the following ones); otherwise the target simply arrives earlier. Either way technologies arrive earlier by roughly (points added) divided by (points per turn) turns, and the number learned in the window rises by roughly (points added) divided by the current cost, which grows linearly along the tree in civ2civ3 and more steeply in classic. New technologies unlock units, buildings and governments that the AI starts using, add to score, can trigger a government change when a government technology arrives, may be passed to allies by treaty and stolen by neighbours' diplomats. The edit does not change gold, cities, terrain or relations directly.

\paragraph{Destruction of cities.} Destroying a city removes it with its citizens and buildings and any wonder inside; units homed there are disbanded or rehomed, its worked tiles and trade routes are freed, and the owner's city count, total city size, population (5 x size x (size+1) per city), production, trade and science drop at once by the destroyed cities' share, disproportionately so when the destroyed cities are large ones, since population is convex in size; culture loses the destroyed cities' own accumulated culture while the national history is kept. The AI does not rebuild quickly: Settlers are only built where the founder want is high, so the count recovers at the civilization's normal expansion pace if free land is within reach and otherwise not at all, and neighbours may settle the freed land. Score falls with the lost citizens and later with the slower technology pace. Its standing as a war target moves in both directions: fewer and smaller cities lower the value the AI's target formula assigns to it, while fewer defending units lower the fear term, so neighbours may or may not become more inclined to attack. Personality, technologies known and diplomatic states are not changed by the edit.

\paragraph{A technology is granted.} Granting a technology adds it to the civilization's known set without spending research points; if it was the current research target the accumulated points carry to the next target, and if it was a prerequisite the path to the research goal shortens. If the technology enables a government, the AI re-evaluates every available government every few turns (about 5 to 20, growing with the city count, and immediately after a reload) and switches when the new one scores higher for its situation, so the change follows the enabling technology within that lag and passes through one to five turns of Anarchy (no taxes or science, culture stalls) before the new rules take effect: for example Communism pays no gold for 1-coin buildings and has no distance-based corruption but a lower rate cap (80\%) than Democracy (90\%); Republic adds trade and pays unit upkeep in shields, while Democracy adds the same trade but pays doubled gold upkeep for units, and both suffer war weariness. A technology that unlocks units or buildings changes what cities build over the following 10 to 30 turns, and allies may receive it by treaty. The edit does not change gold, cities, terrain or relations directly.

\paragraph{Map knowledge revealed.} Revealing map tiles gives one civilization knowledge of the revealed tiles' terrain and of the cities standing on them at that moment; what it later sees is again limited to its vision. The AI uses map knowledge for settling (the founder want only considers known free sites), for exploration (explorers stop when nothing is unknown), for routing units and ferries, and for target evaluation (only known cities can be targeted). The effects are indirect and modest: fewer explorer units, possibly earlier or farther settling if free land exists, and earlier contact with distant civilizations, which starts relations at Peace and makes embassies, treaties and later wars with them possible. The edit does not change any quantity of the civilization directly, nor its technologies, relations or personality.

\paragraph{Personality: Aggressive.} Aggressive (10 to 150, default 50) scales the AI's appetite for war. In this engine it scales the grudge the AI builds each turn against its preferred war target before declaring (about 5 attitude points per turn at 10, 25 at 50, 75 at 150), raises the attitude threshold below which it declares war so that less hatred is needed (about -360 at 10, -210 at 50, -15 at 150), lowers how much hatred still allows a ceasefire (love above -850 at 10, above -300 at 150), and scales the minimum time after a declaration during which it refuses any ceasefire (about 2 to 4 turns at 10, 10 to 20 at 50, 30 to 60 at 150). An upward shift makes a declaration against the most disliked contact more likely within 20 to 40 turns, in proportion to its size, raises the military share of production and the military unit count, drains the treasury under gold-upkeep governments and worsens the target's relations; a downward shift makes the civilization avoid starting wars, though allies can still drag it in. It has no direct effect on research, growth, production formulas or world rules.

\paragraph{Personality: Expansionist.} Expansionist (10 to 150, default 50) scales the founder want with which cities decide to build Settlers and the value the AI puts on free city sites. A higher value means more Settlers and more cities founded per 100 turns, but only where free land within reach exists, since land is the binding constraint; more small cities raise the city count and, slowly, total size, population and score, while spreading units and upkeep thinner. A lower value slows or stops founding while the existing cities grow as before. It does not affect war behaviour, research, tax rates or relations directly, although a civilization with many Settlers is a slightly more attractive war target in the AI's formula and borders growing into a neighbour's land raise tension.

\paragraph{Personality: Trader.} Trader (10 to 150, default 50) scales, linearly in this engine, the AI's want for Caravans built to establish trade routes; caravans for helping wonders are requested by the wonder city's need, although any caravan may be redirected to help a wonder. Trade routes add permanent trade to both cities (routes between a civilization's own cities are allowed here). A higher value shows as more Caravans in the unit list (civilian count up), more trade routes and hence more gold and science over 30 to 100 turns, and possibly faster wonder completion; a lower value the opposite. It does not affect military production, war behaviour, growth or research rules directly.

\paragraph{Personality: Builder.} Builder (10 to 150, default 50) discounts the military advisor's want where it competes with the domestic advisor in each city (about x1.67 at 10, x1.0 at 50, x0.5 at 150), except when a city is in danger or the civilization is on a war footing, and scales the building wants themselves in the same direction (about x0.5 at 10, x2.3 at 150). A higher value shifts production from units to buildings and wonders: fewer military units per city, more improvements (and with them culture, science and gold multipliers) and somewhat faster score growth; a lower value the opposite. It does not directly change war declarations, research, growth or relations, though a smaller army makes the civilization a somewhat more attractive war target.

\paragraph{Hidden attitude (love) toward one player.} Love (-1000 to +1000) is the AI's hidden attitude toward one other player, recomputed every turn: about +15 per turn in any non-war state (+10 more when allied, +7.5 per shared enemy, and +60 more while a ceasefire holds), -15 per turn at war plus penalties for that player's units inside its borders, a 4 percent decay toward zero, and the grudge it builds toward its chosen war target. It gates diplomacy: war is declared when love falls below the aggression-dependent threshold and the player has been chosen as target; a ceasefire needs love above the ceasefire threshold, a peace treaty above +125, an alliance above +250 from Peace with no war countdown. The edit shifts the stock once and the per-turn dynamics then pull it toward the level the current relation supports, so a large drop can start a war countdown or a declaration against that player within a few to a few dozen turns (and the target's relations and enemy count change accordingly), while a large rise can lead to a ceasefire or peace if at war, or to an alliance from Peace if love stays high. It does not change the other player's attitude, nor any economic quantity directly.

\paragraph{World rule: disaster frequency.} The disaster setting is a world rule: the per-city, per-turn chance of a random disaster (Earthquake or Fire destroys a building, Flood costs food or a citizen, Plague costs a citizen, Famine empties the granary, Industrial Accident creates pollution). Multiplying the setting multiplies the expected number of such events proportionally in every city of every civilization from the edit onward; each event costs a building, a citizen, a granary or a polluted tile, so total city size, population and culture grow slightly more slowly and production is diverted to rebuilding. The effect is world-wide, small per event and accumulates with city count and time; it does not change relations, research rules, personality or tax rates.

\par
\endgroup

\end{document}